\documentclass[10pt,twocolumn,letterpaper]{article}

\usepackage[utf8]{inputenc}
\usepackage{cvpr}
\usepackage{times}
\usepackage{epsfig}
\usepackage{graphicx}
\usepackage{amsmath,amsfonts,amssymb,amsthm}
\usepackage{color}
\usepackage{caption}
\usepackage{subcaption}
\usepackage{xspace}
\usepackage{array}
\usepackage{cite}
\usepackage[pagebackref=true,breaklinks=true,letterpaper=true,colorlinks,bookmarks=false]{hyperref}
\usepackage{booktabs}

\newcommand{\bI}{\mathbf{I}}

\newcommand{\bP}{\mathbf{P}}

\newcommand{\br}{\mathbf{r}}\newcommand{\bR}{\mathbf{R}}

\newcommand{\bt}{\mathbf{t}}

\newcommand{\bx}{\mathbf{x}}\newcommand{\bX}{\mathbf{X}}
\newcommand{\by}{\mathbf{y}}\newcommand{\bY}{\mathbf{Y}}
\newcommand{\bz}{\mathbf{z}}

\newcommand{\nE}{\mathbb{E}}

\newcommand{\cL}{\mathcal{L}}

\newcommand{\cT}{\mathcal{T}}

\newcommand{\figref}[1]{Fig.~\ref{#1}}
\newcommand{\secref}[1]{Section~\ref{#1}}

\newcommand{\tabref}[1]{Table~\ref{#1}}

\makeatletter
\DeclareRobustCommand\onedot{\futurelet\@let@token\@onedot}
\def\@onedot{\ifx\@let@token.\else.\null\fi\xspace}
\def\eg{e.g\onedot} 
\def\ie{i.e\onedot} 
 
\def\etc{etc\onedot}

\def\wrt{wrt\onedot}

\def\etal{et~al\onedot}

\makeatother

\newcommand{\boldparagraph}[1]{\vspace{0.2cm}\noindent{\bf #1:} }

\definecolor{darkgreen}{rgb}{0,0.7,0}
\definecolor{orange}{RGB}{255,127,0}

\newcolumntype{C}[1]{>{\centering\arraybackslash}p{#1}}

\cvprfinalcopy
\def\cvprPaperID{5106}

\ifcvprfinal\fi

\begin{document}

\title{Superquadrics Revisited: Learning 3D Shape Parsing beyond Cuboids}

\author{Despoina Paschalidou$^{1,4}$ \quad Ali Osman Ulusoy$^{2}$ \quad Andreas Geiger$^{1,3,4}$\\
$^1$Autonomous Vision Group, MPI for Intelligent Systems T{\"u}bingen\\
$^2$Microsoft \quad
$^3$University of T{\"u}bingen\quad
$^4$Max Planck ETH Center for Learning Systems\\
{\tt\small \{firstname.lastname\}@tue.mpg.de}}

\maketitle

\begin{abstract}
Abstracting complex 3D shapes with parsimonious part-based representations has been a long standing goal in computer vision. This paper presents a learning-based solution to this problem which goes beyond the traditional 3D cuboid representation by exploiting superquadrics as atomic elements. We demonstrate that superquadrics lead to more expressive 3D scene parses while being easier to learn than 3D cuboid representations. Moreover, we provide an analytical solution to the Chamfer loss which avoids the need for computational expensive reinforcement learning or iterative prediction. Our model learns to parse 3D objects into consistent superquadric representations without supervision. Results on various ShapeNet categories as well as the SURREAL human body dataset demonstrate the flexibility of our model in capturing fine details and complex poses that could not have been modelled using cuboids.
\end{abstract}

\section{Introduction} \label{sec:intro}

\begin{figure}[t!]
	\centering
	\begin{subfigure}[b]{1.0\linewidth}
		\centering
		\includegraphics[width=0.3\linewidth]{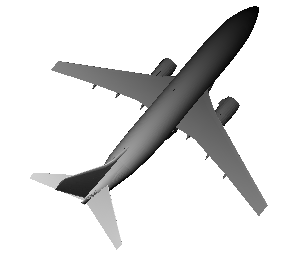}
	    \hfill
		\includegraphics[width=0.26\linewidth]{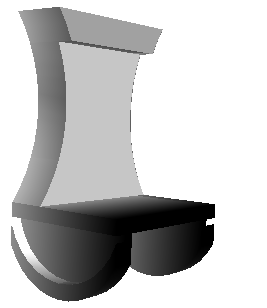}
	    \hfill
		\includegraphics[width=0.3\linewidth]{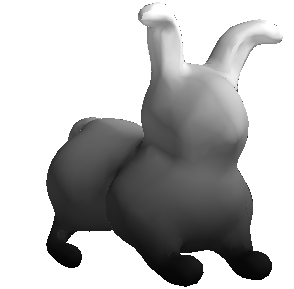}
        \caption{Input Mesh}
        \label{fig:teaser_input}
	\end{subfigure}\vspace{0.1cm}
    \vspace{-0.7em}
	\begin{subfigure}[b]{1.0\linewidth}
		\centering
		\includegraphics[width=0.3\linewidth]{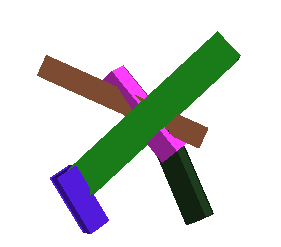}
	    \hfill
		\includegraphics[width=0.3\linewidth]{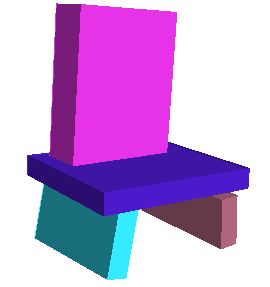}
	    \hfill
		\includegraphics[width=0.3\linewidth]{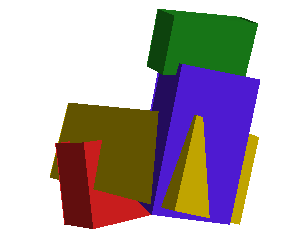}
        \caption{Inferred Cuboid Representation \cite{Tulsiani2017CVPRa}}
        \label{fig:teaser_cuboid}
	\end{subfigure}\vspace{0.1cm}
	\begin{subfigure}[b]{1.0\linewidth}
        \vspace{0.4em}
		\centering
		\includegraphics[width=0.3\linewidth]{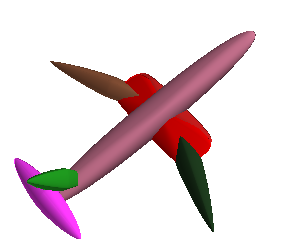}
        \hfill
		\includegraphics[width=0.28\linewidth]{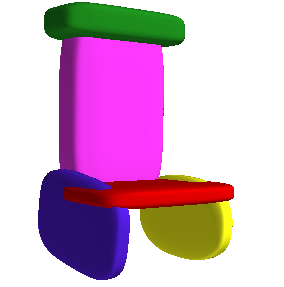}
	    \hfill
		\includegraphics[width=0.3\linewidth]{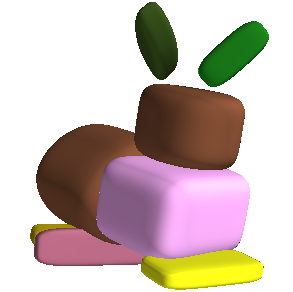}
        \caption{Inferred Superquadric Representation (Ours)}
        \label{fig:teaser_ours}
   \end{subfigure}
   \caption{\textbf{3D Shape Parsing.}
   	We consider the problem of learning to parse unstructured 3D data (\eg, meshes or point clouds) into compact part-based representations.
   	Prior work \cite{Tulsiani2017CVPRa, Zou2017ICCV, Niu2018CVPR} has considered cuboid representations (\subref{fig:teaser_cuboid}) which capture 
   	the overall object structure, but lack expressiveness.
   	In this work, we propose an unsupervised model for superquadrics (\subref{fig:teaser_ours}), which allows us to capture details such as the body of the airplane and the ears of the rabbit.
   }
   \label{fig:teaser}
   \vspace{-1.2em}
\end{figure}

Evolution has developed a remarkable visual system that allows humans to robustly perceive their 3D environment.
It has long been hypothesized \cite{Biederman1986Human} that the human visual system processes the vast amount of raw visual input into compact parsimonious representations, where complex objects are decomposed into a small number of shape primitives that can each be represented using low-dimensional descriptions.
Indeed, experiments show that humans can understand complex scenes from renderings of simple shape primitives such as cuboids or geons~\cite{Biederman1987PsychologicalReview}.

Likewise, machines would tremendously benefit from being able to parse 3D data into compact low-dimensional representations. Such representations would provide useful cues for recognition, detection, shape manipulation and physical reasoning such as path planning and grasping. 
In the early days of computer vision, researchers explored shape primitives such as 3D polyhedral shapes \cite{Roberts1963PhD}, generalized cylinders \cite{Binford1971Visual}, geons \cite{Biederman1986Human} and superquadrics~\cite{Pentland1986AAAI}. 
However, it proved very difficult to extract such representations from images due to the lack of computation power and data at the time.
Thus, the research community shifted their focus away from the shape primitive paradigm. 

In the last decade, major breakthroughs in shape extraction were due to deep neural networks coupled with the abundance of visual data. Recent works focus on learning 3D reconstruction using 2.5D \cite{Hartmann2017ICCV, Paschalidou2018CVPR, Pohan2018CVPR, Yao2018ECCV}, volumetric \cite{Choy2016ECCV, Girdhar2016ECCV, Wu2016NIPS, Ji2017ICCV, Haene2017ARXIV, Riegler2017CVPR}, mesh \cite{Groueix2018CVPR, Liao2018CVPR} and point cloud \cite{Fan2017CVPR, Qi2017NIPS} representations.
However, none of the above are sufficiently parsimonious or interpretable to allow for higher-level 3D scene understanding as required by intelligent systems.

Very recently, shape primitives have been revisited in the context of deep learning. %
In particular, \cite{Tulsiani2017CVPRa, Zou2017ICCV, Niu2018CVPR} have demonstrated that deep neural networks enable to reliably extract 3D cuboids from meshes and even RGB images. 

Inspired by these works, we propose a novel deep neural network to efficiently extract parsimonious 3D representations in an unsupervised fashion, conditioned on a 3D shape or 2D image as input. In particular, this paper makes the following \textbf{contributions}:

First, we note that 3D cuboid representations used in prior works~\cite{Tulsiani2017CVPRa, Zou2017ICCV, Niu2018CVPR} are not sufficiently expressive to model many natural and man-made shapes as illustrated in \figref{fig:teaser}. Thus, cuboid-based representation may require a large number of primitives to accurately represent common shapes. Instead, in this paper, we propose to utilize superquadrics, which
have been successfully used in computer graphics~\cite{Barr1981CGA} and classical computer vision~\cite{Pentland1986AAAI, Solina1990PAMI, Terzopoulos1990ICCV}. Superquadrics are able to represent a diverse class of shapes such as cylinders, spheres, cuboids, ellipsoids \emph{in a single continuous parameter space} (see \figref{fig:teaser}+\ref{fig:superquadrics}).
Moreover, their continuous parametrization is particularly amenable to deep learning, as their shape is smooth and varies continuously with their parameters. This allows for faster optimization, and hence faster and more stable training as evidenced by our experiments.

Second, we provide an analytical closed-form solution to the Chamfer distance function which can be evaluated in linear time \wrt the number of primitives. This allows us to compute gradients \wrt the model parameters using standard error backpropagation \cite{Rumelhart1986NATURE} without resorting to computational expensive reinforcement learning techniques as required by prior work~\cite{Tulsiani2017CVPRa}. We consequently mitigate the need for designing an auxiliary reward function.
Instead, we formulate a simple parsimony loss to favor configurations with a small number of primitives.
We demonstrate the strengths of our model by learning to parse 3D shapes from the ShapeNet \cite{Chang2015ARXIV} and the SURREAL \cite{Varol2017CVPR}. We observe that our model converges faster than \cite{Tulsiani2017CVPRa} and leads to more accurate reconstructions. Our code is publicly available\footnote{\url{https://github.com/paschalidoud/superquadric_parsing}}.

\section{Related Work} \label{sec:related}

In this section, we discuss the most relevant work on deep learning-based 3D shape modeling approaches and review the origins of superquadric representations.

\subsection{3D Reconstruction}

The simplest representation for 3D reconstruction from one or more images are 2.5D depth maps as they can be inferred using standard 2D convolutional neural networks
~\cite{Hartmann2017ICCV, Paschalidou2018CVPR, Ji2017ICCV, Yao2018ECCV}. Since depth maps are view-based, these methods require additional post-processing algorithms to fuse information from multiple viewpoints in order to capture the entire object geometry.
As opposed to depth maps, volumetric representations~\cite{Girdhar2016ECCV, Haene2017ARXIV, Choy2016ECCV, Riegler2017CVPR, Tatarchenko2017ICCV} naturally capture the entire 3D shape. While, hierarchical 3D data structures such as octrees accelerate 3D convolutions, the high memory requirements remain a limitation of existing volumetric methods. 
An alternative line of work \cite{Fan2017CVPR, Qi2007CVPR} focuses on learning to reconstruct 3D point sets. A natural limitation of these approaches is the lack of surface connectivity in the representation. To address this limitation, \cite{Liao2018CVPR, Groueix2018CVPR, Wang2018ECCV,Rezende2016NIPS} proposed to directly learn 3D meshes.

While some of the aforementioned models are able to capture fine surface details, none of them lends itself to parsimonious, semantic interpretations. In this work, we utilize superquadrics which provide a concise and yet accurate representation with significantly less parameters.

\subsection{Constructive Solid Geometry}

Towards the goal of concise representations, researchers exploited constructive solid geometry (CSG) \cite{Laidlaw1986SIGGRAPH} for shape modeling~\cite{Sharma2018CVPR, Ellis2018NIPS}. Sharma \etal \cite{Sharma2018CVPR} leverage an encoder-decoder architecture to generate a sequence of simple boolean operations to act on a set of primitives that can be either squares, circles or triangles. 
In a similar line of work, Ellis \etal \cite{Ellis2018NIPS} learn a programmatic representation of a hand-written drawing, by first extracting simple primitives, such as lines, circles and rectangles and a set of drawing commands that is used to synthesize a \LaTeX~program. 
In contrast to \cite{Sharma2018CVPR, Ellis2018NIPS}, our goal is not to obtain accurate 3D geometry by iteratively applying boolean operations on shapes. Instead, we aim to decompose the depicted object into a parsimonious interpretable representation where each part has a semantic meaning associated with it.
Besides, we do not suffer from ambiguities of an iterative construction process, where different executions lead to the same result.

\subsection{Learning-based Scene Parsing} 

Recently, shape primitives have been revisited in the context of deep learning \cite{Tulsiani2017CVPRa, Zou2017ICCV, Niu2018CVPR}.
Niu \etal \cite{Niu2018CVPR} propose to use a recurrent neural network (RNN) to iteratively predict cuboid primitives as well as symmetry relationships from RGB images. They first train an encoder which encodes the input image and its segmentation into a 80-dimensional latent code. Starting from this root feature, they iteratively decode the structure into cuboids, splitting nodes based on adjacency and symmetry relationships.
In related work, Zou \etal \cite{Zou2017ICCV} utilize LSTMs in combination with mixture density networks to generate cuboid representations from depth maps encoded by a 32-dimensional feature vector. However, both works \cite{Niu2018CVPR,Zou2017ICCV} require supervision in terms of the primitive parameters as well as the sequence of predictions. This supervision must either be provided by manual annotation or using greedy heuristics as in \cite{Niu2018CVPR,Zou2017ICCV}.

In contrast, our approach is unsupervised and does not suffer from ambiguities caused by different possible prediction sequences that lead to the same cuboid assembly. 
Furthermore, \cite{Niu2018CVPR,Zou2017ICCV} exploit simple cuboid representations which do not capture more complex shapes that are common in natural and man-made scenes (\eg, curved objects, spheres).
In this work, we propose to use superquadrics~\cite{Barr1981CGA} which yield a more diverse shape vocabulary and hence lead to more expressive scene abstractions as illustrated in \figref{fig:teaser}.

A primary inspiration for this paper is the seminal work by Tulsiani \etal \cite{Tulsiani2017CVPRa}, who proposed a method for 3D shape abstraction using a non-iterative approach which does not require supervision. Instead, they use a convolutional network architecture for predicting the shape and pose parameters of 3D cuboids as well as their probability of existence. They demonstrate that learning shape abstraction from data allows for obtaining consistent parses across different instances in an unsupervised fashion.

In this paper, we extend the model of Tulsiani \etal \cite{Tulsiani2017CVPRa} in the following directions. First, we utilize superquadrics, instead of cuboids, which leads to more accurate scene abstractions.
Second, we demonstrate that the bi-directional Chamfer distance 
is tractable and doesn't require reinforcement learning \cite{Williams1992ML} or specification of rewards \cite{Tulsiani2017CVPRa}.
In particular, we show that there exists an analytical closed-form solution which can be evaluated in linear time. %
This allows us to compute gradients \wrt the model parameters using standard error propagation \cite{Rumelhart1986NATURE} which facilitates learning. In addition, we add a new simple parsimony loss to favor configurations with a small number of primitives.

\subsection{Superquadrics}
\begin{figure}
    \centering
    \includegraphics[width=1.0\linewidth]{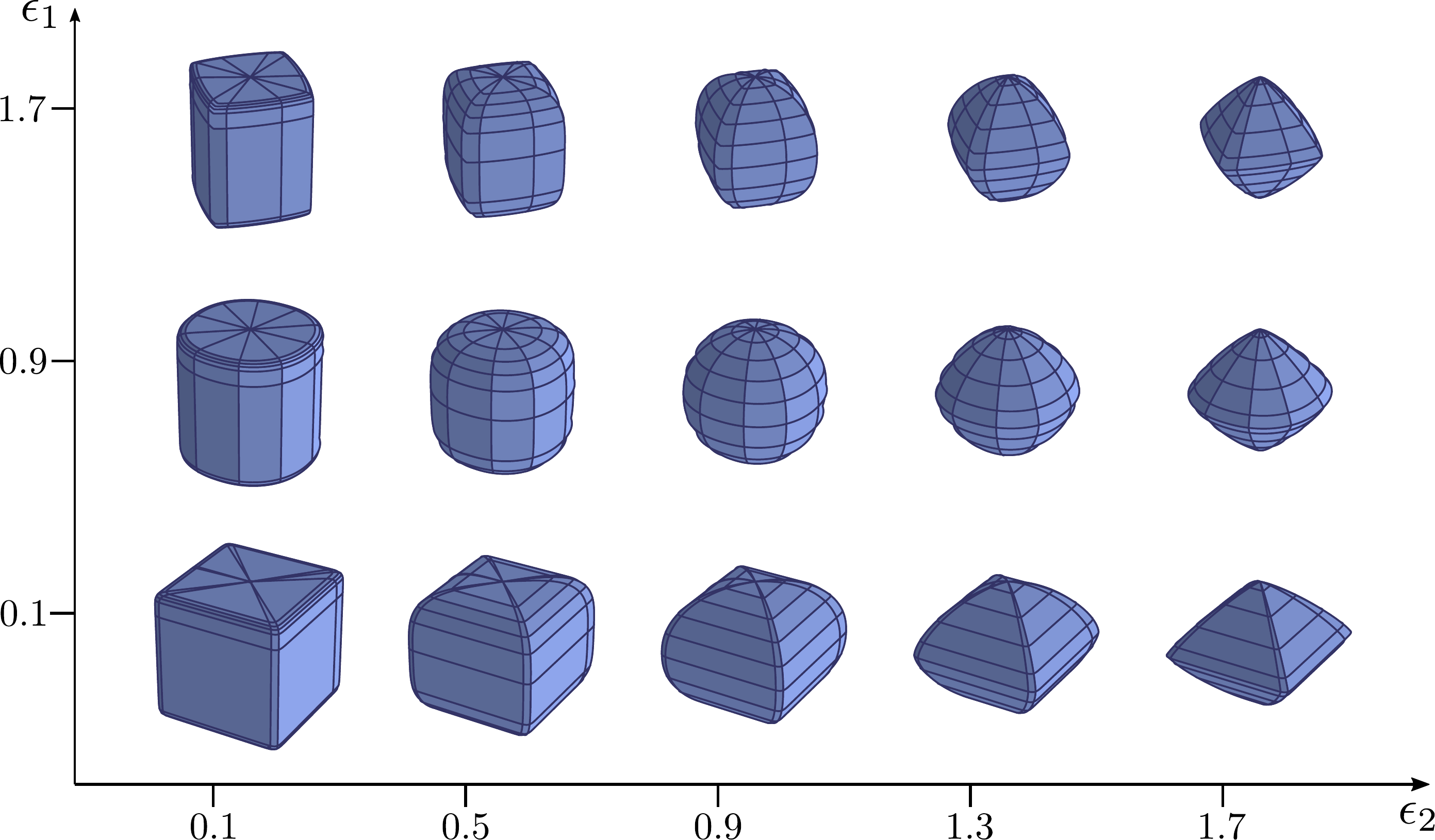}
    \caption{\textbf{Superquadrics Shape Vocabulary.} Due to their ability to model various shapes with little parameters, superquadrics are a natural choice for geometric primitives.}
    \label{fig:superquadrics}
    \vspace{-1.2em}
\end{figure}

Superquadrics are a parametric family of surfaces that can be used to describe cubes, cylinders, spheres, octahedra, ellipsoids \etc ~\cite{Barr1981CGA}. In contrast to geons~\cite{Biederman1986Human}, superquadric surfaces can be described using a fairly simple parameterization. In contrast to generalized cylinders~ \cite{Biederman1986Human}, superquadrics are able to represent a larger variety of shapes. 
See \figref{fig:superquadrics} for an illustration of the shape space.

In 1986, Pentland introduced superquadrics to the computer vision community~ \cite{Pentland1986AAAI}. Solina \etal \cite{Solina1990PAMI} formulated the task of fitting superquadrics to a point cloud as a least-squares minimization problem. Chevalier \etal \cite{Chevalier2003WSCG} followed a two-stage approach, where the point cloud is first partitioned into regions and then each region is fit with a superquadric. As a thorough survey on superquadrics is beyond the scope of this paper, we refer to \cite{Jaklic2000, Solina1994JCIT} for details.

In contrast to these classical works on superquadric fitting using non-linear least squares, we present the first approach to train a deep network to predict superquadrics directly from 2D or 3D inputs.
This allows our network to distill statistical dependencies \wrt the arrangement and geometry of the primitives from data, leading to semantically meaningful parts at inference time.
Towards this goal, we utilize a convolutional network that predicts superquadric poses and attributes, and develop a novel loss function that allow us to train this network efficiently from data.
Our model is able to directly learn superquadric surfaces from an unordered 3D point cloud without any supervision on the primitive parameters nor a 3D segmentation as input.

\section{Method}
\label{sec:method}

We now describe our model. We start by introducing the model parameters, followed by the loss functions and the superquadric parametrization we employ.

Given an input $\bI$ (\eg, image, volume, point cloud) and an oriented point cloud $\bX$ of the target object, our goal is to estimate the parameters $\theta$ of a neural network $\phi_{\theta}(\bI)$ that predicts a set of $M$ primitives that best describe the target object.
Every primitive is fully described by a set of parameters $\lambda_m$ that define its shape, size and its position and orientation in the 3D space.
For details about the parameterization of the superquadric representation, we refer the reader to \secref{subsec:sq_parametrization}.

Since not all objects and scenes require the same number of primitives, we enable our model to predict a variable number of primitives, hence allowing it to decide whether a primitive should be part of the assembled object or not. To achieve this, we follow \cite{Tulsiani2017CVPRa} and associate every primitive with a binary random variable $z_m \in \{0, 1\}$ which follows a Bernoulli distribution $p(z_m)=\gamma_m^{z_m}(1-\gamma_m)^{1-z_m}$ with parameter $\gamma_m$. The random variable $z_m$ indicates whether the $m^{th}$ primitive is part of the scene $(z_m = 1)$ or not $(z_m = 0)$. We refer to these variables as \emph{existence variables} and denote the set of all existence variables as $\bz = \{z_1, \dots, z_M\}$.
Our goal is to learn a neural network
\begin{equation}
\phi_{\theta}: \bI \mapsto \bP
\end{equation}
which maps an input $\bI$ to a primitive representation $\bP$ where
$\bP=\{(\lambda_m,\gamma_m)\}_{m=1}^M$ comprises the primitive parameters $\lambda_m$ and the existence probability $\gamma_m$ for $M$ primitives.
Note that $M$ is only an upper bound on the number of predicted primitives. The final primitive representation is obtained by sampling the existence of each primitive, $z_m \sim \text{Bernoulli}(\gamma_m)$.

One of the key challenges when training such models is related to the lack of direct supervision in the form of primitive annotations. However, despite the absence of supervision, one can still measure the discrepancy between the predicted object and the target object.
Towards this goal, we formulate a bi-directional reconstruction objective $\mathcal{L}_{D}(\bP , \bX)$
and incorporate a Minimum Description Length (MDL) prior $\mathcal{L}_{\gamma}(\bP)$, which favors parsimony, \ie a small number of primitives. Our overall loss function is given as:
\begin{equation}
\mathcal{L}(\bP , \bX) = \mathcal{L}_{D}(\bP , \bX) + \mathcal{L}_{\gamma}(\bP)
\end{equation}
We now describe both losses functions in detail.

\subsection{Reconstruction Loss}
\label{sec:reconstruction_loss}
The reconstruction loss measures the discrepancy between the predicted shape and the target shape.
While we experimented with the truncated bi-directional loss
of Tulsiani \etal \cite{Tulsiani2017CVPRa},
we empirically found that the standard Chamfer distance \cite{Fan2017CVPR} works better in practice and results in less local minima. An empirical analysis on this is provided in our supplementary material. Thus, we use the Chamfer distance in our experiments
\begin{equation}
\mathcal{L}_{D}(\bP , \bX) = \mathcal{L}_{P\rightarrow X}(\bP, \bX)  + \mathcal{L}_{X\rightarrow P}(\bX, \bP) 
\label{eq:loss_distance}
\end{equation}
where $\mathcal{L}_{P\rightarrow X}$ measures the distance from the predicted primitives $\bP$ to the point cloud $\bX$ and $\mathcal{L}_{X\rightarrow P}$ measures the distance from the point cloud $\bX$ to the primitives $\bP$.
We weight the two distance measures in \eqref{eq:loss_distance} with $1.2$ and $0.8$, respectively, which empirically led to good results.

\boldparagraph{Primitive-to-Pointcloud}
We represent the target point cloud as a set of 3D points $\bX = \{\bx_i \}_{i=1}^N$.
Similarly, we approximate the continuous surface of primitive $m$ by a set of points $\bY_m = \{\by_k^m \}_{k=1}^K$.
Details of our sampling strategy are provided in~\secref{subsec:implementation}.
This discretization allows us to express the distance between a superquadric and the target point cloud in a convenient form.
In particular, for each point on the primitive $\by_k^m$, we compute its closest point on the target point cloud $\bx_i$, and average this distance across all points in $\bY_m$ as follows: 
\begin{equation}
\mathcal{L}^m_{P\rightarrow X}(\bP, \bX) = \frac{1}{K} \sum_{k=1}^K ~\Delta^m_k
\label{eq:loss_1_m}
\end{equation}
where
\begin{equation}
\Delta^m_k = \min_{i=1,..,N} \, {\Vert\cT_m(\bx_i) - \by_k^m \Vert}_2
\label{eq:delta_k}
\end{equation}
denotes the minimal distance from the $k$'th point $\by_k^m$ on the $m$'th primitive to the target point cloud $\bX$.
Here, $\cT_m(\bx)=\bR(\lambda_m)\,\bx + \bt(\lambda_m)$ is a function that transforms a 3D point $\bx_i$ in world coordinates into the local coordinate system of the $m^{th}$ primitive.
Note that both $\mathbf{R}$ and $\mathbf{t}$ depend on $\lambda_m$ and are hence estimated by our network.
By taking the expectation \wrt the existence variables $\bz$ and assuming independence of the existence variables: $p(\bz) = \prod_m p(z_m)$, we obtain the joint loss over all primitives as
\begin{equation}
    \begin{aligned}
    \mathcal{L}_{P\rightarrow X}(\bP, \bX) &= \nE_{p(\bz)} \left[ \sum_{m=1}^M \mathcal{L}^m_{P\rightarrow X}(\bP, \bX) \right] \\ &= \sum_{m=1}^M \gamma_m \, \mathcal{L}^m_{P\rightarrow X}(\bP, \bX)
        \end{aligned}
    \label{eq:loss_1}
\end{equation}
Note that this loss encourages the predicted primitives to stay close to the target point cloud.

\boldparagraph{Pointcloud-to-Primitive}
While $\mathcal{L}_{P\rightarrow X}$ measures the distance from the primitives to the point cloud, $\mathcal{L}_{X\rightarrow P}$ measures the distance from the point cloud to the primitives to ensure that each observation is explained by at least one primitive. 
We start by defining $\Delta^m_i$ as the minimal distance from point $\bx_i$ to the surface of the $m$'th primitive: 
\begin{equation}
\Delta^m_i = \min_{k=1,..,K} {\Vert \cT_m(\bx_i) - \by_k^m \Vert}_2
\label{eq:delta_i}
\end{equation}
Note that in contrast to \eqref{eq:delta_k}, we minimize over the $K$ points from the estimated primitive.
Similarly to \eqref{eq:loss_1}, we take the expectation of $\Delta^m_i$ over $p(\bz)$. In contrast to \eqref{eq:loss_1}, we sum over each point in the target point cloud $\bX$ and retrieve the distance to the closest primitive $m$ that exists ($z_m=1$): 
\begin{equation}
    \begin{aligned}
    &\cL_{X\rightarrow P}(\bX,\bP) = \nE_{p(\bz)} \left[\sum_{\bx_i \in \bX} 
     ~\min_{\substack{m | z_m=1}} \, \Delta^m_i \right] 
    \end{aligned}
    \label{eq:loss_2}
\end{equation}
Note that na\"ive computation of Eq.~\ref{eq:loss_2} becomes very slow for a large number of primitives $M$ as it requires evaluating the quantity inside the expectation $2^M$ times.
In this work, we propose a novel approach to simplify this computation that results in a linear number of evaluations. Without loss of generality, let us assume that the $\Delta^m_i$'s are sorted in ascending order:
\begin{equation}
    \begin{aligned}
        \Delta^1_i \leq \Delta^2_i \leq \dots \leq \Delta^M_i
    \end{aligned}
    \label{eq:ordering}
\end{equation}
Assuming this ordering, we can state the following: if the first primitive exists, the first primitive will be the one closest to point $\bx_i$ of the target point, if the first primitive does not exist and the second does, then the second primitive is closest to point $\bx_i$ and so on and so forth. More formally, this property can be stated as follows: 
\begin{equation}
    \begin{aligned}
    &\min_{\substack{m | z_m=1}} \Delta^m_i = 
    &\begin{cases}
        \Delta^1_i , & \text{if}\ z_1=1 \\
        \Delta^2_i, & \text{if}\ z_1=0, z_2=1 \\
        \vdots \\
        \Delta^M_i, & \text{if}\ z_m=0, \dots, z_M=1 \\
        \end{cases}
    \end{aligned}
    \label{eq:minimum_distances}
\end{equation}
This allows us to simplify Eq.~\ref{eq:loss_2} as follows
\begin{equation}
    \cL_{X\rightarrow P}(\bX,\bP) = \sum_{\bx_i \in \bX} \sum_{m=1}^M 
   \Delta^m_i \, \gamma_m \prod_{\bar{m}=1}^{m-1}(1 - \gamma_{\bar{m}})
   \label{eq:loss_2_simplified}
\end{equation}
where $\gamma_{\bar{m}}$ is a shorthand notation which denotes the existence probability of a primitive closer than primitive $m$.
Note that this function requires only $M$, instead of $2^M$, evaluations of the function $\Delta^m_i$ which is one of the main results of this paper.
For a detailed derivation of \eqref{eq:loss_2_simplified}, we refer the reader to the supplementary material.

\subsection{Parsimony Loss}
\label{sec:mdl_loss}

Despite the bidirectional loss formulation above, our model suffers from the trivial solution $\cL_{D}(\bP,\bX)=0$ which is attained for $\gamma_1=\cdots=\gamma_m=0$. Moreover, multiple primitives with identical parameters yield the same loss function as a single primitive by dispersing their existence probability. We thus introduce a regularizer loss on the existence probabilities $\gamma$ which alleviates both problems:
\begin{equation}
\mathcal{L}_{\gamma}(\bP) = \max\left(\alpha-\alpha\sum_{m=1}^M \gamma_m, 0\right)
+ \beta \sqrt{\sum_{m=1}^M \gamma_m}
\label{eq:loss_gamma}
\end{equation}
The first term of \eqref{eq:loss_gamma} makes sure that the aggregate existence probability over all primitives is at least one (\ie, we expect at least one primitive to be present) and the second term enforces a parsimonious scene parse by exploiting a loss function sub-linear in $\sum_m \gamma_m$ which encourages sparsity. $\alpha$ and $\beta$ are weighting factors which are set to $1.0$ and $10^{-3}$ respectively.

\subsection{Superquadric Parametrization}
\label{subsec:sq_parametrization}

Having specified our network and the loss function, we now provide details about the superquadric representation and its parameterization $\lambda$. Note that, in this section, we omit the primitive index $m$ for clarity. Superquadrics define a family of parametric surfaces that can be fully described by a set of $11$ parameters~\cite{Barr1981CGA}. The explicit superquadric equation defines the surface vector $\br$ as
\begin{equation}
    \mathbf{r}(\eta, \omega) =
    \begin{bmatrix}
        \alpha_{1}\cos^{\epsilon_{1}}\eta \cos^{\epsilon_{2}}\omega \\
        \alpha_{2}\cos^{\epsilon_{1}}\eta \sin^{\epsilon_{2}}\omega \\
        \alpha_{3}\sin^{\epsilon_{1}}\eta
    \end{bmatrix}
    \quad 
    \begin{aligned}
        -\pi/2 &\leq \eta \leq \pi/2\\
        -\pi &\leq \omega \leq \pi
    \end{aligned}
    \label{eq:parametric_eq}
\end{equation}
where $\mathbf{\alpha} = [\alpha_{1}, \alpha_{2}, \alpha_{3}]$ determine the size and $\mathbf{\epsilon} = [\epsilon_{1}, \epsilon_{2}]$ determine the global shape of the superquadric, see supplementary material for examples.
Following common practice \cite{Vaskevicius2017PAMI}, we bound the values $\epsilon_{1}$ and $\epsilon_{2}$ to the range $[0.1, 1.9]$ so as to prevent non-convex shapes which are less likely to occur in practice. Eq.~\ref{eq:parametric_eq} produces a superquadric in a canonical pose. In order to  allow any position and orientation, we augment the primitive parameter $\lambda$ with an additional rigid body motion represented by a translation vector $\mathbf{t} = [t_{x}, t_{y}, t_{z}]$ and a quaternion $\mathbf{q} = [q_{0}, q_{1}, q_{2}, q_{3}]$ which determine the coordinate system transformation $\cT(\bx)=\bR(\lambda)\,\bx + \bt(\lambda)$ above. 

\begin{figure}
	\centering
	\includegraphics[width=\linewidth]{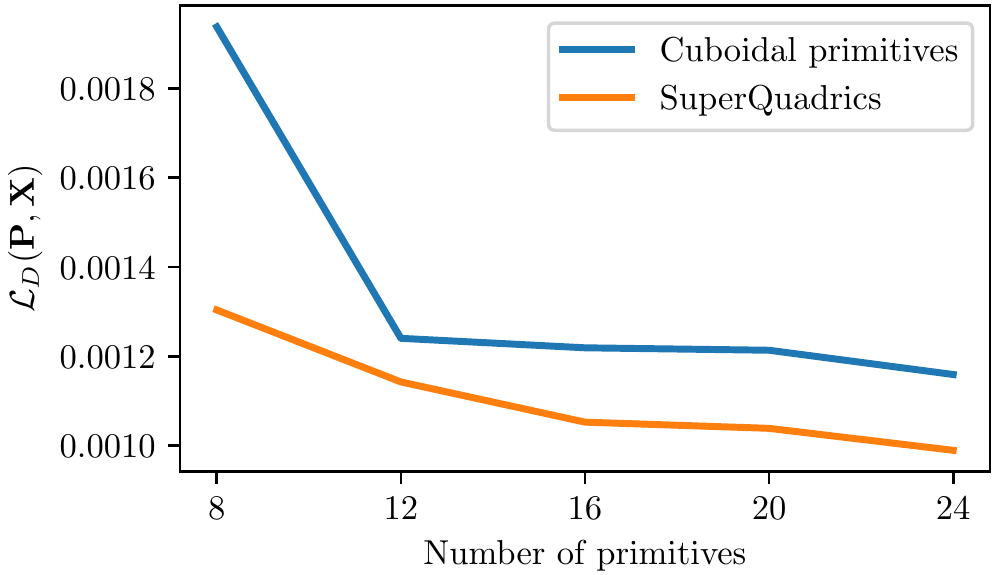}
	\caption{{\bf{Reconstruction Loss \wrt \#Primitives.}} We illustrate the reconstruction loss on the test set of the ShapeNet \textit{animal} category for a different number of primitives. Superquadrics ({\color{orange}{orange}}) consistently outperform cuboid primitives ({\color{blue}{blue}}) due to their diverse shape vocabulary that allows them to better capture fine details of the input shapes.}
	\label{fig:reconstruction_loss}
    \vspace{-1.2em}
\end{figure}

\subsection{Implementation}
\label{subsec:implementation}

Our network architecture comprises an encoder and a set of linear layers followed by non-linearities that independently predict the pose, shape and size of the superquadric surfaces.
The encoder architecture is chosen based on the input type (\eg image, voxelized input, \etc). In our experiments, for a binary occupancy grid as input, our encoder consists of five blocks of 3D convolution layers, followed by batch normalization and Leaky ReLU non-linearities.
The result is passed to five independent heads that regress translation $\bf{t}$, rotation $\bf{q}$, size $\bf{\alpha}$, shape $\bf{\epsilon}$ and probability of existence $\bf{\gamma}$ for each primitive.
Additional details about our network architecture as well as results using an image-based encoder are provided in the supplementary material.

For evaluating our loss \eqref{eq:loss_distance}, we sample points on the superquadric surface.
To achieve a uniform point distribution, we sample $\eta$ and $\omega$ as proposed in \cite{Pilu1995BMVC}.
During training, we uniformly sample $1000$ points, from the surface of the target object, as well as $200$ points from the surface of every superquadric. Note that sampling points on the surface of the objects results in a stochastic approximator of the expected loss. The variance of this approximator is inversely proportional to the number of sampled points. We experimentally observe that our model is not sensitive to the number of sampled points.
For optimization, we use ADAM \cite{Kingma2015ICLR} with learning rate $0.001$ and a batch size of $32$ for $40$k iterations. To further increase parsimony, we then fix all parameters except $\bf{\gamma}$ for additional $5$k iterations. This step removes remaining overlapping primitives as also observed in \cite{Tulsiani2017CVPRa}.

\section{Experimental Evaluation}
\label{sec:results}

\begin{figure}
	\centering
	\includegraphics[width=\linewidth]{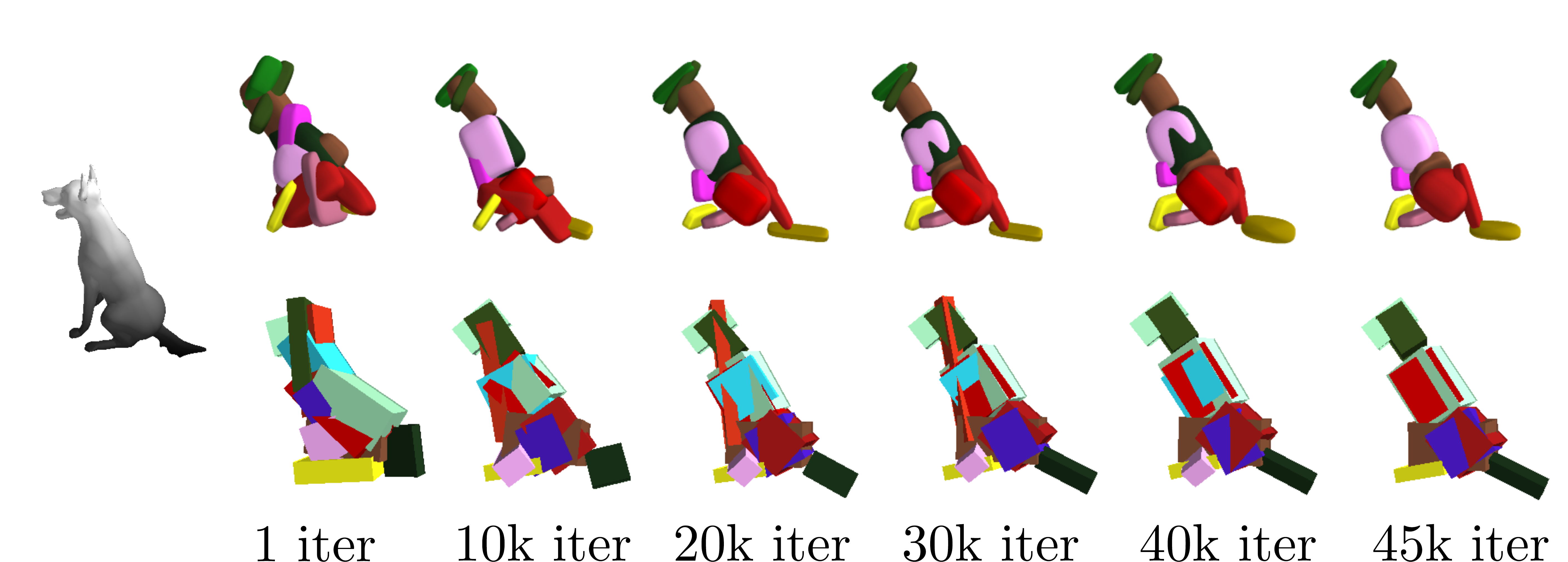}
    \vspace{-0.2em}
	\caption{{\bf{Training Evolution.}} We visualize the qualitative evolution of superquadrics (top) and cuboids (bottom) during training.
		Superquadrics converge faster to more accurate representations, whereas cuboids cannot capture details such as the open mouth of the dog, even after convergence.}
	\label{fig:training_evolution}
	\vspace{-1.2em}
\end{figure}

In this section, we present a set of experiments to evaluate the performance of our network in terms of parsing an input 3D shape into a set of superquadric surfaces.

\begin{figure}[t!]
	\centering
    \includegraphics[width=\linewidth, height=1.1\linewidth]{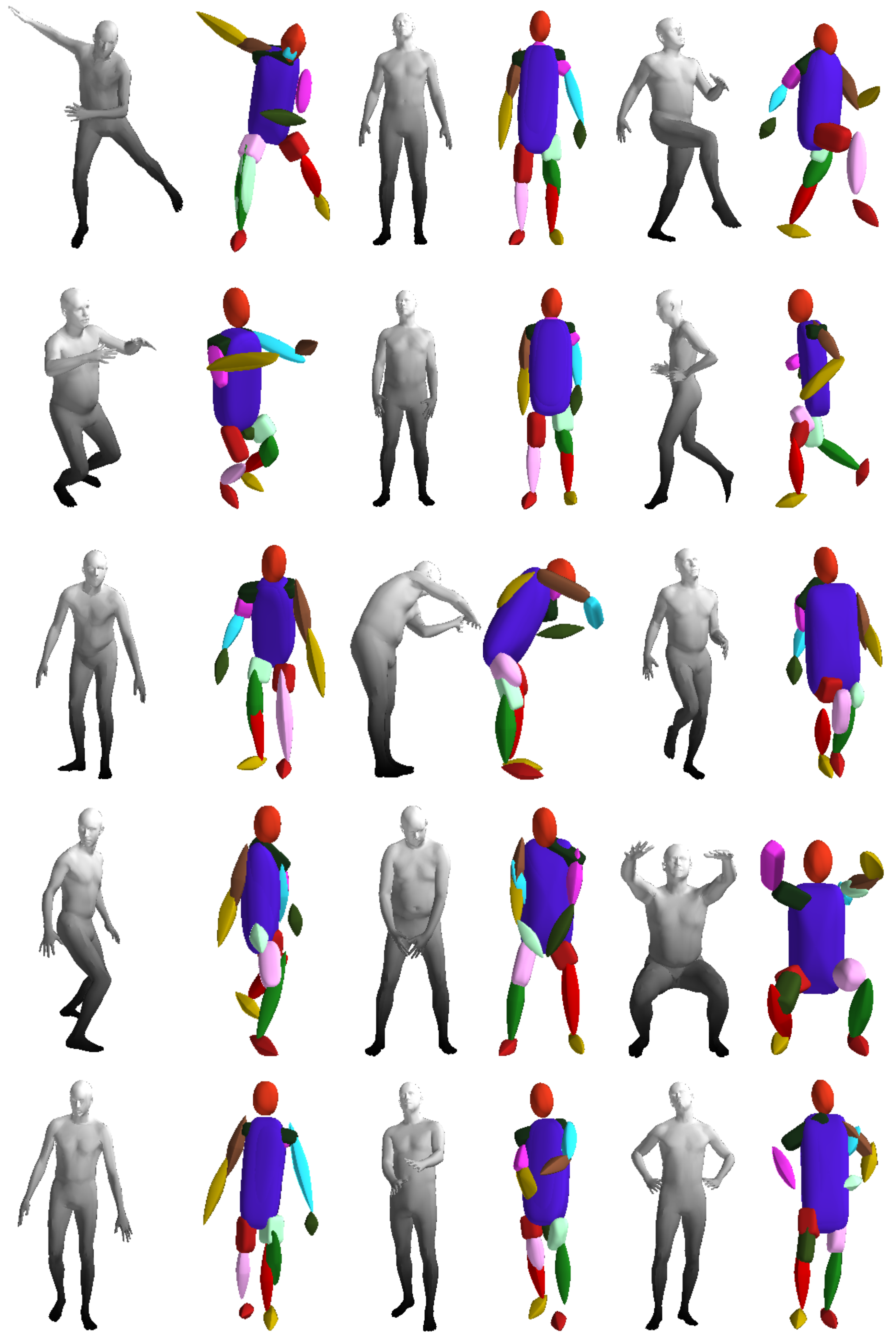}
    \caption{\textbf{Qualitative Results on SURREAL.} Our network learns semantic mappings of body parts across different body shapes and articulations. For instance, the network uses the same primitive for the left forearm across instances.}
    \label{fig:surreal}
    \vspace{-1.2em}
\end{figure}

\begin{figure*}[t!]
	\centering
	\begin{subfigure}[b]{0.10\textwidth}
		\centering
		\includegraphics[width=0.7\textwidth]{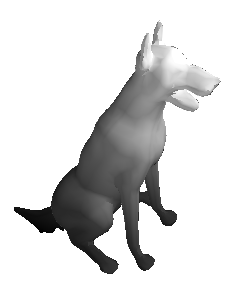}
	\end{subfigure}
	\hfill
	\begin{subfigure}[b]{0.10\textwidth}
		\centering
		\includegraphics[width=0.9\textwidth]{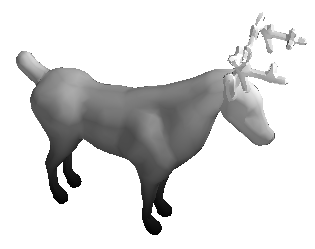}
	\end{subfigure}
	\hfill
	\begin{subfigure}[b]{0.10\textwidth}
		\centering
		\includegraphics[width=\textwidth]{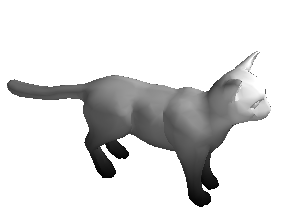}
	\end{subfigure}
	\hfill
	\begin{subfigure}[b]{0.10\textwidth}
		\centering
		\includegraphics[width=0.55\textwidth]{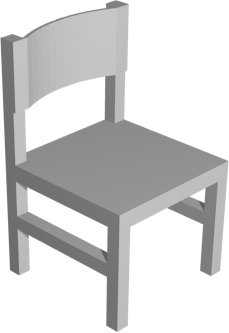}
	\end{subfigure}
	\hfill
	\begin{subfigure}[b]{0.10\textwidth}
		\centering
		\includegraphics[width=\textwidth]{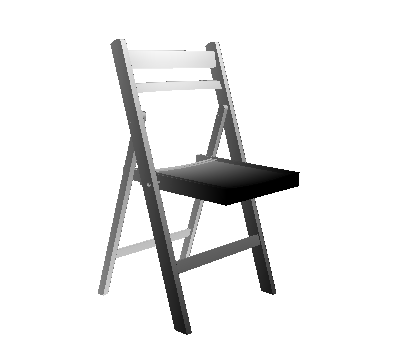}
	\end{subfigure}
	\hfill
	\begin{subfigure}[b]{0.10\textwidth}
		\centering
		\includegraphics[width=0.55\textwidth]{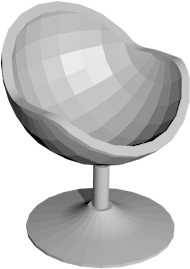}
	\end{subfigure}
	\hfill
	\begin{subfigure}[b]{0.10\textwidth}
		\centering
		\includegraphics[width=\textwidth]{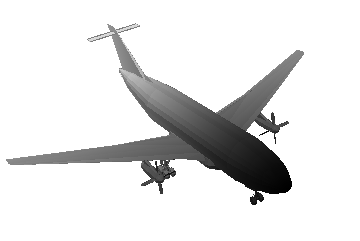}
	\end{subfigure}
	\hfill
	\begin{subfigure}[b]{0.10\textwidth}
		\centering
		\includegraphics[width=\textwidth]{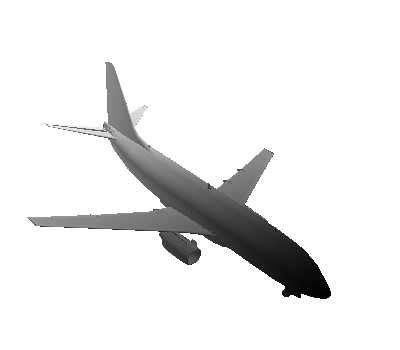}
	\end{subfigure}
	\hfill
	\begin{subfigure}[b]{0.10\textwidth}
		\centering
		\includegraphics[width=\textwidth]{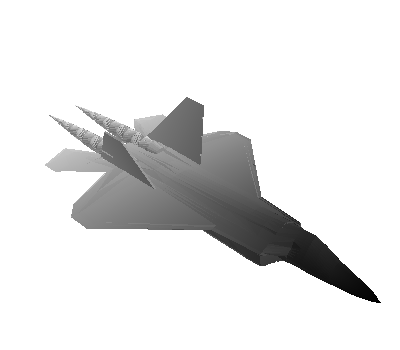}
	\end{subfigure}
	\vskip\baselineskip
    \vspace{-1.5em}
	\begin{subfigure}[b]{0.10\textwidth}
		\centering
		\includegraphics[width=\textwidth]{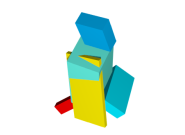}
	\end{subfigure}
	\hfill
	\begin{subfigure}[b]{0.10\textwidth}
		\centering
		\includegraphics[width=\textwidth]{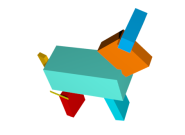}
	\end{subfigure}
	\hfill
	\begin{subfigure}[b]{0.10\textwidth}
		\centering
		\includegraphics[width=\textwidth]{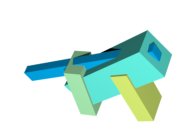}
	\end{subfigure}
	\hfill
	\begin{subfigure}[b]{0.10\textwidth}
		\centering
		\includegraphics[width=0.55\textwidth]{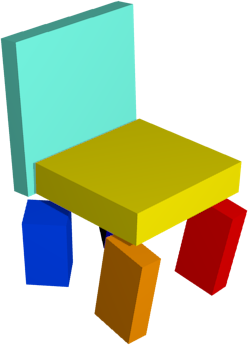}
	\end{subfigure}
	\hfill
	\begin{subfigure}[b]{0.10\textwidth}
		\centering
		\includegraphics[width=0.6\textwidth]{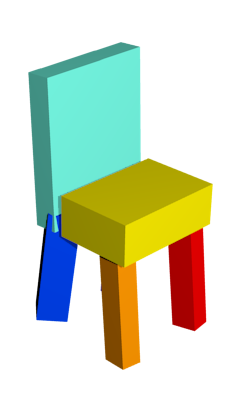}
	\end{subfigure}
	\hfill
	\begin{subfigure}[b]{0.10\textwidth}
		\centering
		\includegraphics[width=0.95\textwidth]{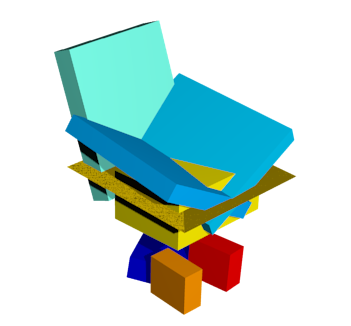}
	\end{subfigure}
	\hfill
	\begin{subfigure}[b]{0.10\textwidth}
		\centering
		\includegraphics[width=\textwidth]{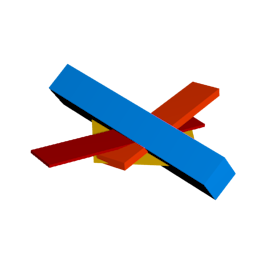}
	\end{subfigure}
	\hfill
	\begin{subfigure}[b]{0.10\textwidth}
		\centering
		\includegraphics[width=\textwidth]{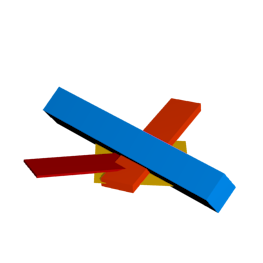}
	\end{subfigure}
	\hfill
	\begin{subfigure}[b]{0.099\textwidth}
		\centering
		\includegraphics[width=\textwidth]{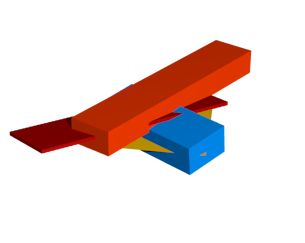}
	\end{subfigure}
	\vskip\baselineskip
    \vspace{-1.5em}
	\begin{subfigure}[b]{0.099\textwidth}
		\centering
		\includegraphics[width=0.8\textwidth]{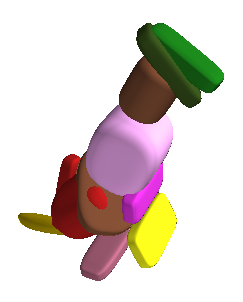}
	\end{subfigure}
	\hfill
	\begin{subfigure}[b]{0.10\textwidth}
		\centering
		\includegraphics[width=\textwidth]{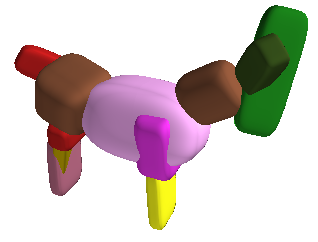}
	\end{subfigure}
	\hfill
	\begin{subfigure}[b]{0.10\textwidth}
		\centering
		\includegraphics[width=\textwidth]{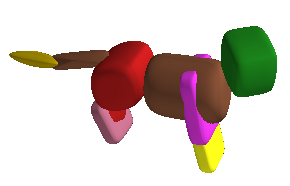}
	\end{subfigure}
	\hfill
	\begin{subfigure}[b]{0.10\textwidth}
		\centering
		\includegraphics[width=\textwidth]{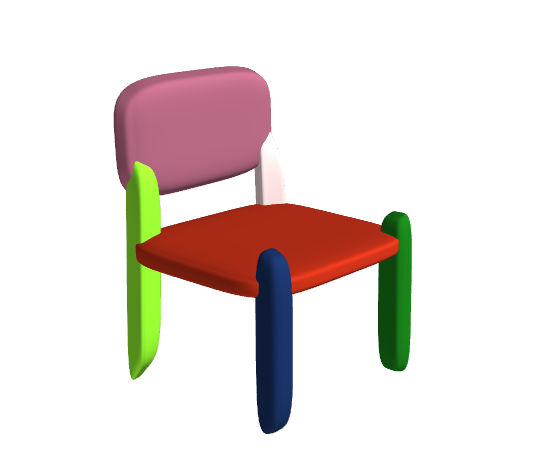}
	\end{subfigure}
	\hfill
	\begin{subfigure}[b]{0.10\textwidth}
		\centering
		\includegraphics[width=\textwidth]{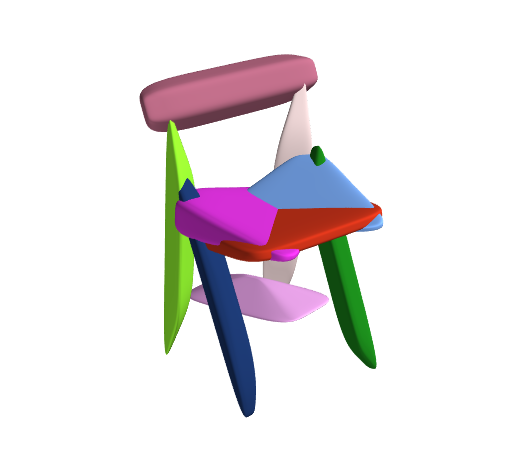}
	\end{subfigure}
	\hfill
	\begin{subfigure}[b]{0.10\textwidth}
		\centering
		\includegraphics[width=\textwidth]{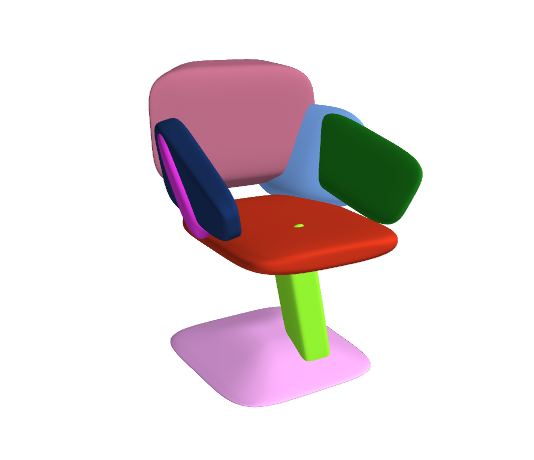}
	\end{subfigure}
	\hfill
	\begin{subfigure}[b]{0.10\textwidth}
		\centering
		\includegraphics[width=\textwidth]{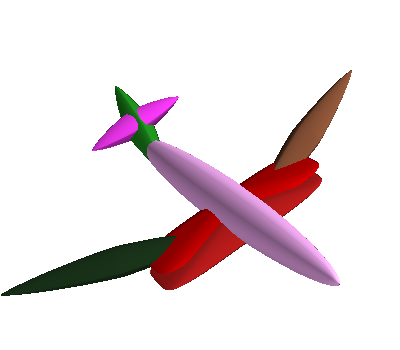}
	\end{subfigure}
	\hfill
	\begin{subfigure}[b]{0.10\textwidth}
		\centering
		\includegraphics[width=\textwidth]{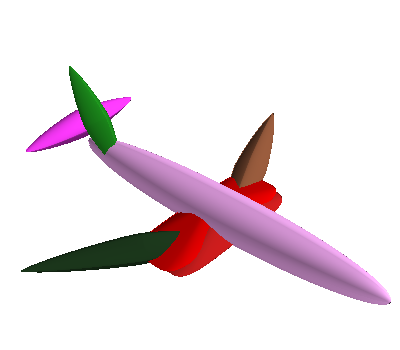}
	\end{subfigure}
	\hfill
	\begin{subfigure}[b]{0.10\textwidth}
		\centering
		\includegraphics[width=\textwidth]{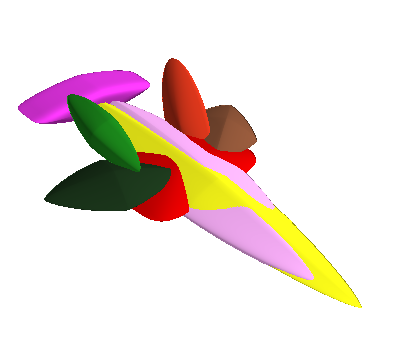}
	\end{subfigure}
	\caption{{\bf Qualitative Results on ShapeNet.} We visualize predictions for the object categories \textit{animals}, \textit{aeroplane} and \textit{chairs} from the ShapeNet dataset.
		The top row illustrates the ground-truth meshes from every object. The middle row depicts the corresponding predictions using the cuboidal primitives estimated by \cite{Tulsiani2017CVPRa}. The bottom row shows the corresponding predictions using our learned superquadric surfaces. Similarly to \cite{Tulsiani2017CVPRa}, we observe that the predicted primitive representations are consistent across instances. For example, the primitive depicted in green describes the right wing of the aeroplane, while for the animals class, the yellow primitive describes the front legs of the animal.}
	\label{fig:shapenet_results}
    \vspace{-1.0em}
\end{figure*}

\begin{figure}[t!]
	\centering
	\includegraphics[width=\linewidth]{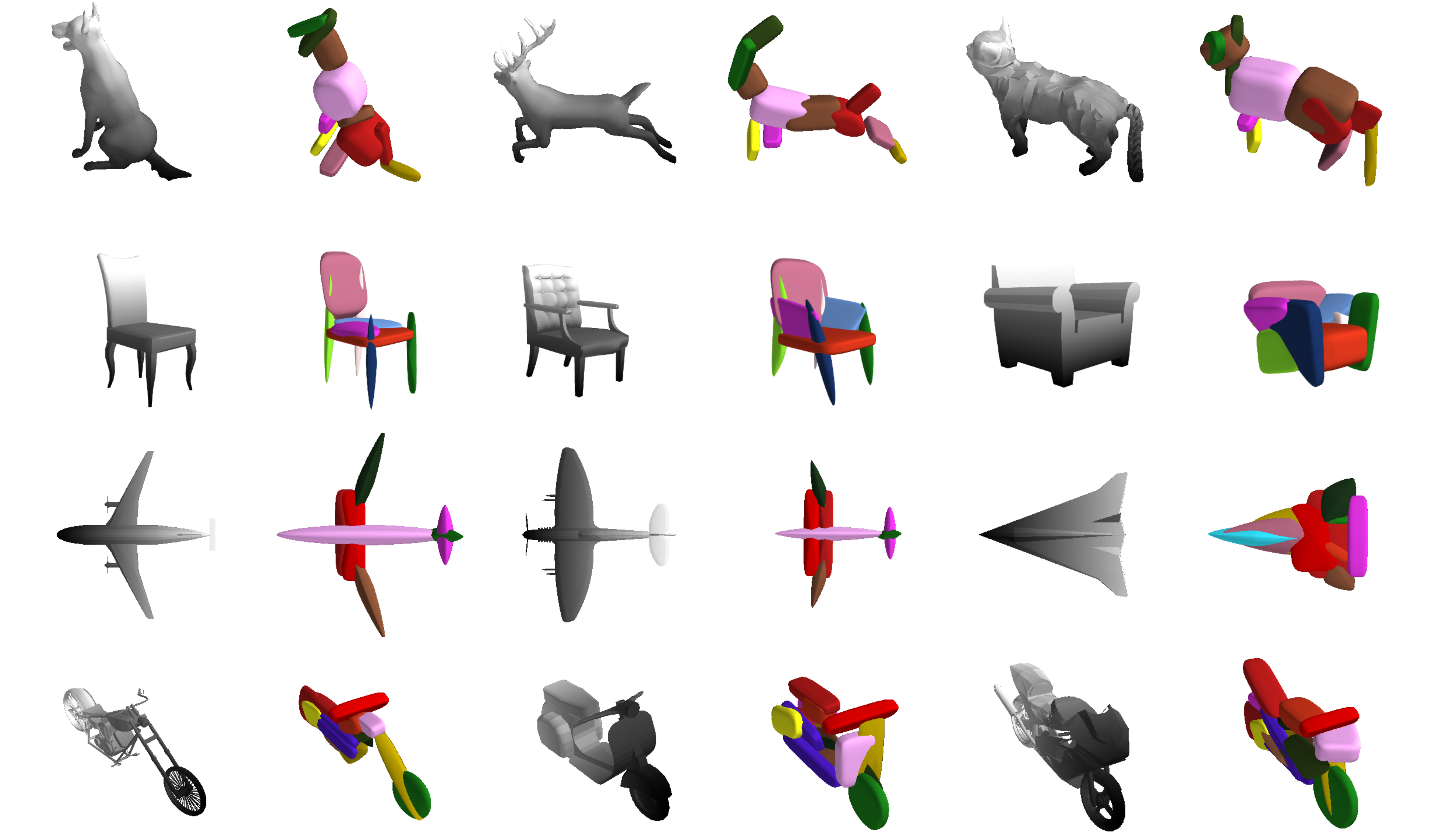}
	\caption{\textbf{Attention to Details.} Superquadrics allow for modeling fine details such as the tails and ears of animals as well as the wings and the body of the airplanes and wheels of the motorbikes which are hard to capture using cuboids.}
	\label{fig:attention_in_details}
	\vspace{-1.2em}
\end{figure}%

\boldparagraph{Datasets}
We provide results on two 3D datasets.
First, we use the \textit{aeroplane}, \textit{chair} and \textit{animals} categories from ShapeNet~\cite{Chang2015ARXIV}. Following \cite{Tulsiani2017CVPRa}, we train one model per object category using a voxelized binary occupancy grid of size $32\times 32 \times 32$ as input.
Second, we use the SURREAL dataset from Varol \etal \cite{Varol2017CVPR} which comprises humans in various poses (\eg, standing, walking, sitting). Using the SMPL model~\cite{Loper2015SIGGRAPH},
we rendered $5000$ meshes, from which $4500$ are used for training and $500$ for testing. 
For additional qualitative results on both datasets, we refer the reader to our supplementary material.

\boldparagraph{Baselines}
Most related to ours is the cuboid parsing approach of Tulsiani \etal~\cite{Tulsiani2017CVPRa}.
Other approaches to cuboid-based scene parsing \cite{Niu2018CVPR,Zou2017ICCV} require ground-truth shape annotations and thus cannot be fairly compared to unsupervised techniques.
We thus compare to Tulsiani \etal~\cite{Tulsiani2017CVPRa}, using their publicly available code\footnote{\url{https://github.com/shubhtuls/volumetricPrimitives}}.

\subsection{Superquadrics vs. Cuboids}
\label{sec:sq_vs_cubes}

We first compare the modeling accuracy of superquadric surfaces \wrt cuboidal shapes which have been extensively used in related work~\cite{Tulsiani2017CVPRa, Zou2017ICCV, Niu2018CVPR}.
Towards this goal, we fit \textit{animal} shapes from ShapeNet by optimizing the distance loss function in \eqref{eq:loss_distance} while varying the maximum number of allowed primitives $M$.
To ensure a fair comparison, we use the proposed model for both cases.
Note that this is trivially possible as cuboids are a special case of superquadrics.
To minimize the effects of network initialization and local minima in the optimization, we repeat the experiment three times with random initializations and visualize the average loss in \figref{fig:reconstruction_loss}. The results show that for any given number of primitives, superquadrics consistently achieve a lower loss, and hence higher modeling fidelity. We further visualize the qualitative evolution of the network during training in \figref{fig:training_evolution}. This figure demonstrates that compared to cuboids, superquadrics better model the object shape, and more importantly that the network is able to converge faster.

\subsection{Results on ShapeNet}
\label{sec:eval_shapenet}

We evaluate the quality of the predicted primitives using our reconstruction loss from \eqref{eq:loss_distance} on the ShapeNet dataset and compare to the cuboidal primitives as estimated by Tulsiani \etal \cite{Tulsiani2017CVPRa}. We associate every primitive with a unique color, thus primitives illustrated with the same color correspond to the same object part.
For both approaches we set the maximal number of primitives to $M=20$.
From \figref{fig:shapenet_results}, we observe that our predictions consistently capture both the structure as well as fine details (\eg, body, tails, head), whereas the corresponding cuboidal primitives from \cite{Tulsiani2017CVPRa} focus mainly on the structure of the predicted object.

\figref{fig:attention_in_details} shows additional results in which our model successfully predicts animals, airplanes and also more complicated motorbike parts.
For instance, we observe that our model is able to capture the open mouth of the dog using two superquadrics as shown in \figref{fig:attention_in_details} (left-most animal in third row). In addition, we notice that our model dynamically allocates a variable number of primitives depending on the complexity of the input shape. For example, the left-most airplane in \figref{fig:shapenet_results}, is modelled with $6$ primitives whereas the jetfighter (right-most) that has a more complicated shape is modelled with $9$ primitives. This can also be observed for the animal category, where our model chooses a single primitive for the body of the cat (rightmost animal in \figref{fig:shapenet_results}) while for all the rest it uses two.
We remark that our expressive shape abstractions allow for differentiating between different types of objects such as scooter/chopper/racebike or airliner/fighter by truthfully capturing the shape of the individual object parts.

\begin{figure}[t!]
	\centering
	\includegraphics[width=\linewidth]{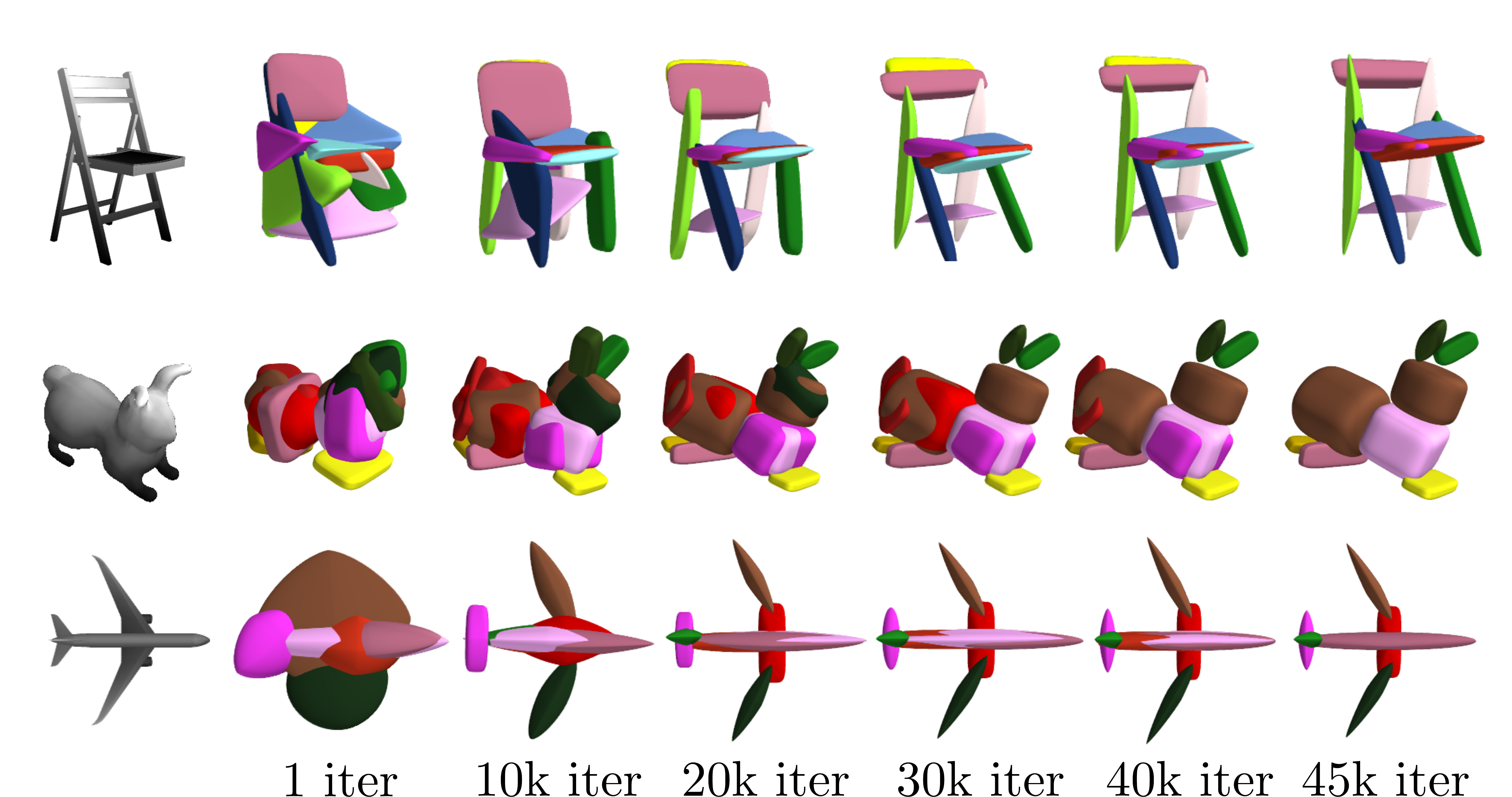}
	\caption{{\bf{Evolution of Primitives.}} We illustrate the evolution of superquadrics during training. Note how our model first focuses on the overall structure of the object and starts to attend to finer details at later iterations.}
	\label{fig:primitives_evolution}
    \vspace{-1.5em}
\end{figure}

\figref{fig:primitives_evolution} visualizes the training evolution of the predicted superquadrics for three object categories.
While initially, the model focuses on the overall structure of the object using mostly blob-shaped superquadrics ($\epsilon_1$ and $\epsilon_2$ close to $1.0$), as training progresses it starts attending to details.
After convergence, the predicted superquadrics closely match the shape of the corresponding (unknown) object parts.

\begin{figure*}[t!]
	\centering
    \begin{subfigure}[t]{0.5\textwidth}
        \centering
	    \includegraphics[width=\linewidth]{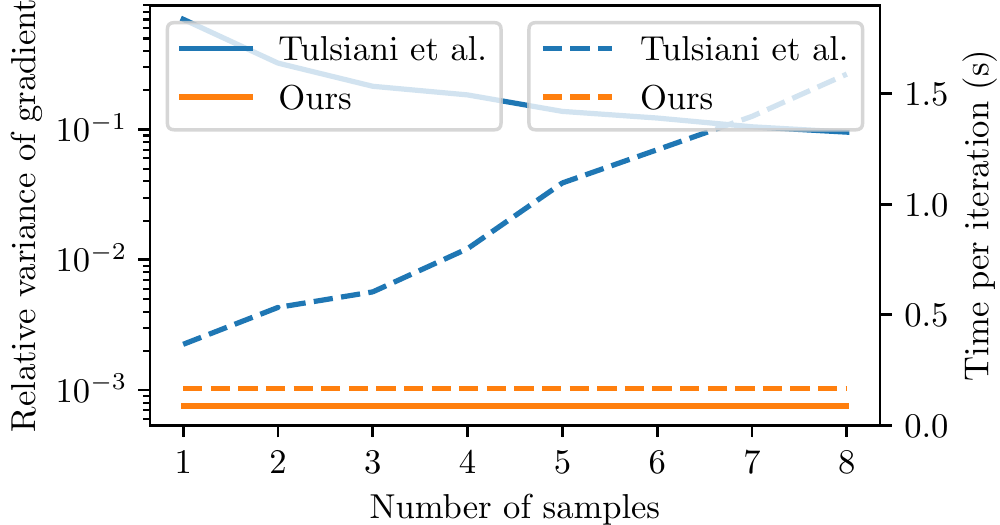}
	    \caption{\textbf{Gradient Variance and Iteration Time.}}
	    \label{fig:gradient_time}
    \end{subfigure}%
    ~
    \begin{subfigure}[t]{0.5\textwidth}
	    \centering
	    \includegraphics[width=\linewidth]{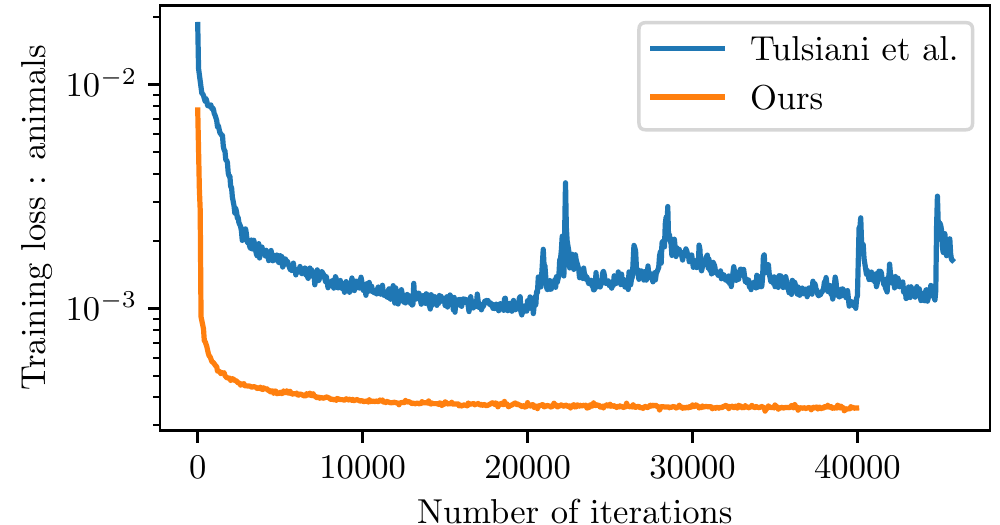}
	    \caption{\textbf{Evolution of Training Loss.}}
	    \label{fig:evolution_of_training_loss}
    \end{subfigure}
    \vspace{-0.3em}
	    \caption{\figref{fig:gradient_time} depicts the variance of the gradient estimates for $\gamma$ over 300 iterations (solid) as well as the computation time per iteration (dashed) for \cite{Tulsiani2017CVPRa} ({\color{blue}{blue}}) and our method ({\color{orange}{orange}}). Our analytical loss function provides gradients with orders of magnitude less variance while at the same time decreasing runtime. In \figref{fig:evolution_of_training_loss}, we compare the training loss evolution of \cite{Tulsiani2017CVPRa} ({\color{blue}{blue}}) to ours ({\color{orange}{orange}}). The sampling based approach of \cite{Tulsiani2017CVPRa} leads to large oscillations while ours converges smoothly.
}
        \vspace{-1.2em}
\end{figure*}%

\subsection{Results on SURREAL}

In addition to ShapeNet, we also demonstrate results on the SURREAL human body dataset in \figref{fig:surreal}. The benefits of superquadrics over simpler shape abstractions are accentuated in this dataset due to the complicated shapes of the human body.
Note that our model successfully captures details that require modeling beyond cuboids: For instance, our model predicts pointy octahedral shapes for the feet, ellipsoid shapes for the head and a flattened elongated superellipsoid for the main body without any supervision on the primitive parameters.
Another interesting aspect of our model is the consistency of the predicted primitives, \ie, the same primitives (highlighted with the same color) consistently represent feet, legs, arms \etc across different poses.
For more complicated poses, correspondences are sometimes mirrored. We speculate that this behavior is caused by symmetries of the human body.

\subsection{Analytical Loss Formulation}

In this section, we compare the evolution of our training loss in Equation \eqref{eq:loss_distance} to the evolution of the training loss proposed by Tulsiani \etal \cite{Tulsiani2017CVPRa} in \figref{fig:evolution_of_training_loss}.
While it is important to mention that the absolute values are not comparable due to the slightly different loss formulations, we observe that our loss converges faster with less oscillations. Note that at iteration 20k, \cite{Tulsiani2017CVPRa} starts updating the existence probabilities using reinforcement learning \cite{Williams1992ML} which further increases oscillations.
In contrast, our loss function decays smoothly and does not require sampling-based gradient estimation.

To further analyze the advantages of our analytical loss formulation, we calculate the variance of the gradient estimates for the existence probabilities $\gamma$ over 300 training iterations. \figref{fig:gradient_time} compares the variance of the gradients of \cite{Tulsiani2017CVPRa} to the variance of the gradients of the proposed analytical loss (solid lines). Note that the variance in our gradient is orders of magnitude lower compared to \cite{Tulsiani2017CVPRa} as we do not require sampling \cite{Williams1992ML} for approximating the gradients. Simultaneously, we obtain a lower runtime per iteration (dashed lines).
While using more samples lowers the variance of gradients approximated using Monte Carlo estimation \cite{Tulsiani2017CVPRa}, runtime per iteration increases linearly with the number of samples. In contrast, our method does not require sampling and outperforms \cite{Tulsiani2017CVPRa} in terms of runtime even for the case of gradient estimates based on a single sample.
We remark that in both cases the runtime is computed for the entire iteration, considering both, the forward and the backward pass.
A quantitative comparison is provided in \tabref{tab:quantitative_shapenet}. Note that in contrast to \cite{Tulsiani2017CVPRa} we optimize for the Chamfer distance.
 \begin{table}
    \centering
    \resizebox{\columnwidth}{!}{
     \begin{tabular}{@{}l cccccc@{}}
         & \multicolumn{3}{c}{Chamfer Distance} & \multicolumn{3}{c}{Volumetric IoU}\\
         \toprule
          & Chairs & Aeroplanes & Animals & Chairs & Aeroplanes & Animals \\
          \midrule
          \cite{Tulsiani2017CVPRa} & 0.0121& 0.0153& 0.0110 & 0.1288 & 0.0650& 0.3339\\
          Ours & \textbf{0.0006}& \textbf{0.0003} & \textbf{0.0003} & \textbf{0.1408}& \textbf{0.1808} & \textbf{0.7506}\\
         \bottomrule
     \end{tabular}}
     \caption{{\bf Quantitative evaluation.} We report the mean Chamfer distance (smaller is better) and the mean Volumetric IoU (larger is better) for our model compared to \cite{Tulsiani2017CVPRa}.}
     \label{tab:quantitative_shapenet}
     \vspace{-1.4em}
 \end{table}

\section{Conclusion and Future Work}
\label{sec:conclusion}

We propose the first learning-based approach for parsing 3D objects into consistent superquadric representations. Our model successfully captures both the structure as well as the details of the target objects by accurately learning to predict superquadrics in an unsupervised fashion from data. %

In future work, we plan to extend our model by including parameters for global deformations such as tapering and bending. We anticipate that this will significantly benefit the fitting process as the available shape vocabulary will be further increased. Finally, we also plan to extend our model to large-scale scenes. We believe that developing novel hierarchical strategies as in \cite{Zou2017ICCV} is key for unsupervised 3D scene parsing at room-, building- or even city-level scales.

\section*{Acknowledgments}
We thank Michael Black for early discussions on superquadrics. This research was supported by the Max Planck ETH Center for Learning Systems. 

{\small
	\bibliographystyle{ieee}
	\bibliography{bibliography_long,bibliography,bibliography_custom}

\begin{thebibliography}{10}\itemsep=-1pt

\bibitem{Barr1981CGA}
Alan~H Barr.
\newblock Superquadrics and angle-preserving transformations.
\newblock {\em IEEE Computer Graphics and Applications (CGA)}, 1981.

\bibitem{Biederman1986Human}
Irving Biederman.
\newblock Human image understanding: Recent research and a theory.
\newblock {\em Computer Vision, Graphics, and Image Processing}, 1986.

\bibitem{Biederman1987PsychologicalReview}
Irving Biederman.
\newblock Recognition-by-components: a theory of human image understanding.
\newblock {\em Psychological Review}, 94(2):115, 1987.

\bibitem{Binford1971Visual}
I Binford.
\newblock Visual perception by computer.
\newblock In {\em IEEE Conference of Systems and Control}, 1971.

\bibitem{Chang2015ARXIV}
Angel~X. Chang, Thomas~A. Funkhouser, Leonidas~J. Guibas, Pat Hanrahan,
  Qi{-}Xing Huang, Zimo Li, Silvio Savarese, Manolis Savva, Shuran Song, Hao
  Su, Jianxiong Xiao, Li Yi, and Fisher Yu.
\newblock Shapenet: An information-rich 3d model repository.
\newblock {\em arXiv.org}, 1512.03012, 2015.

\bibitem{Chevalier2003WSCG}
Laurent Chevalier, Fabrice Jaillet, and Atilla Baskurt.
\newblock Segmentation and superquadric modeling of 3d objects.
\newblock In {\em International Conference in Central Europe on Computer
  Graphics, Visualization and Computer Vision (WSCG)}, 2003.

\bibitem{Choy2016ECCV}
Christopher~Bongsoo Choy, Danfei Xu, JunYoung Gwak, Kevin Chen, and Silvio
  Savarese.
\newblock 3d-r2n2: {A} unified approach for single and multi-view 3d object
  reconstruction.
\newblock In {\em Proc. of the European Conf. on Computer Vision (ECCV)}, 2016.

\bibitem{Roberts1963PhD}
Peter Elias and Lawrence~G Roberts.
\newblock {\em Machine perception of three-dimensional solids}.
\newblock PhD thesis, Massachusetts Institute of Technology, 1963.

\bibitem{Ellis2018NIPS}
Kevin Ellis, Daniel Ritchie, Armando Solar{-}Lezama, and Joshua~B. Tenenbaum.
\newblock Learning to infer graphics programs from hand-drawn images.
\newblock In {\em Advances in Neural Information Processing Systems (NIPS)},
  2018.

\bibitem{Fan2017CVPR}
Haoqiang Fan, Hao Su, and Leonidas~J. Guibas.
\newblock A point set generation network for 3d object reconstruction from a
  single image.
\newblock {\em Proc. IEEE Conf. on Computer Vision and Pattern Recognition
  (CVPR)}, 2017.

\bibitem{Girdhar2016ECCV}
Rohit Girdhar, David~F. Fouhey, Mikel Rodriguez, and Abhinav Gupta.
\newblock Learning a predictable and generative vector representation for
  objects.
\newblock In {\em Proc. of the European Conf. on Computer Vision (ECCV)}, 2016.

\bibitem{Groueix2018CVPR}
Thibault Groueix, Matthew Fisher, Vladimir~G. Kim, Bryan~C. Russell, and
  Mathieu Aubry.
\newblock {AtlasNet}: A papier-m\^ach\'e approach to learning 3d surface
  generation.
\newblock In {\em Proc. IEEE Conf. on Computer Vision and Pattern Recognition
  (CVPR)}, 2018.

\bibitem{Haene2017ARXIV}
Christian H{\"{a}}ne, Shubham Tulsiani, and Jitendra Malik.
\newblock Hierarchical surface prediction for 3d object reconstruction.
\newblock {\em arXiv.org}, 1704.00710, 2017.

\bibitem{Hartmann2017ICCV}
Wilfried Hartmann, Silvano Galliani, Michal Havlena, Luc {Van Gool}, and Konrad
  Schindler.
\newblock Learned multi-patch similarity.
\newblock In {\em Proc. of the IEEE International Conf. on Computer Vision
  (ICCV)}, 2017.

\bibitem{He2016CVPR}
Kaiming He, Xiangyu Zhang, Shaoqing Ren, and Jian Sun.
\newblock Deep residual learning for image recognition.
\newblock In {\em Proc. IEEE Conf. on Computer Vision and Pattern Recognition
  (CVPR)}, 2016.

\bibitem{Pohan2018CVPR}
Po-Han Huang, Kevin Matzen, Johannes Kopf, Narendra Ahuja, and Jia-Bin Huang.
\newblock Deepmvs: Learning multi-view stereopsis.
\newblock In {\em Proc. IEEE Conf. on Computer Vision and Pattern Recognition
  (CVPR)}, 2018.

\bibitem{Jaklic2000}
Ales Jaklic, Ales Leonardis, and Franc Solina.
\newblock {\em Segmentation and Recovery of Superquadrics}, volume~20 of {\em
  Computational Imaging and Vision}.
\newblock Springer, 2000.

\bibitem{Ji2017ICCV}
Mengqi Ji, Juergen Gall, Haitian Zheng, Yebin Liu, and Lu Fang.
\newblock {SurfaceNet:} an end-to-end 3d neural network for multiview
  stereopsis.
\newblock In {\em Proc. of the IEEE International Conf. on Computer Vision
  (ICCV)}, 2017.

\bibitem{Kingma2015ICLR}
Diederik~P. Kingma and Jimmy Ba.
\newblock Adam: {A} method for stochastic optimization.
\newblock In {\em Proc. of the International Conf. on Learning Representations
  (ICLR)}, 2015.

\bibitem{Laidlaw1986SIGGRAPH}
David~H Laidlaw, W~Benjamin Trumbore, and John~F Hughes.
\newblock Constructive solid geometry for polyhedral objects.
\newblock In {\em ACM Trans. on Graphics}, 1986.

\bibitem{Liao2018CVPR}
Yiyi Liao, Simon Donne, and Andreas Geiger.
\newblock Deep marching cubes: Learning explicit surface representations.
\newblock In {\em Proc. IEEE Conf. on Computer Vision and Pattern Recognition
  (CVPR)}, 2018.

\bibitem{Loper2015SIGGRAPH}
Matthew Loper, Naureen Mahmood, Javier Romero, Gerard Pons-Moll, and Michael~J.
  Black.
\newblock {SMPL}: A skinned multi-person linear model.
\newblock {\em ACM Trans. on Graphics}, 2015.

\bibitem{Niu2018CVPR}
Chengjie Niu, Jun Li, and Kai Xu.
\newblock Im2struct: Recovering 3d shape structure from a single {RGB} image.
\newblock In {\em Proc. IEEE Conf. on Computer Vision and Pattern Recognition
  (CVPR)}, 2018.

\bibitem{Paschalidou2018CVPR}
Despoina Paschalidou, Ali~Osman Ulusoy, Carolin Schmitt, Luc van Gool, and
  Andreas Geiger.
\newblock Raynet: Learning volumetric 3d reconstruction with ray potentials.
\newblock In {\em Proc. IEEE Conf. on Computer Vision and Pattern Recognition
  (CVPR)}, 2018.

\bibitem{Pentland1986AAAI}
Alex Pentland.
\newblock Parts: Structured descriptions of shape.
\newblock In {\em Proc. of the Conf. on Artificial Intelligence (AAAI)}, 1986.

\bibitem{Pilu1995BMVC}
Maurizio Pilu and Robert~B. Fisher.
\newblock Equal-distance sampling of supercllipse models.
\newblock In {\em Proc. of the British Machine Vision Conf. (BMVC)}, 1995.

\bibitem{Qi2017NIPS}
Charles~R Qi, Li Yi, Hao Su, and Leonidas~J Guibas.
\newblock Pointnet++: Deep hierarchical feature learning on point sets in a
  metric space.
\newblock In {\em Advances in Neural Information Processing Systems (NIPS)},
  2017.

\bibitem{Qi2007CVPR}
G.J. Qi, X.S. Hua, Y. Rui, T. Mei, J. Tang, and H.J. Zhang.
\newblock Concurrent multiple instance learning for image categorization.
\newblock In {\em Proc. IEEE Conf. on Computer Vision and Pattern Recognition
  (CVPR)}, 2007.

\bibitem{Rezende2016NIPS}
Danilo~Jimenez Rezende, S.~M.~Ali Eslami, Shakir Mohamed, Peter Battaglia, Max
  Jaderberg, and Nicolas Heess.
\newblock Unsupervised learning of 3d structure from images.
\newblock In {\em Advances in Neural Information Processing Systems (NIPS)},
  2016.

\bibitem{Riegler2017CVPR}
Gernot Riegler, Ali~Osman Ulusoy, and Andreas Geiger.
\newblock Octnet: Learning deep 3d representations at high resolutions.
\newblock In {\em Proc. IEEE Conf. on Computer Vision and Pattern Recognition
  (CVPR)}, 2017.

\bibitem{Rumelhart1986NATURE}
David~E. Rumelhart, Geoffrey~E. Hinton, and Ronald~J. Williams.
\newblock Learning representations by back-propagating errors.
\newblock {\em Nature}, 323:533--536, 1986.

\bibitem{Sharma2018CVPR}
Gopal Sharma, Rishabh Goyal, Difan Liu, Evangelos Kalogerakis, and Subhransu
  Maji.
\newblock Csgnet: Neural shape parser for constructive solid geometry.
\newblock In {\em Proc. IEEE Conf. on Computer Vision and Pattern Recognition
  (CVPR)}, 2018.

\bibitem{Solina1994JCIT}
Franc Solina.
\newblock Volumetric models in computer vision-an overview.
\newblock {\em Journal of Computing and Information technology}, 1994.

\bibitem{Solina1990PAMI}
Franc Solina and Ruzena Bajcsy.
\newblock Recovery of parametric models from range images: The case for
  superquadrics with global deformations.
\newblock {\em IEEE Trans. on Pattern Analysis and Machine Intelligence
  (PAMI)}, 1990.

\bibitem{Tatarchenko2017ICCV}
M. Tatarchenko, A. Dosovitskiy, and T. Brox.
\newblock Octree generating networks: Efficient convolutional architectures for
  high-resolution 3d outputs.
\newblock In {\em Proc. of the IEEE International Conf. on Computer Vision
  (ICCV)}, 2017.

\bibitem{Terzopoulos1990ICCV}
Demetri Terzopoulos and Dimitris~N. Metaxas.
\newblock Dynamic 3d models with local and global deformations: deformable
  superquadrics.
\newblock In {\em Proc. of the IEEE International Conf. on Computer Vision
  (ICCV)}, 1990.

\bibitem{Tulsiani2017CVPRa}
Shubham Tulsiani, Hao Su, Leonidas~J. Guibas, Alexei~A. Efros, and Jitendra
  Malik.
\newblock Learning shape abstractions by assembling volumetric primitives.
\newblock In {\em Proc. IEEE Conf. on Computer Vision and Pattern Recognition
  (CVPR)}, 2017.

\bibitem{Varol2017CVPR}
G{\"{u}}l Varol, Javier Romero, Xavier Martin, Naureen Mahmood, Michael~J.
  Black, Ivan Laptev, and Cordelia Schmid.
\newblock Learning from synthetic humans.
\newblock In {\em Proc. IEEE Conf. on Computer Vision and Pattern Recognition
  (CVPR)}, 2017.

\bibitem{Vaskevicius2017PAMI}
Narunas Vaskevicius and Andreas Birk.
\newblock Revisiting superquadric fitting: A numerically stable formulation.
\newblock {\em IEEE Trans. on Pattern Analysis and Machine Intelligence
  (PAMI)}, 2017.

\bibitem{Wang2018ECCV}
Nanyang Wang, Yinda Zhang, Zhuwen Li, Yanwei Fu, Wei Liu, and Yu{-}Gang Jiang.
\newblock Pixel2mesh: Generating 3d mesh models from single rgb images.
\newblock In {\em Proc. of the European Conf. on Computer Vision (ECCV)}, 2018.

\bibitem{Williams1992ML}
Ronald~J. Williams.
\newblock Simple statistical gradient-following algorithms for connectionist
  reinforcement learning.
\newblock {\em Machine Learning}, 8:229--256, 1992.

\bibitem{Wu2016NIPS}
Jiajun Wu, Chengkai Zhang, Tianfan Xue, Bill Freeman, and Josh Tenenbaum.
\newblock Learning a probabilistic latent space of object shapes via 3d
  generative-adversarial modeling.
\newblock In {\em Advances in Neural Information Processing Systems (NIPS)},
  2016.

\bibitem{Yao2018ECCV}
Yao Yao, Zixin Luo, Shiwei Li, Tian Fang, and Long Quan.
\newblock Mvsnet: Depth inference for unstructured multi-view stereo.
\newblock In {\em Proc. of the European Conf. on Computer Vision (ECCV)}, 2018.

\bibitem{Zou2017ICCV}
C. Zou, E. Yumer, J. Yang, D. Ceylan, and D. Hoiem.
\newblock 3d-prnn: Generating shape primitives with recurrent neural networks.
\newblock In {\em Proc. of the IEEE International Conf. on Computer Vision
  (ICCV)}, 2017.

\end{thebibliography}
}

\newpage
\clearpage
\appendix
\onecolumn

\begin{minipage}{\textwidth}
   \null
   \vskip .375in
   \begin{center}
      {\Large \bf Supplementary Material for\\Superquadrics Revisited: Learning 3D Shape Parsing beyond Cuboids \par}
      \vspace*{24pt}
      {
      \large
      \lineskip .5em
      \begin{tabular}[t]{c}
      Despoina Paschalidou$^{1,4}$ \quad Ali Osman Ulusoy$^{2}$ \quad Andreas Geiger$^{1,3,4}$\\
          $^1$Autonomous Vision Group, MPI for Intelligent Systems T{\"u}bingen\\
          $^2$Microsoft \quad
          $^3$University of T{\"u}bingen\quad
          $^4$Max Planck ETH Center for Learning Systems\\
          {\tt\small \{firstname.lastname\}@tue.mpg.de}
         \vspace*{1pt}\\%
      \end{tabular}
      \par
      }
      \vskip .5em
      \vspace*{12pt}
   \end{center}
\end{minipage}

\begin{abstract}
In this \textbf{supplementary document}, we present a detailed derivation of the proposed analytical solution to the Chamfer loss, which avoids the need for computationally expensive reinforcement learning or iterative prediction. Moreover, we also present additional qualitative results on more complex object categories from the ShapeNet dataset \cite{Chang2015ARXIV} such as \textit{cars} and \textit{motorbikes} and on the SURREAL human body dataset \cite{Varol2017CVPR}.
Furthermore, we also show results on primitive prediction when using RGB images instead of 3D occupancy grids as input.
Finally, we empirically demonstrate that our bi-directional Chamfer loss formulation indeed works better and results in less local minima than the original bi-directional loss formulation of Tulsiani \etal \cite{Tulsiani2017CVPRa}.
\end{abstract}

\section{Superquadrics}
\label{sec:superquadrics}

\begin{figure}[t!]
	\centering
	\includegraphics[width=0.75\textwidth]{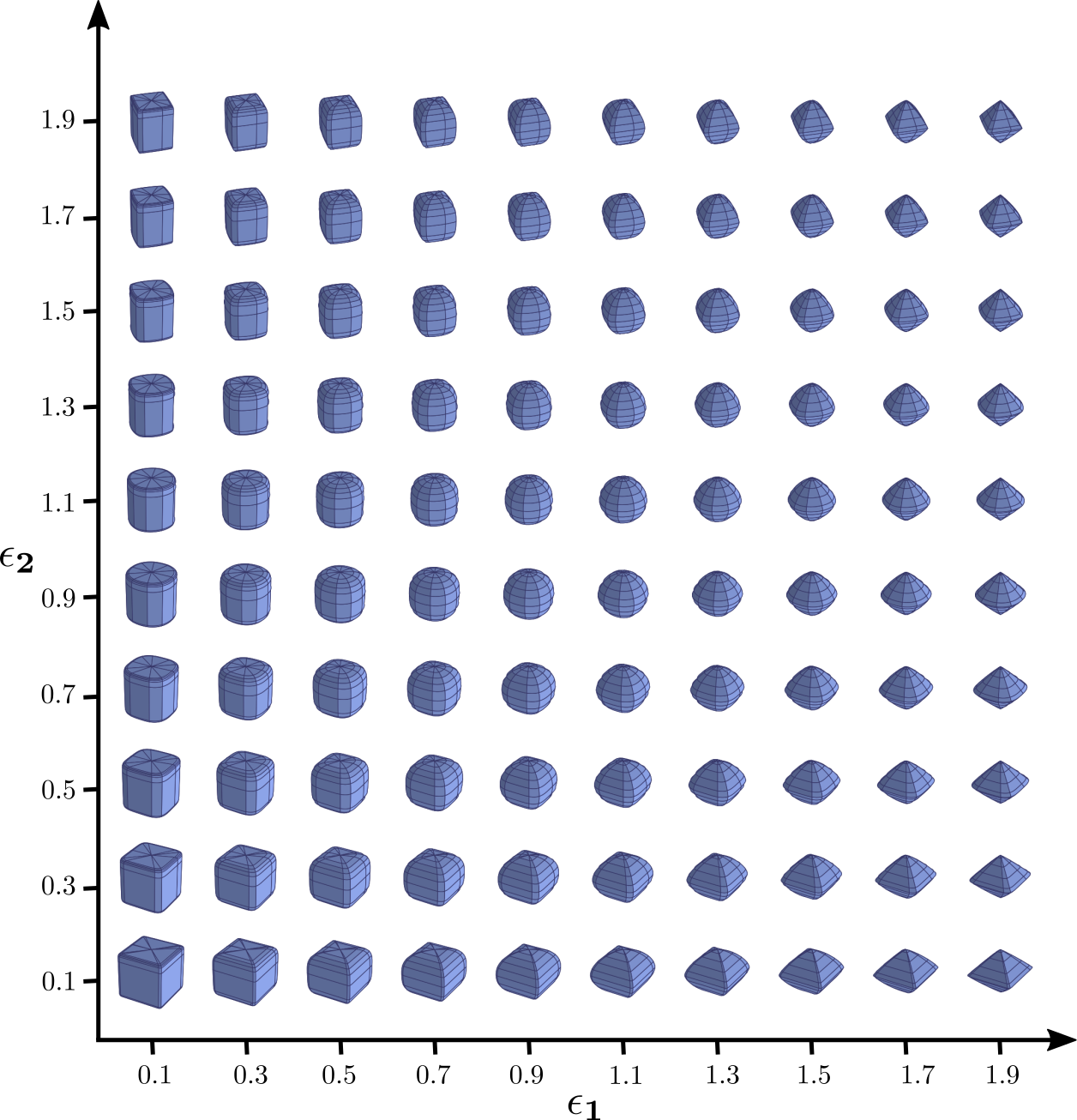}
	\caption{\textbf{Superquadric Shape Space.} Superquadrics are a parametric family of surfaces that can be used to describe cubes, cylinders, spheres, octahedral ellipsoids, \etc~\cite{Barr1981CGA}. This figure visualizes superquadrics when varying the shape parameters $\epsilon_1$ and $\epsilon_2$, while keeping the size parameters $\alpha_1$, $\alpha_2$ and $\alpha_3$ constant.}
	\label{fig:sqs_e1_e2_supp}
\end{figure}%

\begin{figure}[t!]
	\centering
	\includegraphics[width=0.5\textwidth]{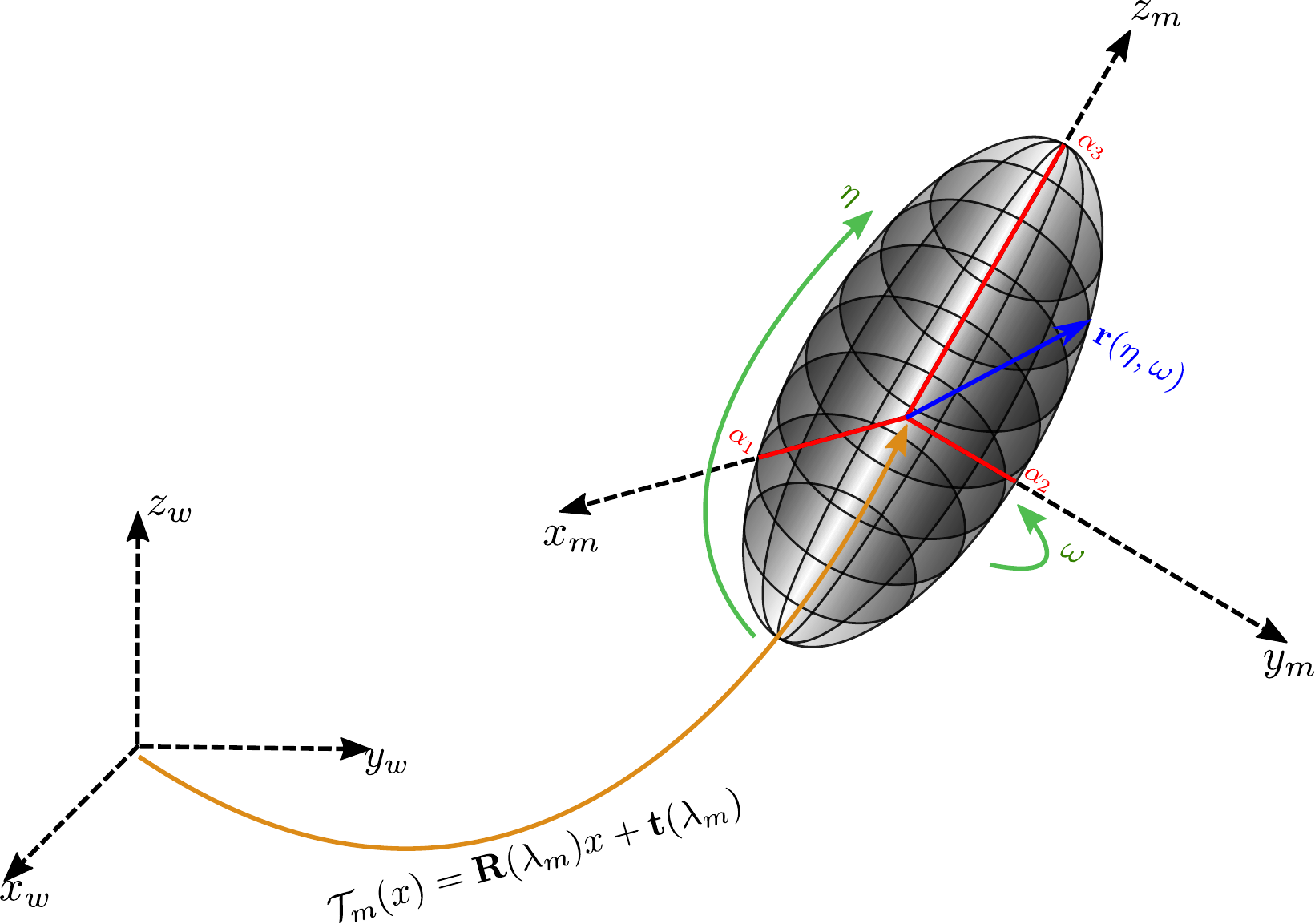}
	\caption{\textbf{Explicit Superquadric Equation}. A $3D$ vector $\mathbf{r}(\eta, \omega)$ defines a closed surface in space as $\eta$ (latitude angle) and $\omega$ (longitude angle) change in the given intervals \eqref{eq:parametric_eq_supp}. The rigid body transformation $\cT_m(x)$ maps a point from the world coordinate system to the local coordinate system of the $m^{th}$ primitive.}
	\label{fig:sqs_world}
\end{figure}%

In this work, we propose superquadrics as a shape primitive representation. Their simple parametrization in combination to their ability to represent a diverse class of shapes makes superquadrics a natural choice for geometric primitives. Moreover, their continuous parametrization is suitable for deep learning as their shape varies continuously with their parameters.
Superquadrics are fully modelled using a set of $11$ parameters \cite{Barr1981CGA}. The \textbf{explicit superquadric equation} defines the surface vector $\br$
\begin{equation}
    \mathbf{r}(\eta, \omega) =
    \begin{bmatrix}
        \alpha_{1}\cos^{\epsilon_{1}}\eta \cos^{\epsilon_{2}}\omega \\
        \alpha_{2}\cos^{\epsilon_{1}}\eta \sin^{\epsilon_{2}}\omega \\
        \alpha_{3}\sin^{\epsilon_{1}}\eta
    \end{bmatrix}
    \quad 
    \begin{aligned}
        -\pi/2 &\leq \eta \leq \pi/2\\
        -\pi &\leq \omega \leq \pi
    \end{aligned}
    \label{eq:parametric_eq_supp}
\end{equation}
where $\mathbf{\alpha} = [\alpha_{1}, \alpha_{2}, \alpha_{3}]$ determine the size and $\mathbf{\epsilon} = [\epsilon_{1}, \epsilon_{2}]$ determine the global shape of the superquadric. \figref{fig:sqs_e1_e2_supp} visualizes the shape of superquadrics for different values of $\epsilon_{1}$ and $\epsilon_{2}$.
In addition to the shape parameters, we also associate a rigid body transformation with each superquadric. This transformation is represented by a translation vector $\mathbf{t} = [t_{x}, t_{y}, t_{z}]$ and a quaternion $\mathbf{q} = [q_{0}, q_{1}, q_{2}, q_{3}]$ that determines the coordinate system transformation $\cT(\bx)=\bR(\lambda)\,\bx + \bt(\lambda)$ from world coordinates to local primitive-centric coordinates. This transformation as well as the angles $\eta,\omega$ and the scale parameters $\alpha_1,\alpha_2,\alpha_3$ are illustrated in \figref{fig:sqs_world}.

\section{Derivation of Pointcloud-to-Primitive Loss}

This section provides the derivation of the pointcloud-to-primitive distance $\mathcal{L}_{X\rightarrow P}(\bX, \bP)$ in Eq.~11 of the main paper.
For completeness, we restate our notation briefly.
We represent the target point cloud as a set of 3D points $\bX = \{\bx_i \}_{i=1}^N$ and we approximate the continuous surface of the $m^{th}$ primitive by a set of 3D points $\bY_m = \{\by_k^m \}_{k=1}^K$. We further denote $\cT_m(\bx)=\bR(\lambda_m)\,\bx + \bt(\lambda_m)$ as the mapping from world coordinates to the local coordinate system of the $m^{th}$ primitive.

The pointcloud-to-primitive distance, $\mathcal{L}_{X\rightarrow P}$, measures the distance from the point cloud to the primitives to ensure that each observation is explained by at least one primitive. It can be expressed as:
\begin{equation}
    \begin{aligned}
    &\cL_{X\rightarrow P}(\bX,\bP) = \nE_{p(\bz)} \left[\sum_{\bx_i \in \bX} 
     ~\min_{\substack{m | z_m=1}} \, \Delta^m_i \right] 
    \end{aligned}
    \label{eq:loss_2_supp}
\end{equation}
where $\Delta^m_i$ denotes the minimal distance from point $\bx_i$ to the surface of the $m$'th primitive: 
\begin{equation}
\Delta^m_i = \min_{k=1,..,K} {\Vert \cT_m(\bx_i) - \by_k^m \Vert}_2
\end{equation}
Assuming independence of the existence variables $p(\bz) = \prod_m p(z_m)$, we can replace the expectations in \eqref{eq:loss_2_supp} with summations as follows:
\begin{equation}
    \begin{aligned}
    &\cL_{X\rightarrow P}(\bX,\bP) = \sum_{z_1}\dots\sum_{z_M} \left[\sum_{\bx_i \in \bX} 
     ~\min_{\substack{m | z_m=1}} \, \Delta^m_i \right]p(\bz)
    \end{aligned}
    \label{eq:loss_3_supp}
\end{equation}
Na\"ive computation of \eqref{eq:loss_3_supp} has exponential complexity, \ie for $M$ primitives it requires evaluating the quantity inside the expectation $2^M$ times. Our key insight is that \eqref{eq:loss_3_supp} can be evaluated in linear time if the distances $\Delta_i^m$ are sorted. Without loss of generality, we assume that the distances are sorted in ascending order. This allows us to state the following:
if the first primitive exists, the first primitive will be the one closest to point $\bx_i$ of the target point, if the first primitive does not exist and the second does, then the second primitive is closest to point $\bx_i$ and so forth. More formally, this property can be stated as follows: 
\begin{equation}
    \begin{aligned}
    &\min_{\substack{m | z_m=1}} \Delta^m_i = 
    &\begin{cases}
        \Delta^1_i , & \text{if}\ z_1=1 \\
        \Delta^2_i, & \text{if}\ z_1=0, z_2=1 \\
        \vdots \\
        \Delta^M_i, & \text{if}\ z_m=0, \dots, z_M=1 \\
        \end{cases}
    \end{aligned}
    \label{eq:minimum_distances_supp}
\end{equation}
Using \eqref{eq:minimum_distances_supp} we can simplify \eqref{eq:loss_3_supp} as follows. We start to carry out the summations over the existence variables one by one. Starting with the summations over $z_1$, \eqref{eq:loss_3_supp} becomes:
\begin{equation}
    \begin{aligned}
    &\cL_{X\rightarrow P}(\bX,\bP) = \sum_{\bx_i \in \bX} \left[
        \underbrace{\gamma_1
        \sum_{z_2}\dots\sum_{z_M} \Delta^1_i \prod_{\bar{m}=2}^M p(z_{\bar{m}})}_{\text{(}\dagger\text{)}} + 
    (1 - \gamma_1) \sum_{z_2}\dots\sum_{z_M} \left[
        ~\min_{\substack{m\geq 2 | z_m=1}} \, \Delta^m_i \right] \prod_{\bar{m}=2}^M p(z_{\bar{m}})
     \right]
    \end{aligned}
    \label{eq:loss_4_supp}
\end{equation}
The expression, marked with $\text{(}\dagger\text{)}$ corresponds to the case for $z_1=1$, namely the $1^{st}$ primitive is part of the scene. From Eq.~\ref{eq:minimum_distances_supp}, we know that
$\min_{\substack{m | z_m=1}} \Delta^m_i = \Delta^1_i$ for $z_1=1$, thus the expression marked with $\text{(}\dagger\text{)}$, can be simplified as follows,
\begin{equation}
    \text{(}\dagger\text{)} = \gamma_1\Delta^1_i
        \underbrace{\sum_{z_2}\dots\sum_{z_M}\prod_{\bar{m}=2}^M p(z_{\bar{m}})}_{
        \text{this term evaluates to 1}} = \gamma_1\Delta^1_i
\end{equation}

Following this strategy, we can iteratively simplify the remaining terms in \eqref{eq:loss_4_supp} and arrive at the analytical form of the pointcloud-to-primitive distance stated in Eq.~11 in the main paper:
\begin{equation}
    \begin{aligned}
    \cL_{X\rightarrow P}(\bX,\bP) &= \sum_{\bx_i \in \bX} \left[
        \gamma_1\Delta^1_i + (1-\gamma_1)\gamma_2\Delta^2_i +
        \dots + (1-\gamma_1)(1-\gamma_2)\dots\gamma_M\Delta^M_i
     \right] \\
    &=\sum_{\bx_i \in \bX} \sum_{m=1}^M 
    \Delta^m_i \, \gamma_m \prod_{\bar{m}=1}^{m-1}(1 - \gamma_{\bar{m}})
    \end{aligned}
\end{equation}
Note that our current formulation assumes that at least one primitive exists in the scene. However, this assumption can be easily relaxed by introducing a ``virtual primitive" with a fixed distance to every 3D point on the target point cloud.

\section{Qualitative Results on SURREAL}
\label{sec:qualitative_surreal}

\begin{figure}[h!]
	\centering
	\begin{subfigure}[b]{0.15\linewidth}
    	\centering
        \includegraphics[width=\linewidth]{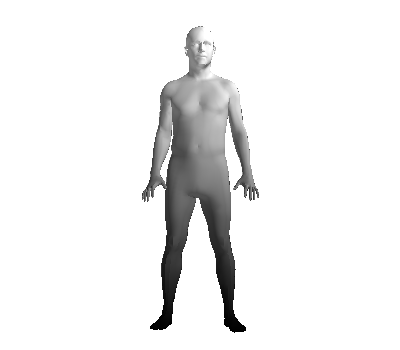}
    \end{subfigure}
    \hfill
    \begin{subfigure}[b]{0.15\linewidth}
    	\centering
        \includegraphics[width=\linewidth]{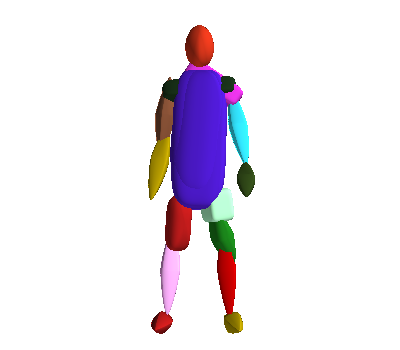}
    \end{subfigure}
    \hfill
    \begin{subfigure}[b]{0.15\linewidth}
    	\centering
        \includegraphics[width=\linewidth]{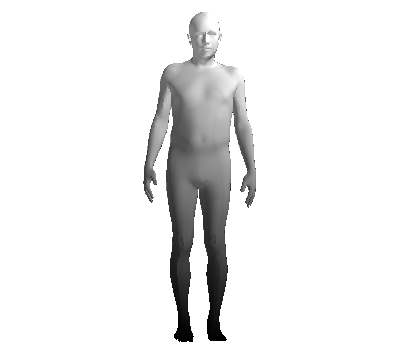}
    \end{subfigure}
    \hfill
    \begin{subfigure}[b]{0.15\linewidth}
    	\centering
        \includegraphics[width=\linewidth]{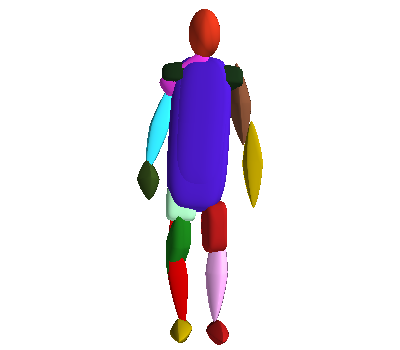}
    \end{subfigure}
    \hfill
    \begin{subfigure}[b]{0.15\linewidth}
    	\centering
        \includegraphics[width=\linewidth]{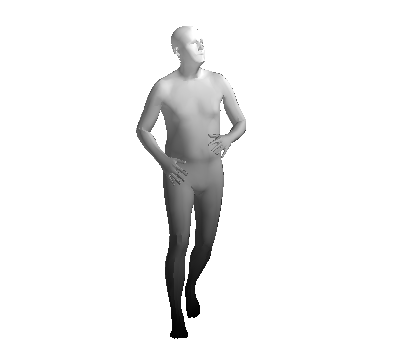}
    \end{subfigure}
    \hfill
    \begin{subfigure}[b]{0.15\linewidth}
    	\centering
        \includegraphics[width=\linewidth]{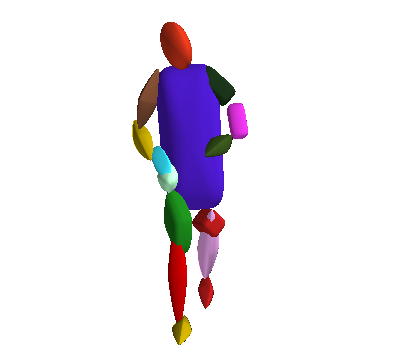}
    \end{subfigure}
    \vskip\baselineskip
	\begin{subfigure}[b]{0.15\linewidth}
    	\centering
        \includegraphics[width=\linewidth]{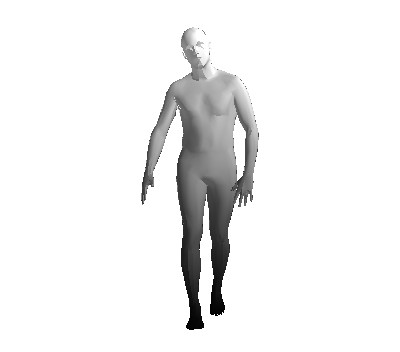}
    \end{subfigure}
    \hfill
    \begin{subfigure}[b]{0.15\linewidth}
    	\centering
        \includegraphics[width=\linewidth]{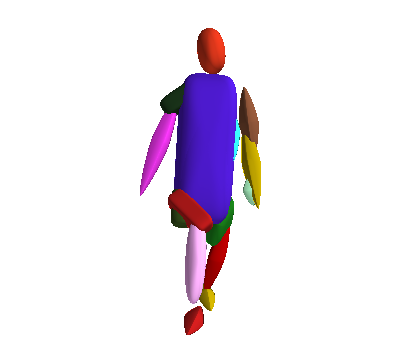}
    \end{subfigure}
    \hfill
    \begin{subfigure}[b]{0.15\linewidth}
    	\centering
        \includegraphics[width=\linewidth]{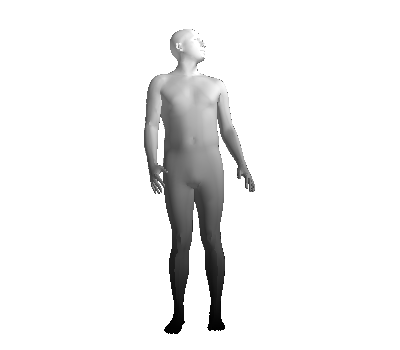}
    \end{subfigure}
    \hfill
    \begin{subfigure}[b]{0.15\linewidth}
    	\centering
        \includegraphics[width=\linewidth]{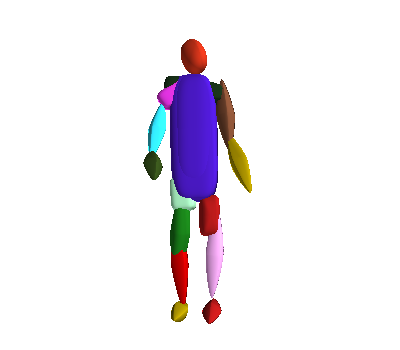}
    \end{subfigure}
    \hfill
    \begin{subfigure}[b]{0.15\linewidth}
    	\centering
        \includegraphics[width=\linewidth]{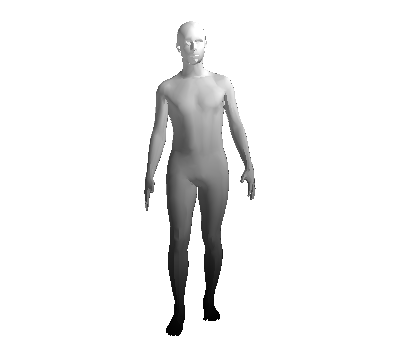}
    \end{subfigure}
    \hfill
    \begin{subfigure}[b]{0.15\linewidth}
    	\centering
        \includegraphics[width=\linewidth]{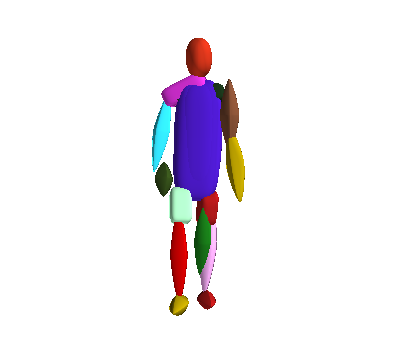}
    \end{subfigure}
    \vskip\baselineskip
	\begin{subfigure}[b]{0.15\linewidth}
    	\centering
        \includegraphics[width=\linewidth]{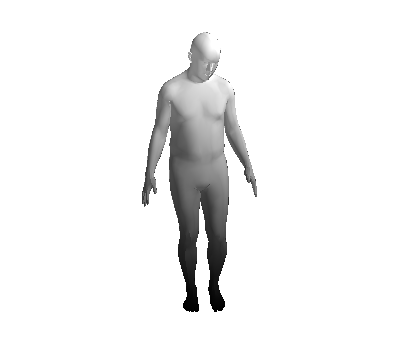}
    \end{subfigure}
    \hfill
    \begin{subfigure}[b]{0.15\linewidth}
    	\centering
        \includegraphics[width=\linewidth]{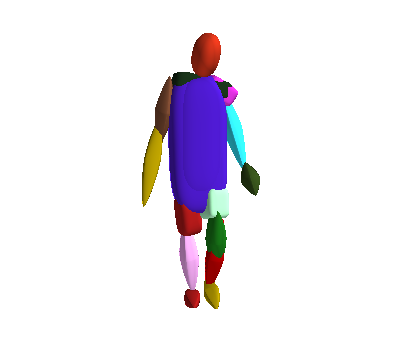}
    \end{subfigure}
    \hfill
    \begin{subfigure}[b]{0.15\linewidth}
    	\centering
        \includegraphics[width=\linewidth]{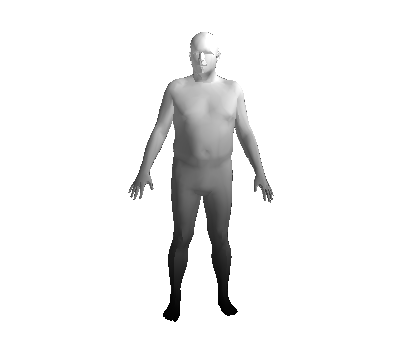}
    \end{subfigure}
    \hfill
    \begin{subfigure}[b]{0.15\linewidth}
    	\centering
        \includegraphics[width=\linewidth]{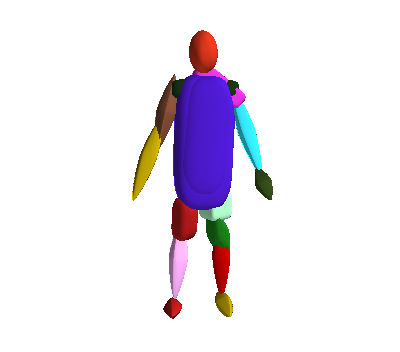}
    \end{subfigure}
    \hfill
    \begin{subfigure}[b]{0.15\linewidth}
    	\centering
        \includegraphics[width=\linewidth]{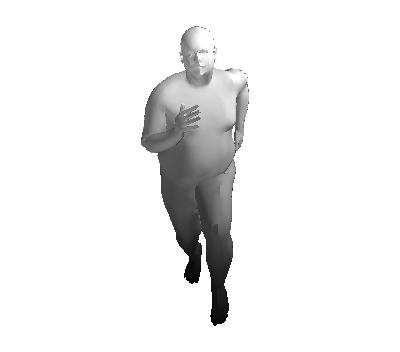}
    \end{subfigure}
    \hfill
    \begin{subfigure}[b]{0.15\linewidth}
    	\centering
        \includegraphics[width=\linewidth]{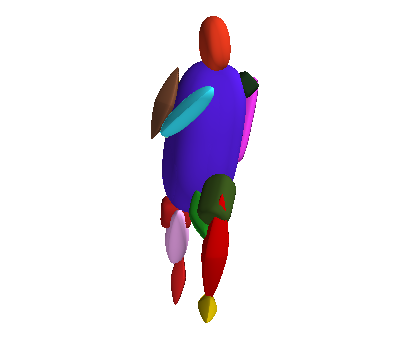}
    \end{subfigure}
    \vskip\baselineskip
	\begin{subfigure}[b]{0.15\linewidth}
    	\centering
        \includegraphics[width=\linewidth]{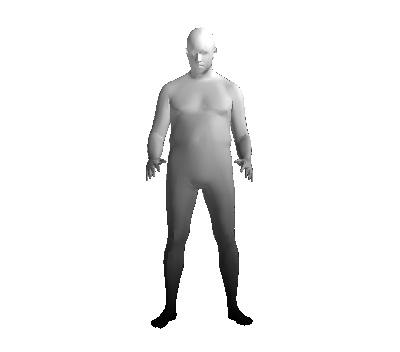}
    \end{subfigure}
    \hfill
    \begin{subfigure}[b]{0.15\linewidth}
    	\centering
        \includegraphics[width=\linewidth]{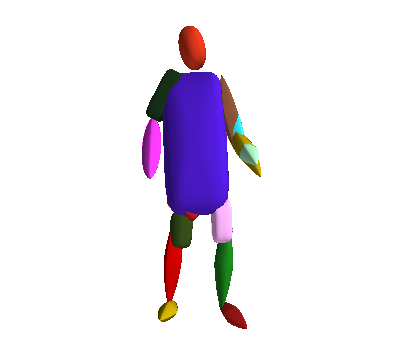}
    \end{subfigure}
    \hfill
    \begin{subfigure}[b]{0.15\linewidth}
    	\centering
        \includegraphics[width=\linewidth]{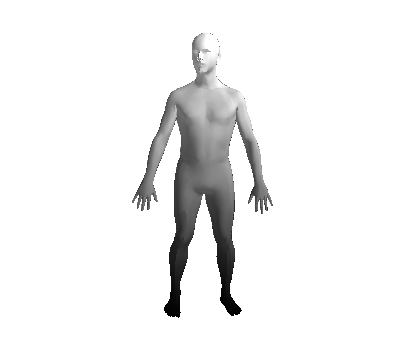}
    \end{subfigure}
    \hfill
    \begin{subfigure}[b]{0.15\linewidth}
    	\centering
        \includegraphics[width=\linewidth]{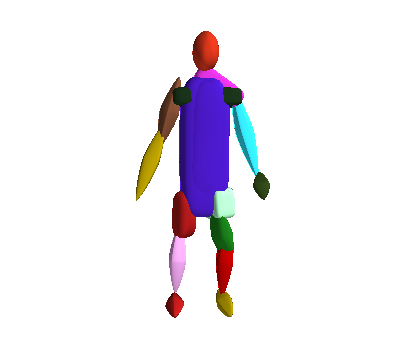}
    \end{subfigure}
    \hfill
    \begin{subfigure}[b]{0.15\linewidth}
    	\centering
        \includegraphics[width=\linewidth]{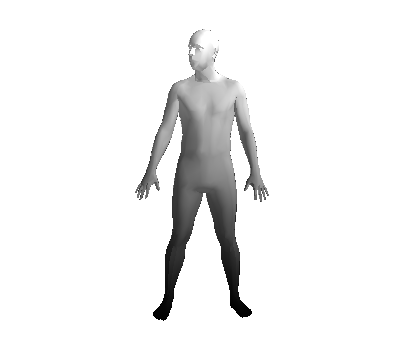}
    \end{subfigure}
    \hfill
    \begin{subfigure}[b]{0.15\linewidth}
    	\centering
        \includegraphics[width=\linewidth]{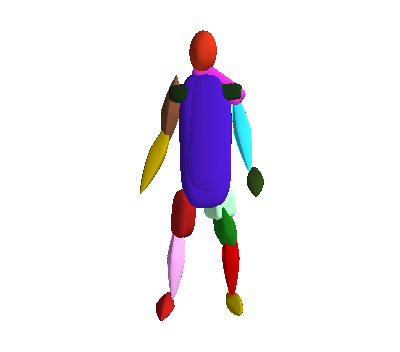}
    \end{subfigure}
    \vskip\baselineskip
	\begin{subfigure}[b]{0.15\linewidth}
    	\centering
        \includegraphics[width=\linewidth]{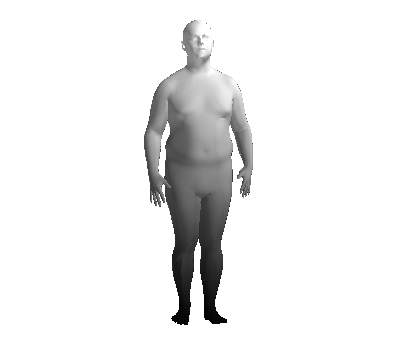}
    \end{subfigure}
    \hfill
    \begin{subfigure}[b]{0.15\linewidth}
    	\centering
        \includegraphics[width=\linewidth]{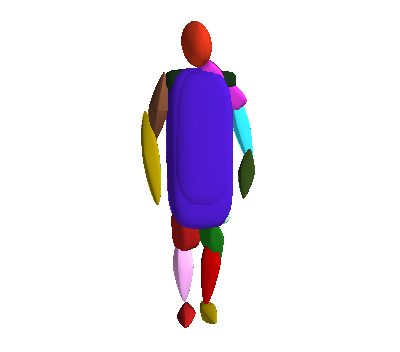}
    \end{subfigure}
    \hfill
    \begin{subfigure}[b]{0.15\linewidth}
    	\centering
        \includegraphics[width=\linewidth]{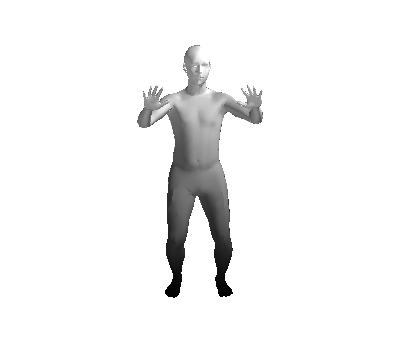}
    \end{subfigure}
    \hfill
    \begin{subfigure}[b]{0.15\linewidth}
    	\centering
        \includegraphics[width=\linewidth]{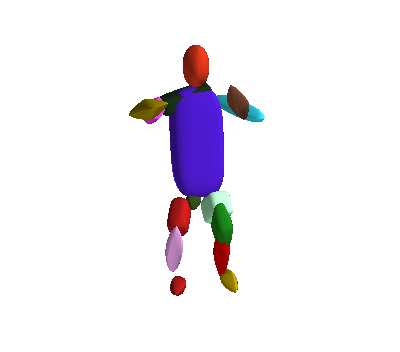}
    \end{subfigure}
    \hfill
    \begin{subfigure}[b]{0.15\linewidth}
    	\centering
        \includegraphics[width=\linewidth]{humans_experiment_gt_000323}
    \end{subfigure}
    \hfill
    \begin{subfigure}[b]{0.15\linewidth}
    	\centering
        \includegraphics[width=\linewidth]{humans_experiment_sq_000323}
    \end{subfigure}
    \vskip\baselineskip
	\begin{subfigure}[b]{0.15\linewidth}
    	\centering
        \includegraphics[width=\linewidth]{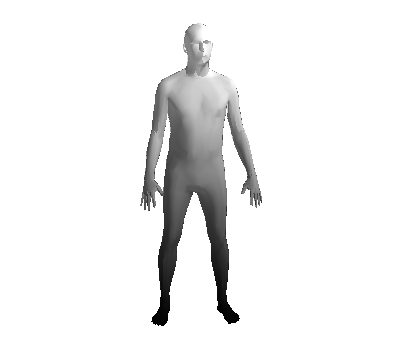}
    \end{subfigure}
    \hfill
    \begin{subfigure}[b]{0.15\linewidth}
    	\centering
        \includegraphics[width=\linewidth]{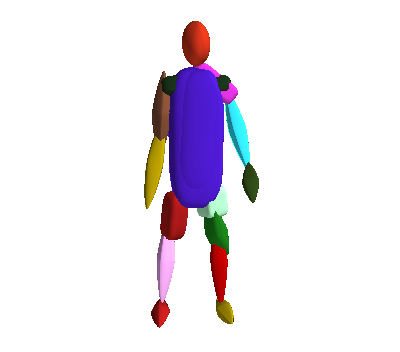}
    \end{subfigure}
    \hfill
    \begin{subfigure}[b]{0.15\linewidth}
    	\centering
        \includegraphics[width=\linewidth]{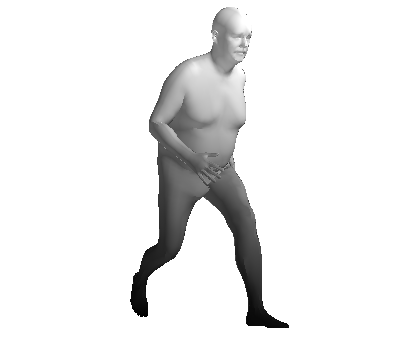}
    \end{subfigure}
    \hfill
    \begin{subfigure}[b]{0.15\linewidth}
    	\centering
        \includegraphics[width=\linewidth]{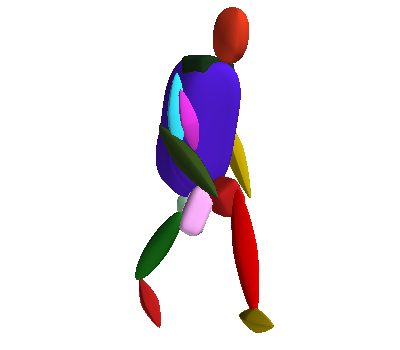}
    \end{subfigure}
    \hfill
    \begin{subfigure}[b]{0.15\linewidth}
    	\centering
        \includegraphics[width=\linewidth]{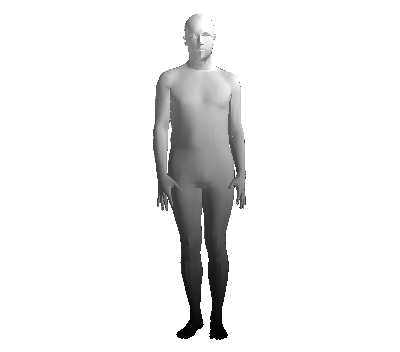}
    \end{subfigure}
    \hfill
    \begin{subfigure}[b]{0.15\linewidth}
    	\centering
        \includegraphics[width=\linewidth]{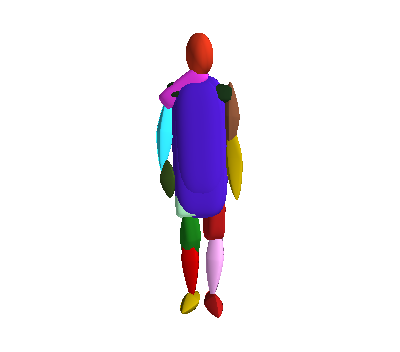}
    \end{subfigure}
    \vskip\baselineskip
	\begin{subfigure}[b]{0.15\linewidth}
    	\centering
        \includegraphics[width=\linewidth]{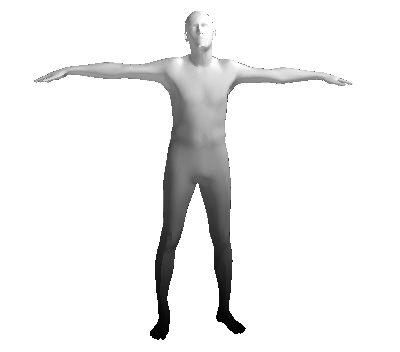}
    \end{subfigure}
    \hfill
    \begin{subfigure}[b]{0.15\linewidth}
    	\centering
        \includegraphics[width=\linewidth]{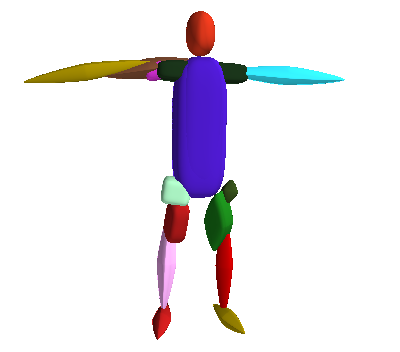}
    \end{subfigure}
    \hfill
	\begin{subfigure}[b]{0.15\linewidth}
    	\centering
        \includegraphics[width=\linewidth]{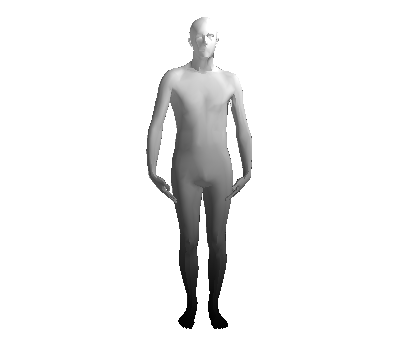}
    \end{subfigure}
    \hfill
    \begin{subfigure}[b]{0.15\linewidth}
    	\centering
        \includegraphics[width=\linewidth]{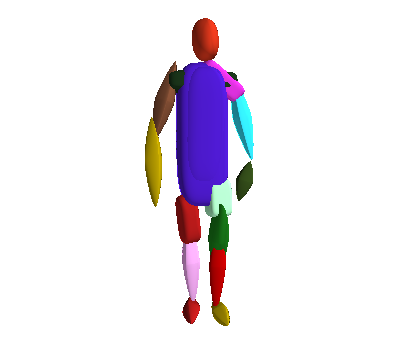}
    \end{subfigure}
    \hfill
	\begin{subfigure}[b]{0.15\linewidth}
    	\centering
        \includegraphics[width=\linewidth]{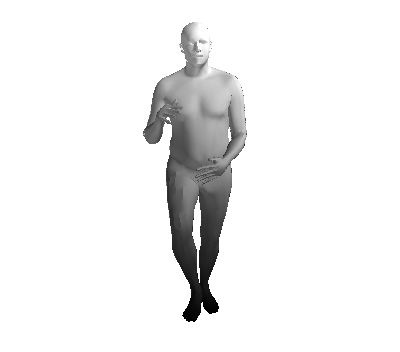}
    \end{subfigure}
    \hfill
    \begin{subfigure}[b]{0.15\linewidth}
    	\centering
        \includegraphics[width=\linewidth]{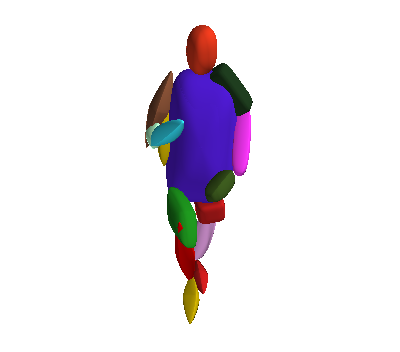}
    \end{subfigure}
    \caption{\textbf{Qualitative Results on SURREAL.} Our network learns semantic mappings of body parts across different body shapes and articulations. For instance, the network uses the same primitive for the left forearm across instances.}
    \label{fig:surreal_supp}
    \vspace{-1em}
\end{figure}

In this section, we provide additional qualitative results on the SURREAL human body dataset. In \figref{fig:surreal_supp}, we illustrate the predicted primitives of humans in various poses and articulations.

We remark that our model is able to accurately capture the various human body parts using superquadric surfaces. Another interesting aspect of our model, which is also observed in \cite{Tulsiani2017CVPRa}, is related to the fact that our model uses the same primitive (highlighted with the same color) to represent the same actual human body part. For example, the head is typically captured using the primitive illustrated with red. For some poses these correspondences are lost. We speculate that this is because the network does not know whether the human is facing in front or behind.

\section{Qualitative Results on ShapeNet}
\label{sec:qualitative_shapenet}

\begin{figure}[t!]
	\centering
	\begin{subfigure}[b]{0.15\linewidth}
    	\centering
        \includegraphics[width=\linewidth]{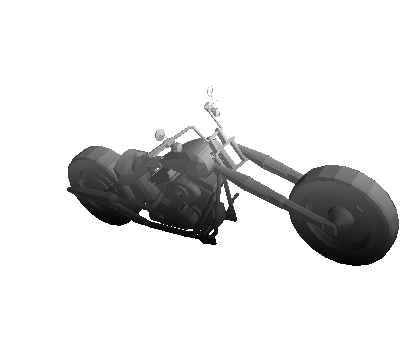}
    \end{subfigure}
    \hfill
    \begin{subfigure}[b]{0.15\linewidth}
    	\centering
        \includegraphics[width=\linewidth]{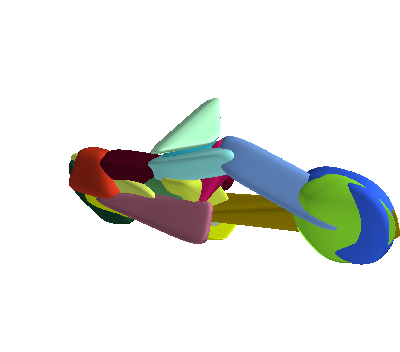}
    \end{subfigure}
    \hfill
    \begin{subfigure}[b]{0.15\linewidth}
    	\centering
        \includegraphics[width=\linewidth]{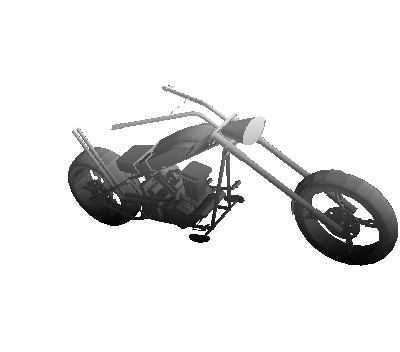}
    \end{subfigure}
    \hfill
    \begin{subfigure}[b]{0.15\linewidth}
    	\centering
        \includegraphics[width=\linewidth]{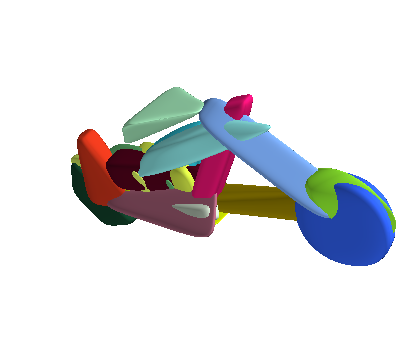}
    \end{subfigure}
    \hfill
    \begin{subfigure}[b]{0.15\linewidth}
    	\centering
        \includegraphics[width=\linewidth]{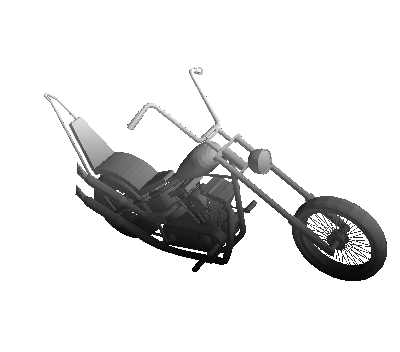}
    \end{subfigure}
    \hfill
    \begin{subfigure}[b]{0.15\linewidth}
    	\centering
        \includegraphics[width=\linewidth]{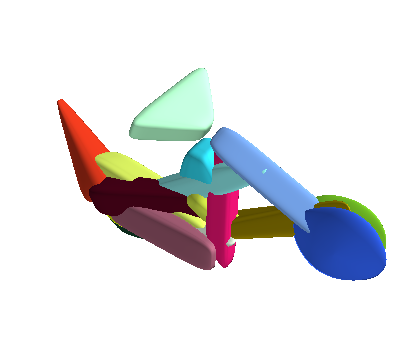}
    \end{subfigure}
    \vskip\baselineskip
	\begin{subfigure}[b]{0.15\linewidth}
    	\centering
        \includegraphics[width=\linewidth]{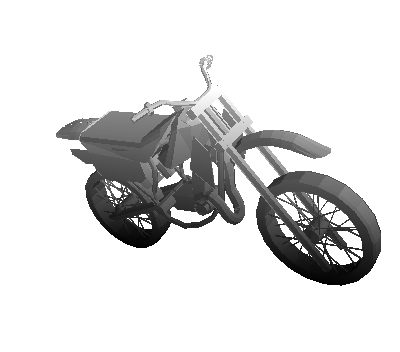}
    \end{subfigure}
    \hfill
    \begin{subfigure}[b]{0.15\linewidth}
    	\centering
        \includegraphics[width=\linewidth]{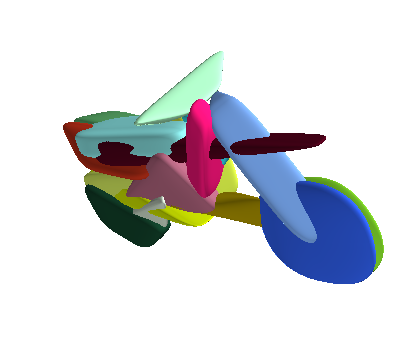}
    \end{subfigure}
    \hfill
    \begin{subfigure}[b]{0.15\linewidth}
    	\centering
        \includegraphics[width=\linewidth]{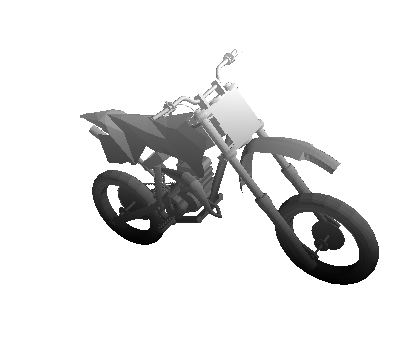}
    \end{subfigure}
    \hfill
    \begin{subfigure}[b]{0.15\linewidth}
    	\centering
        \includegraphics[width=\linewidth]{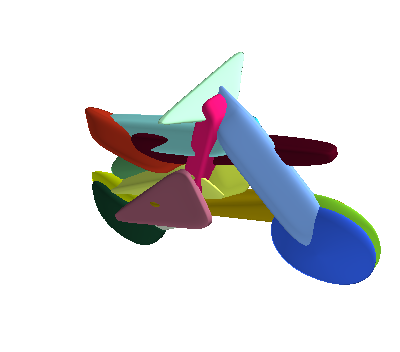}
    \end{subfigure}
    \hfill
    \begin{subfigure}[b]{0.15\linewidth}
    	\centering
        \includegraphics[width=\linewidth]{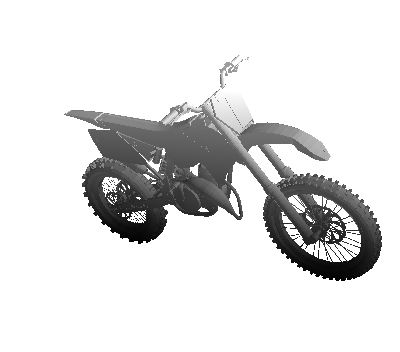}
    \end{subfigure}
    \hfill
    \begin{subfigure}[b]{0.15\linewidth}
    	\centering
        \includegraphics[width=\linewidth]{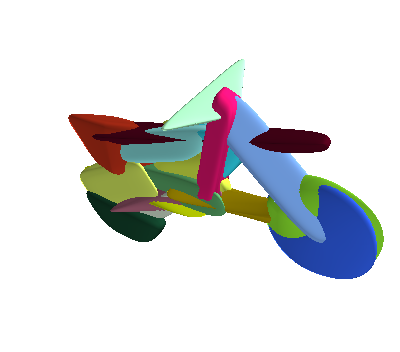}
    \end{subfigure}
    \vskip\baselineskip
	\begin{subfigure}[b]{0.15\linewidth}
    	\centering
        \includegraphics[width=\linewidth]{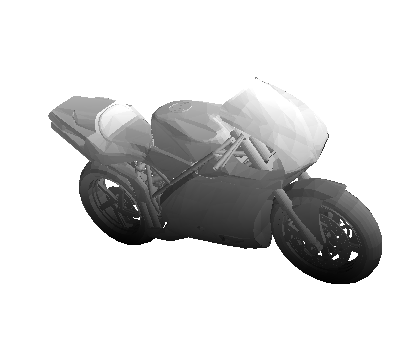}
    \end{subfigure}
    \hfill
    \begin{subfigure}[b]{0.15\linewidth}
    	\centering
        \includegraphics[width=\linewidth]{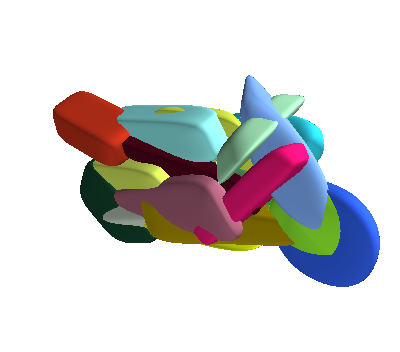}
    \end{subfigure}
    \hfill
    \begin{subfigure}[b]{0.15\linewidth}
    	\centering
        \includegraphics[width=\linewidth]{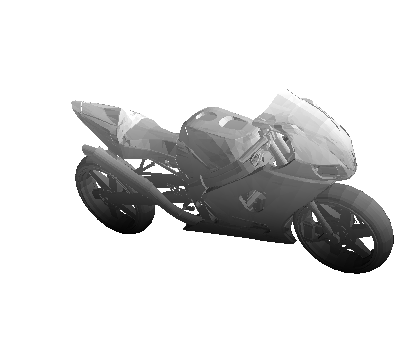}
    \end{subfigure}
    \hfill
    \begin{subfigure}[b]{0.15\linewidth}
    	\centering
        \includegraphics[width=\linewidth]{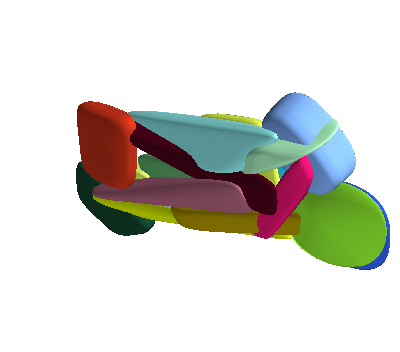}
    \end{subfigure}
    \hfill
    \begin{subfigure}[b]{0.15\linewidth}
    	\centering
        \includegraphics[width=\linewidth]{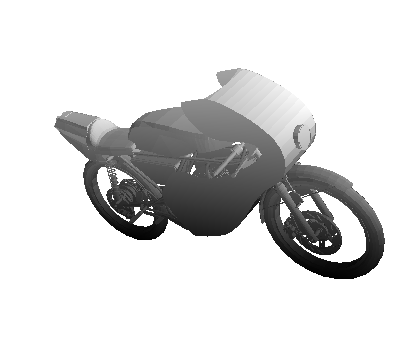}
    \end{subfigure}
    \hfill
    \begin{subfigure}[b]{0.15\linewidth}
    	\centering
        \includegraphics[width=\linewidth]{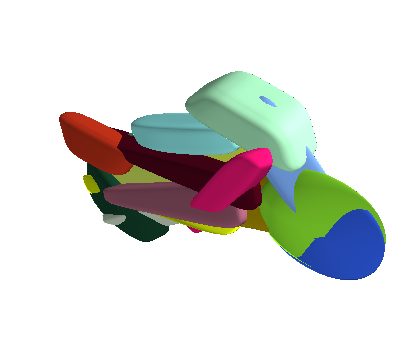}
    \end{subfigure}
    \vskip\baselineskip
	\begin{subfigure}[b]{0.15\linewidth}
    	\centering
        \includegraphics[width=\linewidth]{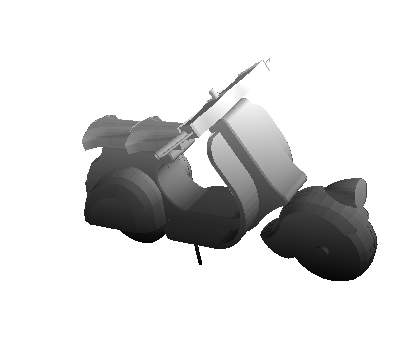}
    \end{subfigure}
    \hfill
    \begin{subfigure}[b]{0.15\linewidth}
    	\centering
        \includegraphics[width=\linewidth]{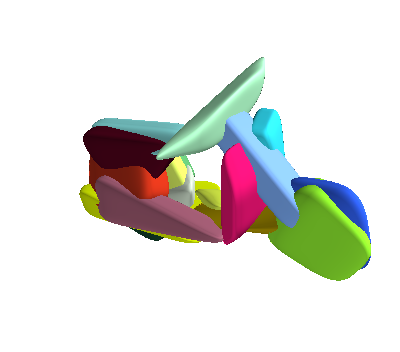}
    \end{subfigure}
    \hfill
    \begin{subfigure}[b]{0.15\linewidth}
    	\centering
        \includegraphics[width=\linewidth]{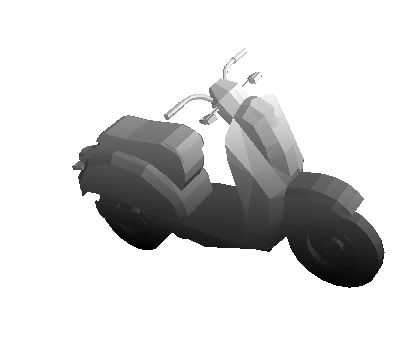}
    \end{subfigure}
    \hfill
    \begin{subfigure}[b]{0.15\linewidth}
    	\centering
        \includegraphics[width=\linewidth]{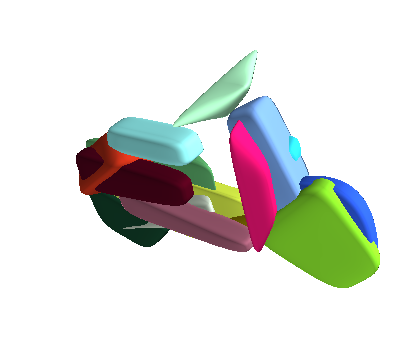}
    \end{subfigure}
    \hfill
    \begin{subfigure}[b]{0.15\linewidth}
    	\centering
        \includegraphics[width=\linewidth]{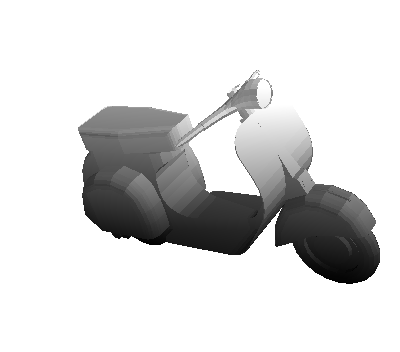}
    \end{subfigure}
    \hfill
    \begin{subfigure}[b]{0.15\linewidth}
    	\centering
        \includegraphics[width=\linewidth]{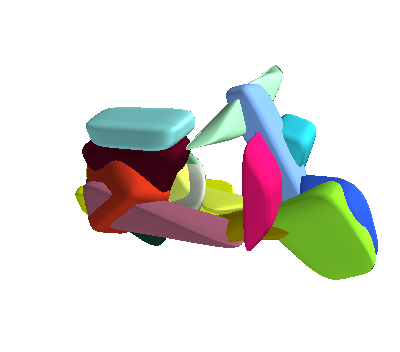}
    \end{subfigure}
    \vskip\baselineskip
	\begin{subfigure}[b]{0.15\linewidth}
    	\centering
        \includegraphics[width=\linewidth]{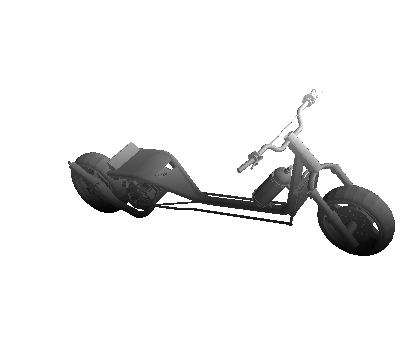}
    \end{subfigure}
    \hfill
    \begin{subfigure}[b]{0.15\linewidth}
    	\centering
        \includegraphics[width=\linewidth]{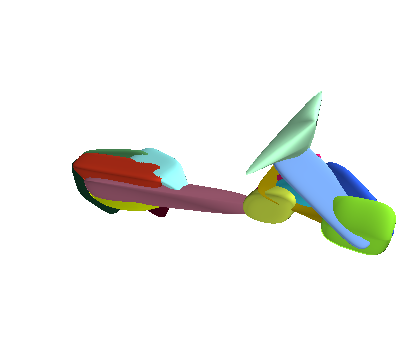}
    \end{subfigure}
    \hfill
    \begin{subfigure}[b]{0.15\linewidth}
    	\centering
        \includegraphics[width=\linewidth]{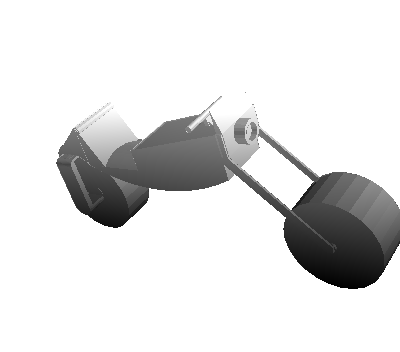}
    \end{subfigure}
    \hfill
    \begin{subfigure}[b]{0.15\linewidth}
    	\centering
        \includegraphics[width=\linewidth]{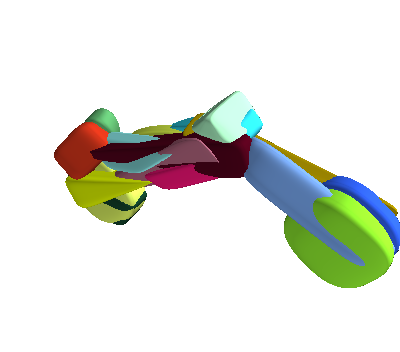}
    \end{subfigure}
    \hfill
    \begin{subfigure}[b]{0.15\linewidth}
    	\centering
        \includegraphics[width=\linewidth]{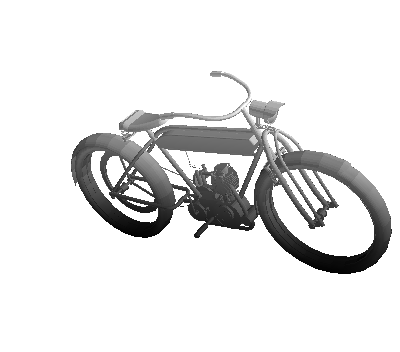}
    \end{subfigure}
    \hfill
    \begin{subfigure}[b]{0.15\linewidth}
    	\centering
        \includegraphics[width=\linewidth]{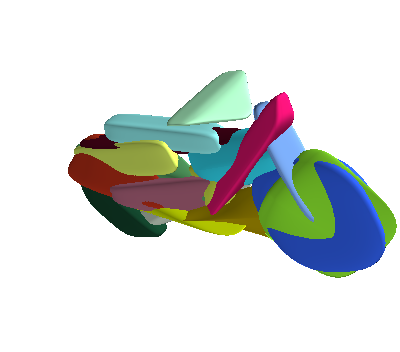}
    \end{subfigure}
    \vskip\baselineskip
	\begin{subfigure}[b]{0.15\linewidth}
    	\centering
        \includegraphics[width=\linewidth]{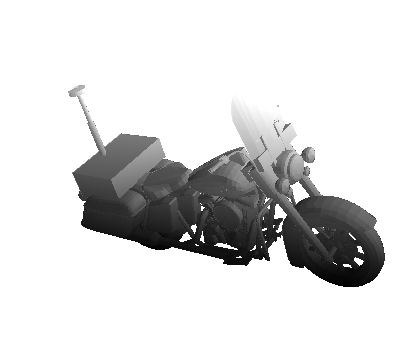}
    \end{subfigure}
    \hfill
    \begin{subfigure}[b]{0.15\linewidth}
    	\centering
        \includegraphics[width=\linewidth]{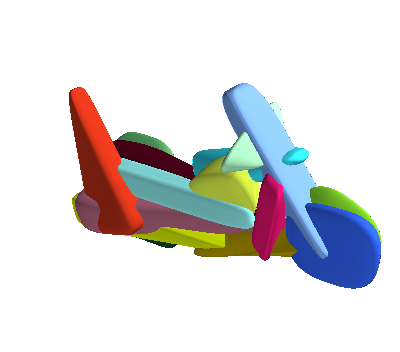}
    \end{subfigure}
    \hfill
	\begin{subfigure}[b]{0.15\linewidth}
    	\centering
        \includegraphics[width=\linewidth]{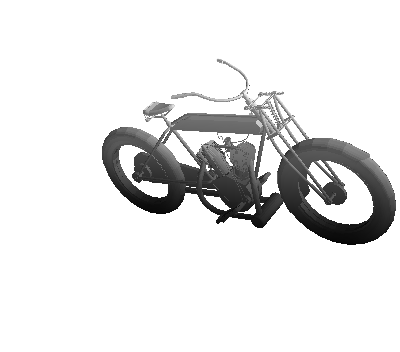}
    \end{subfigure}
    \hfill
    \begin{subfigure}[b]{0.15\linewidth}
    	\centering
        \includegraphics[width=\linewidth]{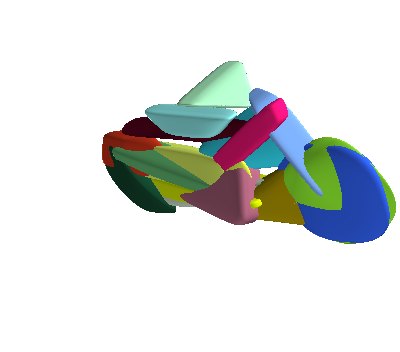}
    \end{subfigure}
    \hfill
	\begin{subfigure}[b]{0.15\linewidth}
    	\centering
        \includegraphics[width=\linewidth]{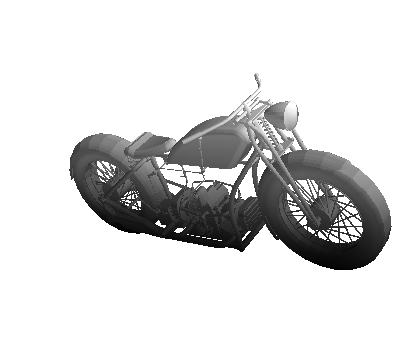}
    \end{subfigure}
    \hfill
    \begin{subfigure}[b]{0.15\linewidth}
    	\centering
        \includegraphics[width=\linewidth]{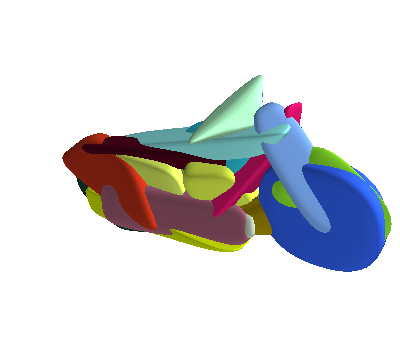}
    \end{subfigure}
    \caption{\textbf{Qualitative Results on \emph{motorbikes} from ShapeNet.} Our network learns semantic mappings of various object parts of different objects within the same category. Our expressive shape abstractions allow for differentiating between different types of motorbikes (scooter, racing bike, chopper \etc), by sucessfully capturing the shape of various indicative parts such as the wheels or the front fork of the bike.}
    \label{fig:bikes_supp}
    \vspace{-1em}
\end{figure}

\begin{figure}[b!]
	\centering
	\begin{subfigure}[b]{0.15\textwidth}
		\centering
		\includegraphics[width=\textwidth]{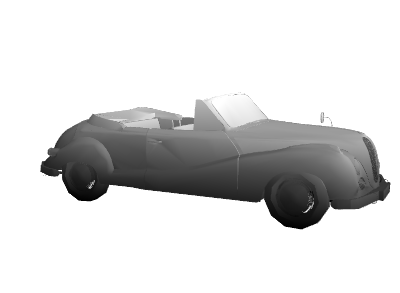}
	\end{subfigure}
	\hfill
	\begin{subfigure}[b]{0.15\textwidth}
		\centering
		\includegraphics[width=\textwidth]{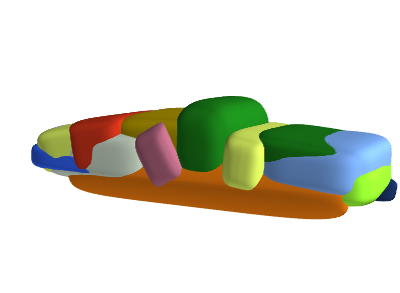}
	\end{subfigure}
	\hfill
	\begin{subfigure}[b]{0.15\textwidth}
		\centering
		\includegraphics[width=\textwidth]{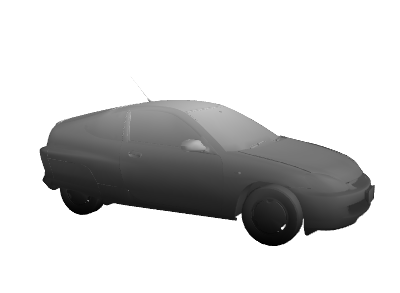}
	\end{subfigure}
	\hfill
	\begin{subfigure}[b]{0.15\textwidth}
		\centering
		\includegraphics[width=\textwidth]{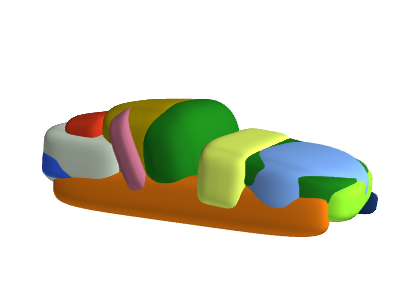}
	\end{subfigure}
	\hfill
	\begin{subfigure}[b]{0.15\textwidth}
		\centering
		\includegraphics[width=\textwidth]{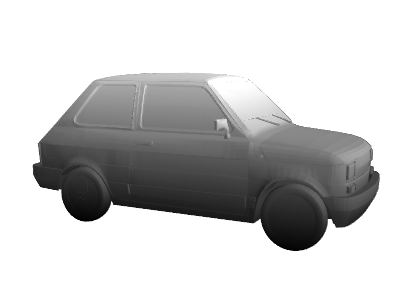}
	\end{subfigure}
	\hfill
	\begin{subfigure}[b]{0.15\textwidth}
		\centering
		\includegraphics[width=\textwidth]{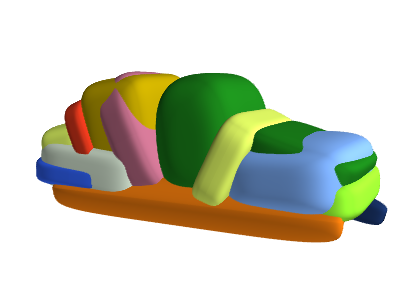}
	\end{subfigure}
	\vskip\baselineskip
	\begin{subfigure}[b]{0.15\textwidth}
		\centering
		\includegraphics[width=\textwidth]{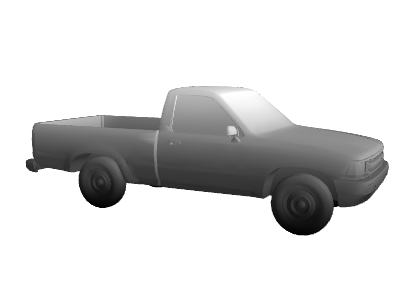}
	\end{subfigure}
	\hfill
	\begin{subfigure}[b]{0.15\textwidth}
		\centering
		\includegraphics[width=\textwidth]{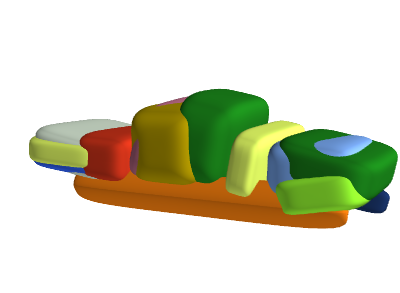}
	\end{subfigure}
	\hfill
	\begin{subfigure}[b]{0.15\textwidth}
		\centering
		\includegraphics[width=\textwidth]{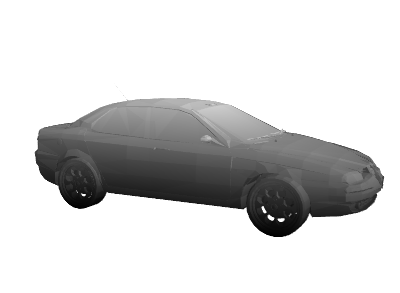}
	\end{subfigure}
	\hfill
	\begin{subfigure}[b]{0.15\textwidth}
		\centering
		\includegraphics[width=\textwidth]{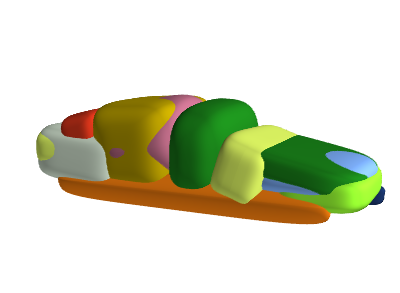}
	\end{subfigure}
	\hfill
	\begin{subfigure}[b]{0.15\textwidth}
		\centering
		\includegraphics[width=\textwidth]{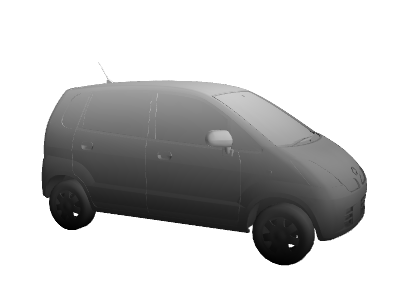}
	\end{subfigure}
	\hfill
	\begin{subfigure}[b]{0.15\textwidth}
		\centering
		\includegraphics[width=\textwidth]{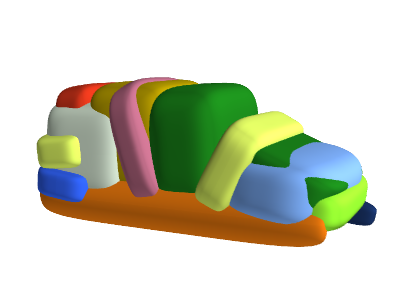}
	\end{subfigure}
	\vskip\baselineskip
	\begin{subfigure}[b]{0.15\textwidth}
		\centering
		\includegraphics[width=\textwidth]{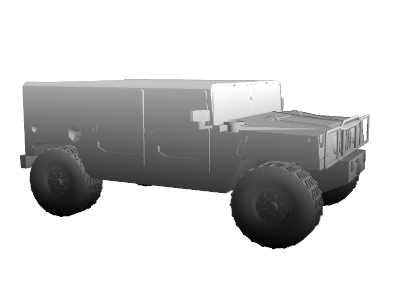}
	\end{subfigure}
	\hfill
	\begin{subfigure}[b]{0.15\textwidth}
		\centering
		\includegraphics[width=\textwidth]{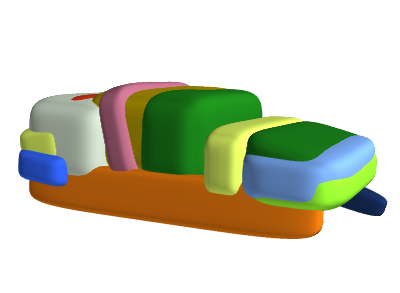}
	\end{subfigure}
	\hfill
	\begin{subfigure}[b]{0.15\textwidth}
		\centering
		\includegraphics[width=\textwidth]{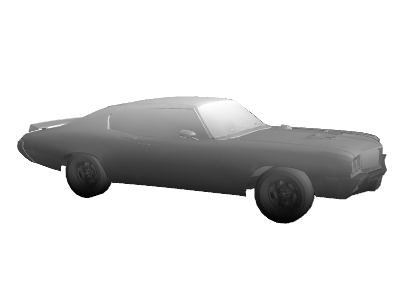}
	\end{subfigure}
	\hfill
	\begin{subfigure}[b]{0.15\textwidth}
		\centering
		\includegraphics[width=\textwidth]{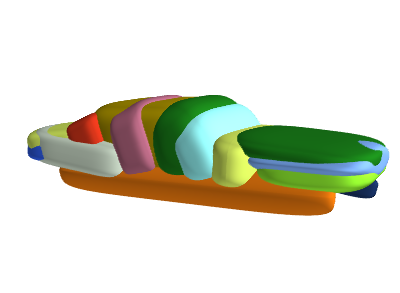}
	\end{subfigure}
	\hfill
	\begin{subfigure}[b]{0.15\textwidth}
		\centering
		\includegraphics[width=\textwidth]{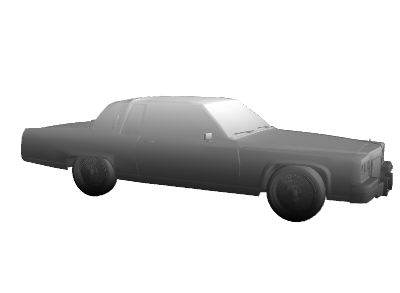}
	\end{subfigure}
	\hfill
	\begin{subfigure}[b]{0.15\textwidth}
		\centering
		\includegraphics[width=\textwidth]{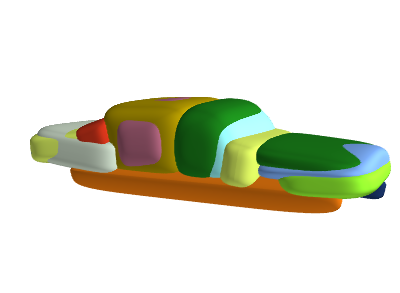}
	\end{subfigure}
    \caption{{\bf Qualitative Results on \emph{cars} from ShapeNet.} We visualize predictions for the object categories \textit{car} from the ShapeNet dataset. Our expressive shape abstractions allow us to differentiate between different car types such as sedan, coupe \etc. This would not have been possible with cuboidal primitives that cannnot model rounded surfaces}
    \label{fig:cars_supp}
\end{figure}

In this section, we provide additional qualitative results on various object types from the ShapeNet dataset \cite{Chang2015ARXIV}. We also demonstrate the ability of our model to capture fine details in more complicated objects such as \textit{motorcycle-bikes} and \textit{cars}. Due to their diverse shape vocabulary, superquadrics can accurately capture the structure of complex objects such as motorbikes and cars. We observe that our model successfully represents the wheels of all bikes using a flattened ellipsoid and the front fork using a pointy ellipsoid. Again, we note that our network consistently associates the same primitive with the same semantic part. For instance, for the motorcycles object category, the primitive colored in red is associated with the saddle, the primitive colored in green is associated with the front wheel \etc. \figref{fig:bikes_supp}$+$\ref{fig:cars_supp} demonstrate several predictions for both classes using superquadric surfaces as geometric primitives.
\begin{figure}[h!]
	\centering
	\begin{subfigure}[b]{0.16\linewidth}
    	\centering
        \includegraphics[width=\linewidth]{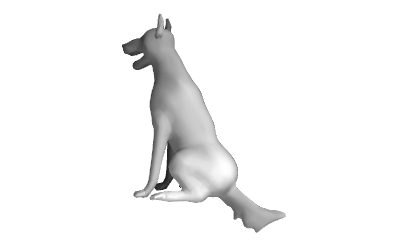}
    \end{subfigure}
    \hfill
    \begin{subfigure}[b]{0.16\linewidth}
    	\centering
        \includegraphics[width=\linewidth]{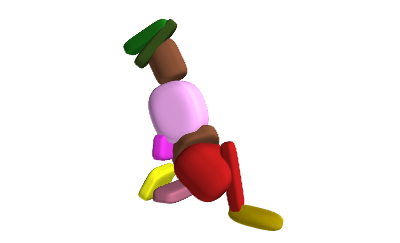}
    \end{subfigure}
    \hfill
    \begin{subfigure}[b]{0.16\linewidth}
    	\centering
        \includegraphics[width=\linewidth]{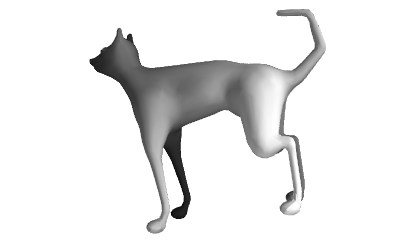}
    \end{subfigure}
    \hfill
    \begin{subfigure}[b]{0.16\linewidth}
    	\centering
        \includegraphics[width=\linewidth]{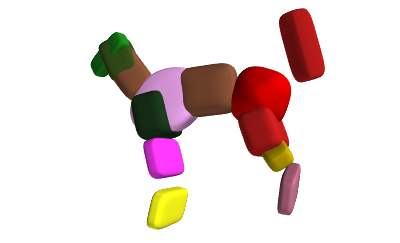}
    \end{subfigure}
    \hfill
    \begin{subfigure}[b]{0.16\linewidth}
    	\centering
        \includegraphics[width=\linewidth]{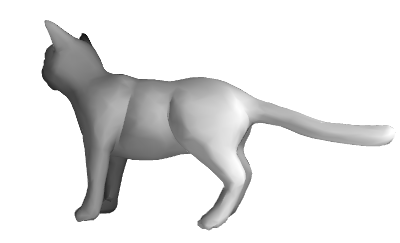}
    \end{subfigure}
    \hfill
    \begin{subfigure}[b]{0.16\linewidth}
    	\centering
        \includegraphics[width=\linewidth]{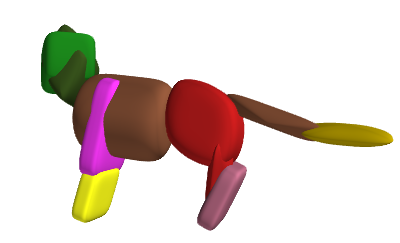}
    \end{subfigure}
    \vskip\baselineskip
	\begin{subfigure}[b]{0.16\linewidth}
    	\centering
        \includegraphics[width=\linewidth]{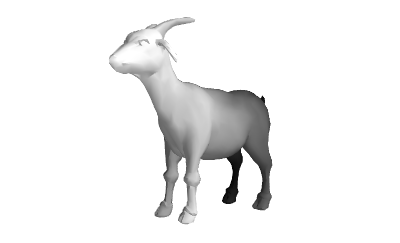}
    \end{subfigure}
    \hfill
    \begin{subfigure}[b]{0.16\linewidth}
    	\centering
        \includegraphics[width=\linewidth]{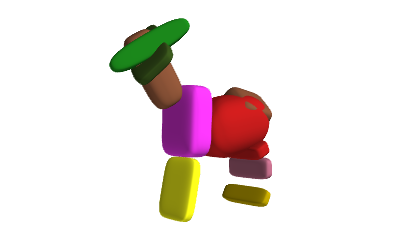}
    \end{subfigure}
    \hfill
	\begin{subfigure}[b]{0.16\linewidth}
    	\centering
        \includegraphics[width=\linewidth]{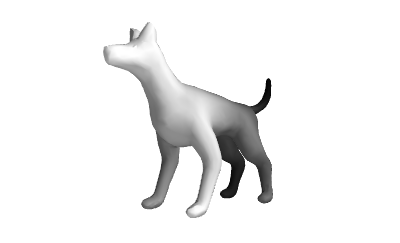}
    \end{subfigure}
    \hfill
    \begin{subfigure}[b]{0.16\linewidth}
    	\centering
        \includegraphics[width=\linewidth]{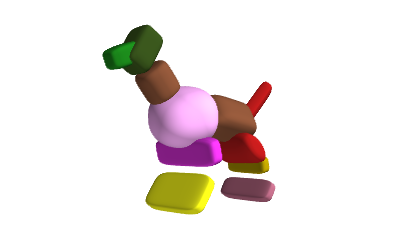}
    \end{subfigure}
    \hfill
	\begin{subfigure}[b]{0.16\linewidth}
    	\centering
        \includegraphics[width=\linewidth]{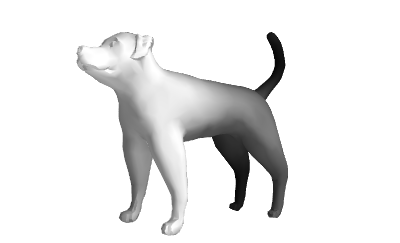}
    \end{subfigure}
    \hfill
    \begin{subfigure}[b]{0.16\linewidth}
    	\centering
        \includegraphics[width=\linewidth]{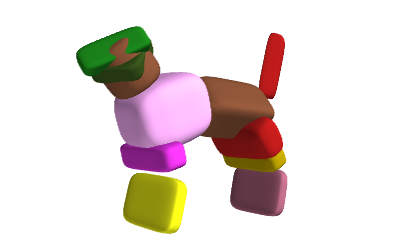}
    \end{subfigure}
    \caption{\textbf{Qualitative Results on \emph{animals} from ShapeNet.} We visualize the predictions of our network on the \emph{animals} class of the ShapeNet dataset. We remark the consistency across primitives and animal parts as well as the ability of our model to cpature details such as ears and tails of animals that could have not beeen captured using cuboidal primitives}
    \label{fig:animals_supp}
    \vspace{-1em}
\end{figure}

Due to superquadrics' large shape vocabulary, our approach derives expressive scene abstractions that allow for differentiating between different types of vehicles both for motorcycles (scooter, racing bike, chopper \etc) (\figref{fig:bikes_supp}), cars (sedan, convertible, coupe, \etc) (\figref{fig:cars_supp}), animals (dogs, cats) (\figref{fig:animals_supp}) despite that our model leverages only up to 20 primitives per object. Note that for cars, wheels are not as easily recovered as for motorbikes due to the lack of supervision and since they are ``geometrically occluded'' by the body of the car.
\begin{figure}[h!]
	\centering
	\begin{subfigure}[b]{0.16\linewidth}
    	\centering
        \includegraphics[width=\linewidth]{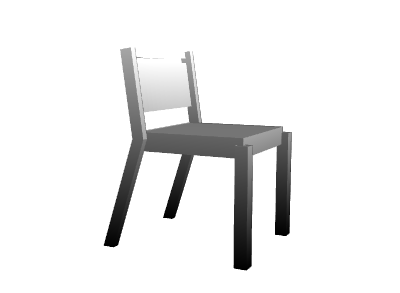}
    \end{subfigure}
    \hfill
    \begin{subfigure}[b]{0.16\linewidth}
    	\centering
        \includegraphics[width=\linewidth]{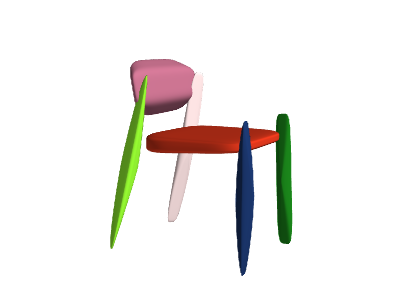}
    \end{subfigure}
    \hfill
    \begin{subfigure}[b]{0.16\linewidth}
    	\centering
        \includegraphics[width=\linewidth]{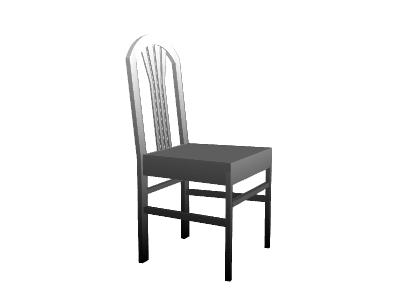}
    \end{subfigure}
    \hfill
    \begin{subfigure}[b]{0.16\linewidth}
    	\centering
        \includegraphics[width=\linewidth]{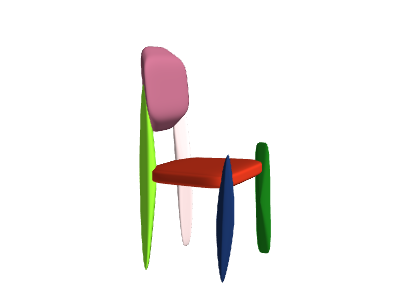}
    \end{subfigure}
    \hfill
    \begin{subfigure}[b]{0.16\linewidth}
    	\centering
        \includegraphics[width=\linewidth]{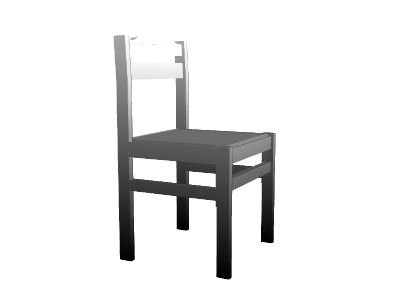}
    \end{subfigure}
    \hfill
    \begin{subfigure}[b]{0.16\linewidth}
    	\centering
        \includegraphics[width=\linewidth]{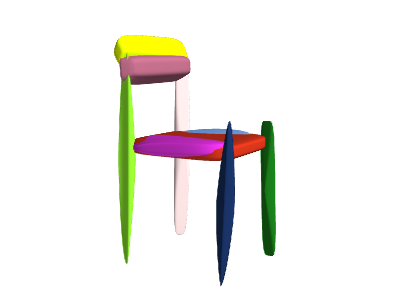}
    \end{subfigure}
    \vskip\baselineskip
	\begin{subfigure}[b]{0.16\linewidth}
    	\centering
        \includegraphics[width=\linewidth]{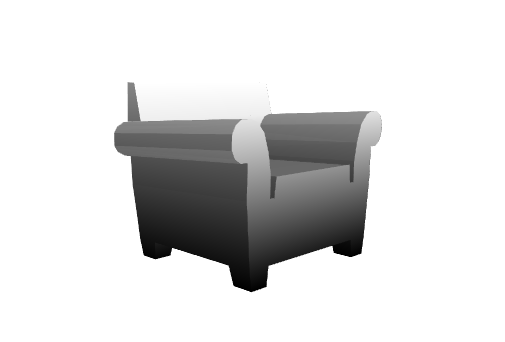}
    \end{subfigure}
    \hfill
    \begin{subfigure}[b]{0.16\linewidth}
    	\centering
        \includegraphics[width=\linewidth]{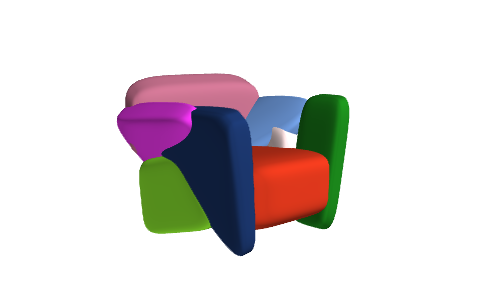}
    \end{subfigure}
    \hfill
	\begin{subfigure}[b]{0.16\linewidth}
    	\centering
        \includegraphics[width=\linewidth]{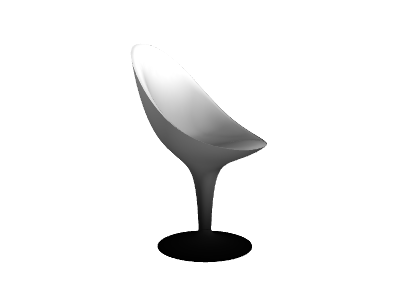}
    \end{subfigure}
    \hfill
    \begin{subfigure}[b]{0.16\linewidth}
    	\centering
        \includegraphics[width=\linewidth]{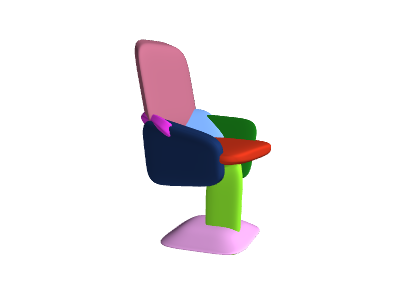}
    \end{subfigure}
    \hfill
	\begin{subfigure}[b]{0.16\linewidth}
    	\centering
        \includegraphics[width=\linewidth]{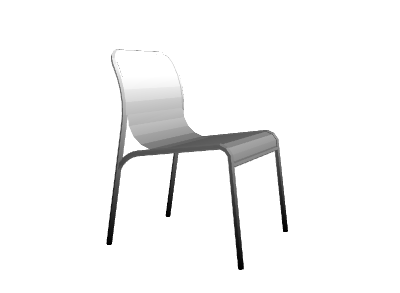}
    \end{subfigure}
    \hfill
    \begin{subfigure}[b]{0.16\linewidth}
    	\centering
        \includegraphics[width=\linewidth]{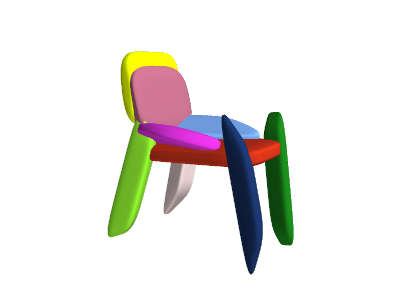}
    \end{subfigure}
    \caption{\textbf{Qualitative Results on \emph{chairs} from ShapeNet.} We visualize the predictions of our network on the \emph{chairs} class of the ShapeNet dataset. We observe the consistency across corespondences between primitives and object parts as well as the ability of our model to capture the shape of rounded parts.}
    \label{fig:chairs_supp}
    \vspace{-1em}
\end{figure}

\figref{fig:animals_supp}$+$\ref{fig:chairs_supp} depicts additional predictions on the \emph{animal} and the \emph{chair} object class of the ShapeNet dataset. We observe that for both categories our model consistently captures both the structure and the fine details of the depicted object. Note that chairs that have rounded legs are associated with flattened ellipsoids (\figref{fig:chairs_supp}), this would not have been possible only with cuboids.

\pagebreak
\clearpage
\newpage
\section{Network Architecture Details}

\begin{figure}[h!]
	\centering
	\includegraphics[width=1.0\textwidth]{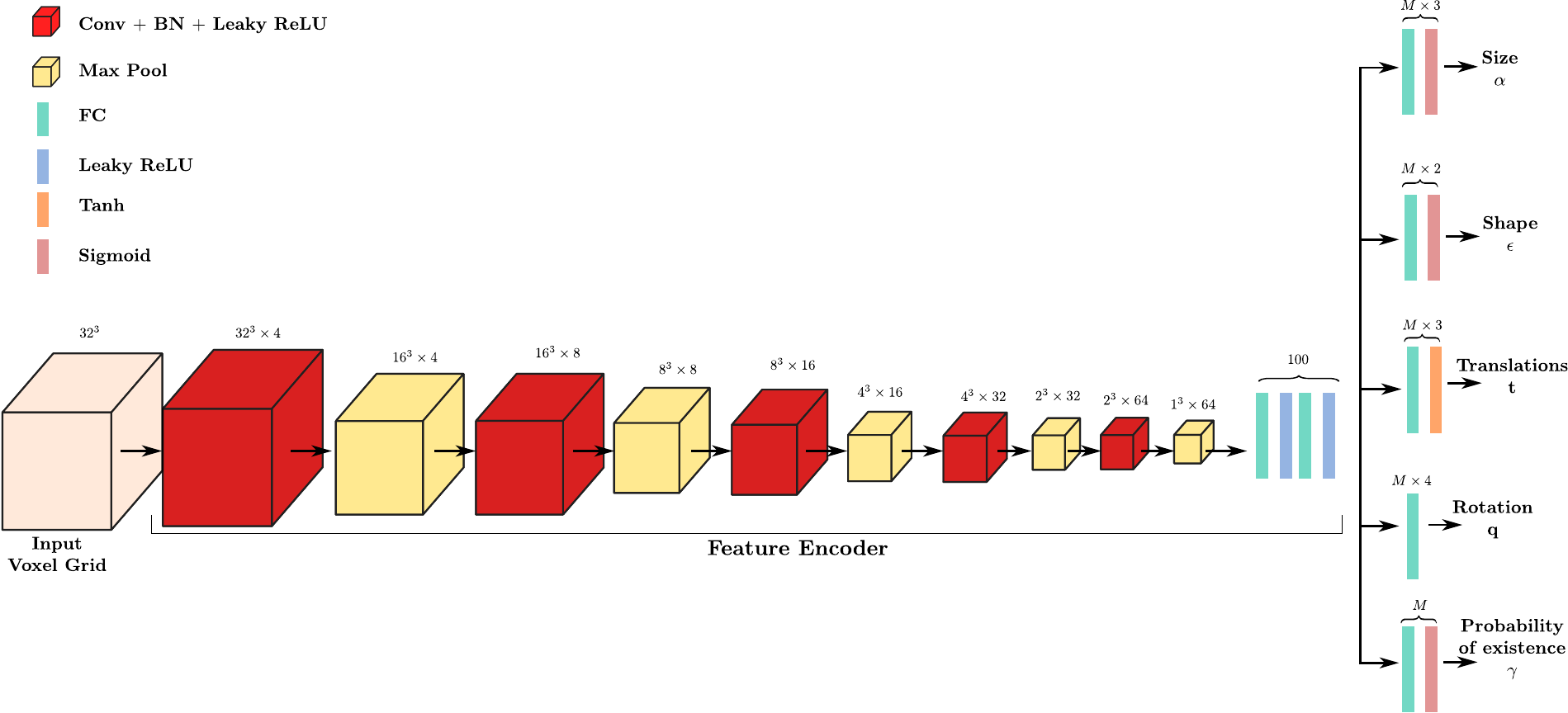}
	\caption{\textbf{Volume-based network architecture.} We visualize the layers that comprise our network architecture. Cubes denote operations that are conducted on 3-dimensional volumes, while rectangles correspond to $K$-dimensional features. The number above each shape (cube or rectangle) corresponds to the dimensionality of that layer. For instance, $16^3 \times 4$ denotes a feature map of size $16^3$ and $4$ channels. Following, our notation, $M$ corresponds to the maximum number of primitives predicted.}
	\label{fig:network_1}
\end{figure}

In this section, we detail the network architecture used throughout our experimental evaluations. Our network comprises of two main parts, an encoder that learns a low-dimensional feature representation for the input and five regressors that predict the parameters of the superquadrics (size $\alpha$, shape $\epsilon$, translations $\mathbf{t}$, rotations $\mathbf{q}$ and $\gamma$ probabilities of existence). As we already explained in our main submission, the encoder architecture is chosen based on the input type (image, voxelized input \etc). In our experiments, we consider a binary occupancy grid as an input and the sequence of layers comprising both the encoder and the regressors are depicted in Figure~\ref{fig:network_1}.

Note that, for the image-based experiment of section~\ref{sec:sqs_from_rgb} in the supplementary, where we consider an image as an input to our model, we replace the encoder architecture in \figref{fig:network_1} with a ResNet18 \cite{He2016CVPR}.

\subsection{Parsimony Loss Details}

We would also like to briefly provide some additional details for our parsimony loss. For completeness, we restate the parsimony loss of Equation $12$ in our main submission,
\begin{equation}
\mathcal{L}_{\gamma}(\bP) = \max\left(\alpha-\alpha\sum_{m=1}^M \gamma_m, 0\right)
+ \beta \sqrt{\sum_{m=1}^M \gamma_m}
\label{eq:loss_gamma_supp}
\end{equation}

Note that the $\displaystyle \sum_{m=1}^M \gamma_m$ corresponds to the expected number of primitives in the predicted parsing. As already mentioned, in our main submission, our model suffers from the trivial solution $\cL_{D}(\bP,\bX)=0$ which is attained for $\gamma_1=\cdots=\gamma_m=0$. To avoid this solution, we introduce the first term of Eq.~\ref{eq:loss_gamma_supp} that penalizes the prediction when the expected number of primitives is less than $1$. The second term penalizes the prediction when the expected number of primitives is large. Note that the maximum value of the second term is $\beta \sqrt{M}$, while the maximum value of the first term is $\alpha$. Therefore, in order to allow the model to use more than one primitive, we set $\beta$ to a value smaller than $\alpha$. Typically $\alpha=1.0$ and $\beta=10^{-3}$.

\pagebreak
\section{Shape Abstraction from a Single RGB Image}
\label{sec:sqs_from_rgb}

In this section, we use the proposed reconstruction loss of Eq. $3$, in the main submission, to extract shape primitives from RGB images instead of occupancy grids. Towards this goal, we render the ShapeNet models to images, and train an image-based network to minimize the same reconstruction loss also used for our volume-based architecture.

More specifically, we replace the encoder architecture, described in Section $3.4$ in our main submission, with the ResNet18 architecture \cite{He2016CVPR}, without the last fully connected layer. The extracted features are subsequently passed to five independent heads that regress translation $\bf{t}$, rotation $\bf{q}$, size $\bf{\alpha}$, shape $\bf{\epsilon}$ and probability of existence $\bf{\gamma}$ for each primitive.
During training, we uniformly sample $1000$ points, from the surface of the target object, as well as $200$ points from the surface of every superquadric. For optimization, we use ADAM \cite{Kingma2015ICLR} with a learning rate of $0.001$ and a batch size of $32$ for $40$k iterations.
We observe that our model accurately captures shape primitives even from a single RGB image as input.

\begin{figure}[h!]
	\centering
	\begin{subfigure}[b]{0.19\textwidth}
		\centering
		\includegraphics[width=0.7\textwidth]{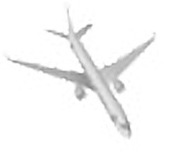}
	\end{subfigure}
	\hfill
	\begin{subfigure}[b]{0.19\textwidth}
		\centering
		\includegraphics[width=0.7\textwidth]{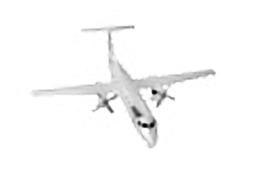}
	\end{subfigure}
	\hfill
	\begin{subfigure}[b]{0.19\textwidth}
		\centering
		\includegraphics[width=0.7\textwidth]{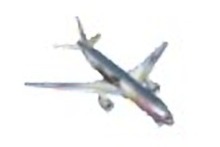}
	\end{subfigure}
	\hfill
	\begin{subfigure}[b]{0.19\textwidth}
		\centering
		\includegraphics[width=0.7\textwidth]{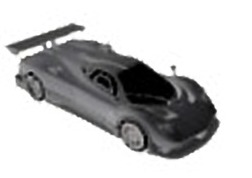}
	\end{subfigure}
	\vskip\baselineskip
	\begin{subfigure}[b]{0.19\textwidth}
		\centering
		\includegraphics[width=0.7\textwidth]{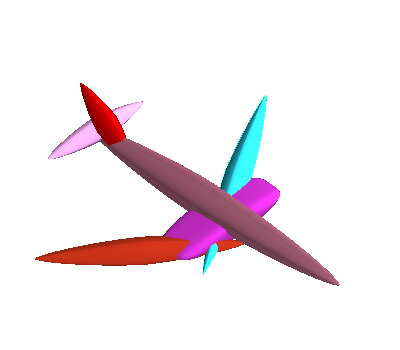}
	\end{subfigure}
	\hfill
	\begin{subfigure}[b]{0.19\textwidth}
		\centering
		\includegraphics[width=0.7\textwidth]{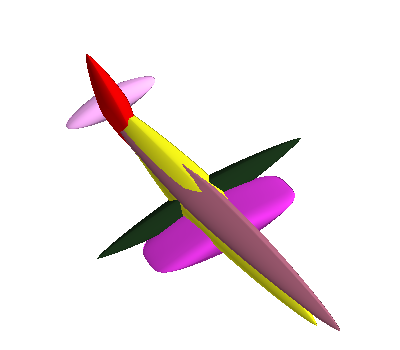}
	\end{subfigure}
	\hfill
	\begin{subfigure}[b]{0.19\textwidth}
		\centering
		\includegraphics[width=0.7\textwidth]{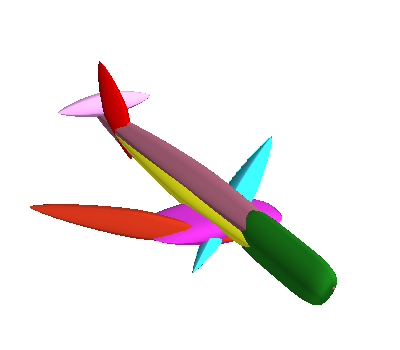}
	\end{subfigure}
	\hfill
	\begin{subfigure}[b]{0.19\textwidth}
		\centering
		\includegraphics[width=.7\textwidth]{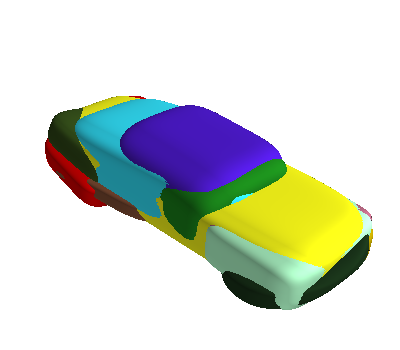}
	\end{subfigure}
	\vskip\baselineskip
	\begin{subfigure}[b]{0.19\textwidth}
		\centering
		\includegraphics[width=.5\textwidth]{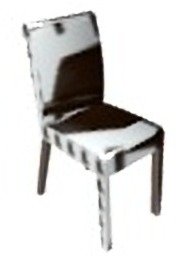}
	\end{subfigure}
	\hfill
	\begin{subfigure}[b]{0.19\textwidth}
		\centering
		\includegraphics[width=.5\textwidth]{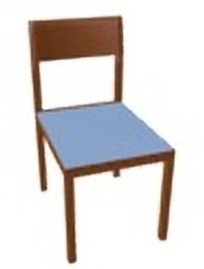}
	\end{subfigure}
	\hfill
	\begin{subfigure}[b]{0.19\textwidth}
		\centering
		\includegraphics[width=0.5\textwidth]{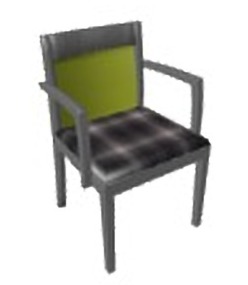}
	\end{subfigure}
	\hfill
	\begin{subfigure}[b]{0.19\textwidth}
		\centering
		\includegraphics[width=0.7\textwidth]{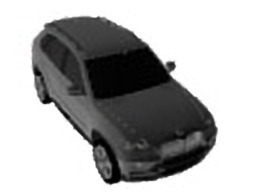}
	\end{subfigure}
	\vskip\baselineskip
	\begin{subfigure}[b]{0.19\textwidth}
		\centering
		\includegraphics[width=0.7\textwidth]{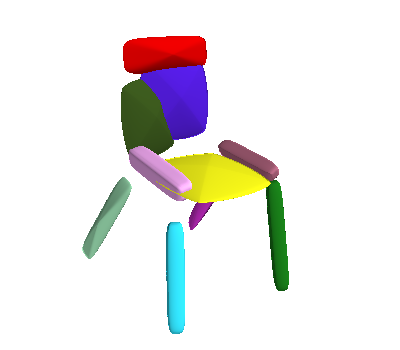}
	\end{subfigure}
	\hfill
	\begin{subfigure}[b]{0.19\textwidth}
		\centering
		\includegraphics[width=0.7\textwidth]{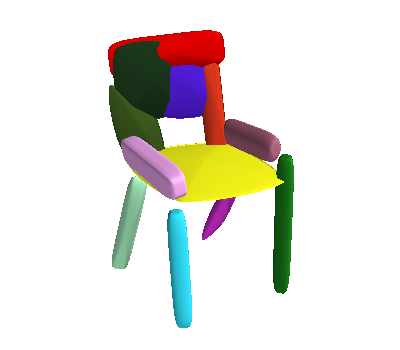}
	\end{subfigure}
	\hfill
	\begin{subfigure}[b]{0.19\textwidth}
		\centering
		\includegraphics[width=0.7\textwidth]{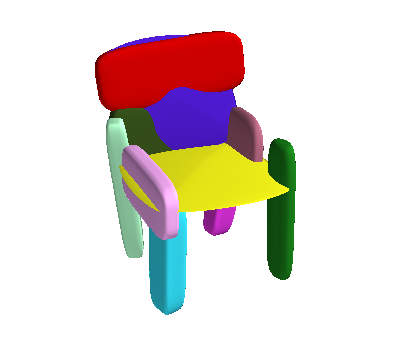}
	\end{subfigure}
	\hfill
	\begin{subfigure}[b]{0.19\textwidth}
		\centering
		\includegraphics[width=0.7\textwidth]{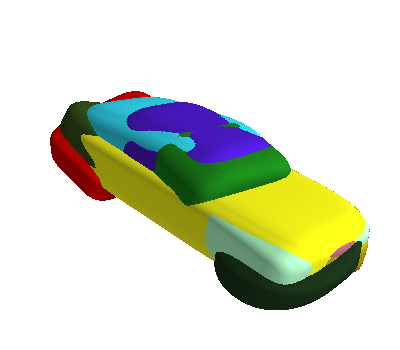}
	\end{subfigure}
    \caption{{\bf Shape Abstraction from a Single RGB Image.} We visualize predictions for various ShapeNet object categories using a single RGB image as input to our model.}
\end{figure}

\section{Quantitative Analysis}

In this section, we provide additional details regarding the quantitative comparison of Table 1 in our main paper. For evaluation, we report two metrics the mean \emph{Chamfer distance} and the mean \emph{Volumetric IoU}.
Volumetric IoU is defined as the quotient of the volume of the two meshes’ intersection and the volume of their union. We obtain unbiased estimates of the volume of the intersection and the union by randomly sampling points from the bounding volume and determining if the points lie inside our outside the ground truth / predicted mesh. The computation of the Chamfer distance is discussed in detail in our main submission throughout Section 3. Regarding the comparison in Table 1 of our main submission, we want to mention that cuboids are a special case of superquadrics, thus fitting objects with cuboids is expected to lead to worse results compared to superquadrics.

\section{Empirical Analysis of Reconstruction Loss}

\begin{figure}[t!]
	\centering
	\begin{subfigure}[b]{0.19\linewidth}
		\centering
		\includegraphics[width=\linewidth]{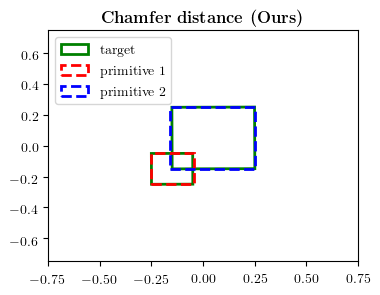}
	\end{subfigure}
	\hfill
	\begin{subfigure}[b]{0.19\linewidth}
		\centering
		\includegraphics[width=\linewidth]{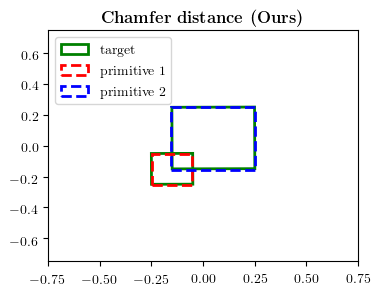}
	\end{subfigure}
	\hfill
	\begin{subfigure}[b]{0.19\linewidth}
		\centering
		\includegraphics[width=\linewidth]{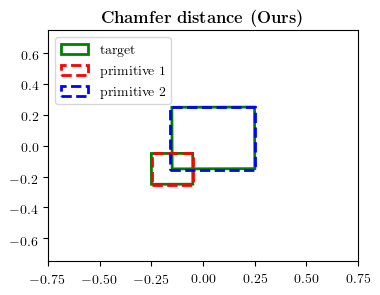}
	\end{subfigure}
	\hfill
	\begin{subfigure}[b]{0.19\linewidth}
		\centering
		\includegraphics[width=\linewidth]{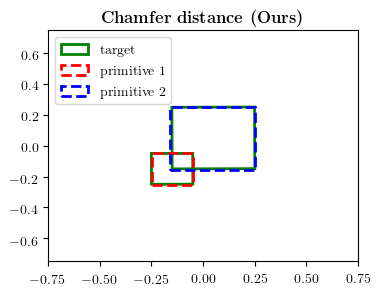}
	\end{subfigure}
	\hfill
	\begin{subfigure}[b]{0.19\linewidth}
		\centering
		\includegraphics[width=\linewidth]{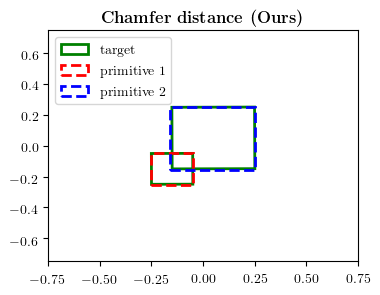}
	\end{subfigure}
	\vskip\baselineskip
	\begin{subfigure}[b]{0.19\linewidth}
		\centering
		\includegraphics[width=\linewidth]{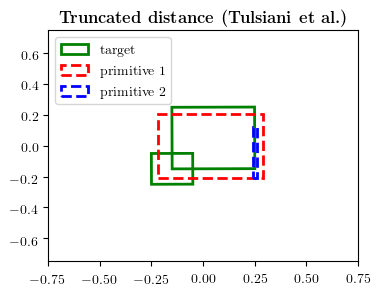}
	\end{subfigure}
	\hfill
	\begin{subfigure}[b]{0.19\linewidth}
		\centering
		\includegraphics[width=\linewidth]{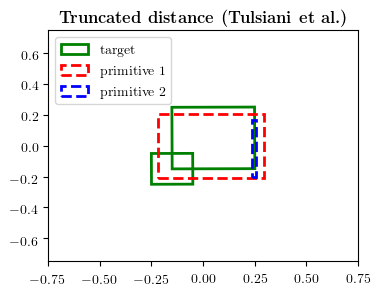}
	\end{subfigure}
	\hfill
	\begin{subfigure}[b]{0.19\linewidth}
		\centering
		\includegraphics[width=\linewidth]{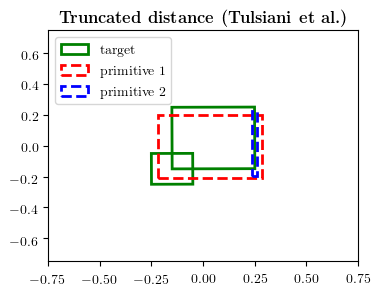}
	\end{subfigure}
	\hfill
	\begin{subfigure}[b]{0.19\linewidth}
		\centering
		\includegraphics[width=\linewidth]{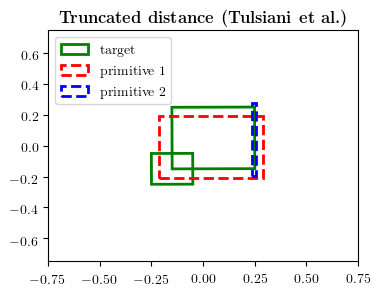}
	\end{subfigure}
	\hfill
	\begin{subfigure}[b]{0.19\linewidth}
		\centering
		\includegraphics[width=\linewidth]{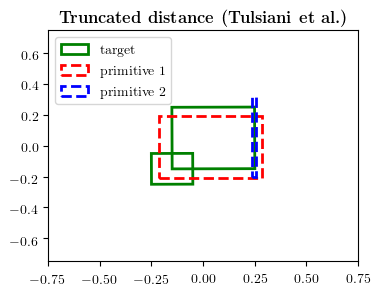}
	\end{subfigure}
	\vskip\baselineskip
	\begin{subfigure}[b]{0.19\linewidth}
		\centering
        \caption{\textbf{Iteration $150$}}
	\end{subfigure}
    \vspace{-0.2em}
	\hfill
	\begin{subfigure}[b]{0.19\linewidth}
		\centering
        \caption{\textbf{Iteration $160$}}
	\end{subfigure}
    \vspace{-0.2em}
	\hfill
	\begin{subfigure}[b]{0.19\linewidth}
		\centering
        \caption{\textbf{Iteration $170$}}
	\end{subfigure}
    \vspace{-0.2em}
	\hfill
	\begin{subfigure}[b]{0.19\linewidth}
		\centering
        \caption{\textbf{Iteration $180$}}
	\end{subfigure}
    \vspace{-0.2em}
	\hfill
	\begin{subfigure}[b]{0.19\linewidth}
		\centering
        \caption{\textbf{Iteration $190$}}
	\end{subfigure}
    \vspace{-0.2em}
	\vskip\baselineskip

	\begin{subfigure}[b]{0.19\linewidth}
		\centering
		\includegraphics[width=\linewidth]{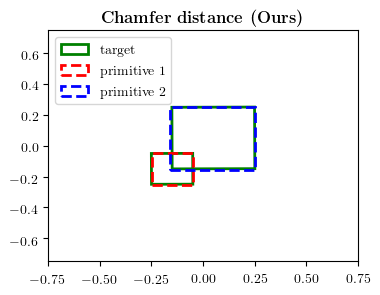}
	\end{subfigure}
	\hfill
	\begin{subfigure}[b]{0.19\linewidth}
		\centering
		\includegraphics[width=\linewidth]{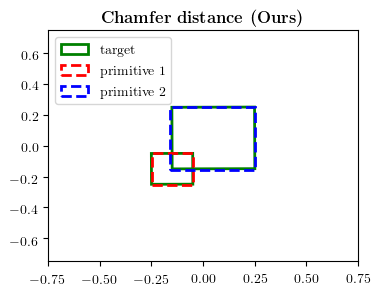}
	\end{subfigure}
	\hfill
	\begin{subfigure}[b]{0.19\linewidth}
		\centering
		\includegraphics[width=\linewidth]{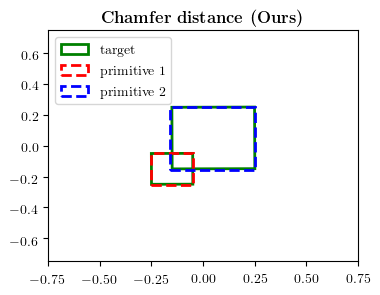}
	\end{subfigure}
	\hfill
	\begin{subfigure}[b]{0.19\linewidth}
		\centering
		\includegraphics[width=\linewidth]{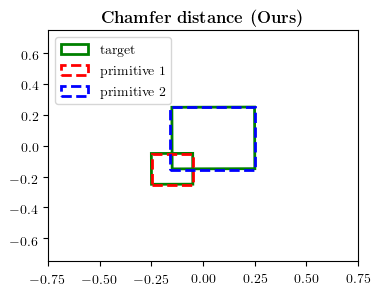}
	\end{subfigure}
	\hfill
	\begin{subfigure}[b]{0.19\linewidth}
		\centering
		\includegraphics[width=\linewidth]{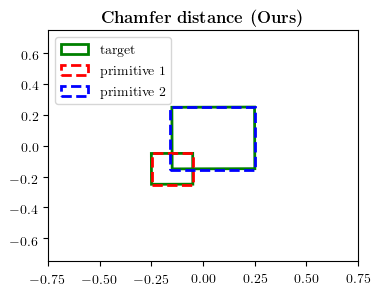}
	\end{subfigure}
	\vskip\baselineskip
	\begin{subfigure}[b]{0.19\linewidth}
		\centering
		\includegraphics[width=\linewidth]{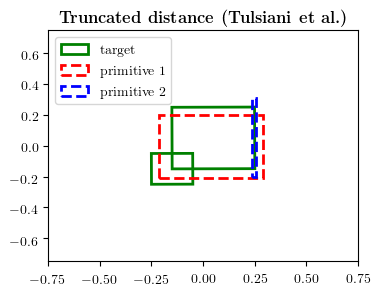}
	\end{subfigure}
	\hfill
	\begin{subfigure}[b]{0.19\linewidth}
		\centering
		\includegraphics[width=\linewidth]{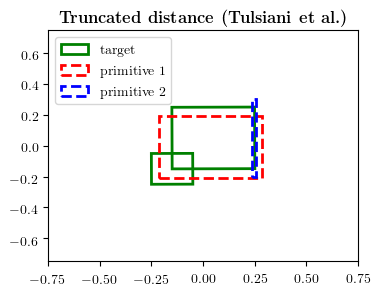}
	\end{subfigure}
	\hfill
	\begin{subfigure}[b]{0.19\linewidth}
		\centering
		\includegraphics[width=\linewidth]{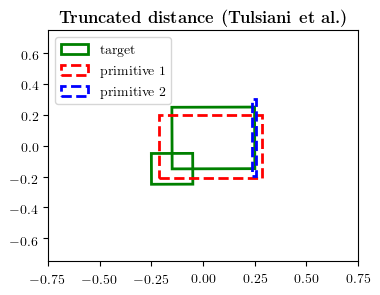}
	\end{subfigure}
	\hfill
	\begin{subfigure}[b]{0.19\linewidth}
		\centering
		\includegraphics[width=\linewidth]{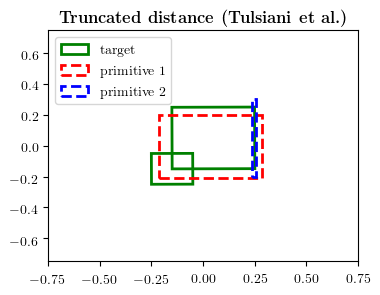}
	\end{subfigure}
	\hfill
	\begin{subfigure}[b]{0.19\linewidth}
		\centering
		\includegraphics[width=\linewidth]{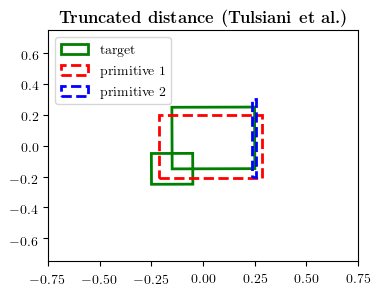}
	\end{subfigure}
	\vskip\baselineskip
	\begin{subfigure}[b]{0.19\linewidth}
		\centering
        \vspace{-0.2em}
        \caption{\textbf{Iteration $200$}}
	\end{subfigure}
    \vspace{-0.2em}
	\hfill
	\begin{subfigure}[b]{0.19\linewidth}
		\centering
        \vspace{-0.2em}
        \caption{\textbf{Iteration $210$}}
	\end{subfigure}
    \vspace{-0.2em}
	\hfill
	\begin{subfigure}[b]{0.19\linewidth}
		\centering
        \vspace{-0.2em}
        \caption{\textbf{Iteration $220$}}
	\end{subfigure}
    \vspace{-0.2em}
	\hfill
	\begin{subfigure}[b]{0.19\linewidth}
		\centering
        \vspace{-0.2em}
        \caption{\textbf{Iteration $230$}}
	\end{subfigure}
    \vspace{-0.2em}
	\hfill
	\begin{subfigure}[b]{0.19\linewidth}
		\centering
        \vspace{-0.2em}
        \caption{\textbf{Iteration $240$}}
	\end{subfigure}
    \vspace{-0.2em}
	\vskip\baselineskip
	\begin{subfigure}[b]{0.19\linewidth}
		\centering
		\includegraphics[width=\linewidth]{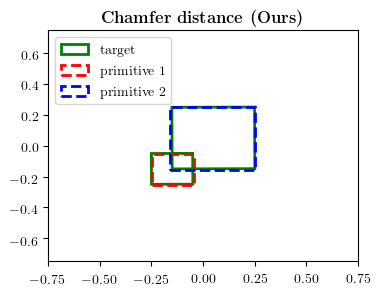}
	\end{subfigure}
	\hfill
	\begin{subfigure}[b]{0.19\linewidth}
		\centering
		\includegraphics[width=\linewidth]{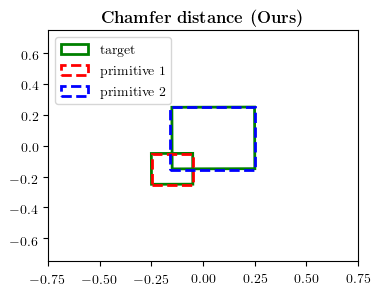}
	\end{subfigure}
	\hfill
	\begin{subfigure}[b]{0.19\linewidth}
		\centering
		\includegraphics[width=\linewidth]{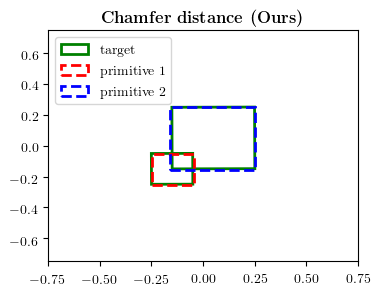}
	\end{subfigure}
	\hfill
	\begin{subfigure}[b]{0.19\linewidth}
		\centering
		\includegraphics[width=\linewidth]{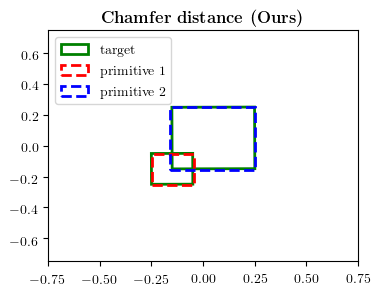}
	\end{subfigure}
	\hfill
	\begin{subfigure}[b]{0.19\linewidth}
		\centering
		\includegraphics[width=\linewidth]{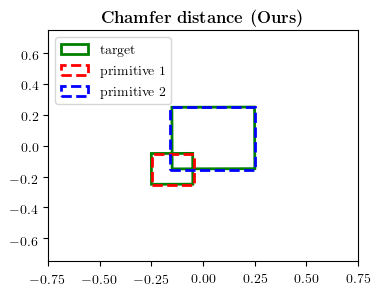}
	\end{subfigure}
	\vskip\baselineskip
	\begin{subfigure}[b]{0.19\linewidth}
		\centering
		\includegraphics[width=\linewidth]{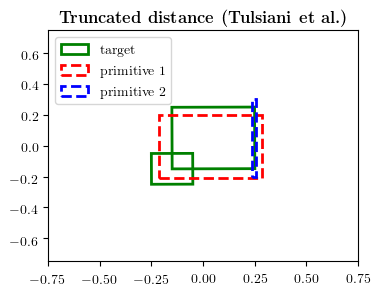}
	\end{subfigure}
	\hfill
	\begin{subfigure}[b]{0.19\linewidth}
		\centering
		\includegraphics[width=\linewidth]{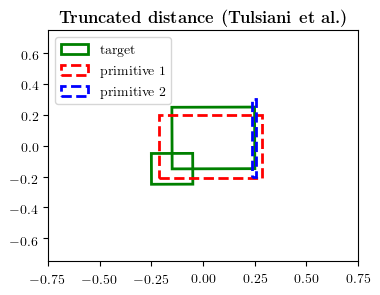}
	\end{subfigure}
	\hfill
	\begin{subfigure}[b]{0.19\linewidth}
		\centering
		\includegraphics[width=\linewidth]{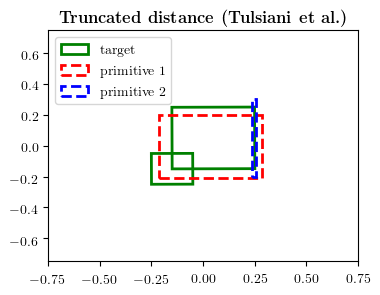}
	\end{subfigure}
	\hfill
	\begin{subfigure}[b]{0.19\linewidth}
		\centering
		\includegraphics[width=\linewidth]{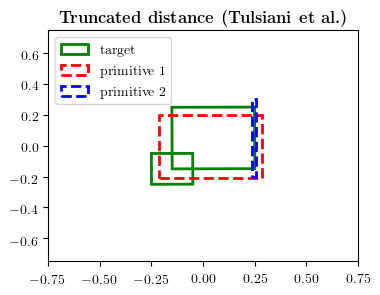}
	\end{subfigure}
	\hfill
	\begin{subfigure}[b]{0.19\linewidth}
		\centering
		\includegraphics[width=\linewidth]{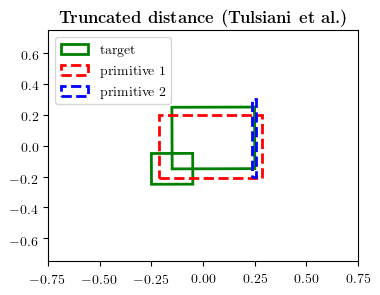}
	\end{subfigure}
	\vskip\baselineskip
	\begin{subfigure}[b]{0.19\linewidth}
		\centering
        \vspace{-0.2em}
        \caption{\textbf{Iteration $250$}}
        \label{fig:local_min_start}
	\end{subfigure}
	\hfill
	\begin{subfigure}[b]{0.19\linewidth}
		\centering
        \vspace{-0.2em}
        \caption{\textbf{Iteration $260$}}
	\end{subfigure}
	\hfill
	\begin{subfigure}[b]{0.19\linewidth}
		\centering
        \vspace{-0.2em}
        \caption{\textbf{Iteration $270$}}
	\end{subfigure}
	\hfill
	\begin{subfigure}[b]{0.19\linewidth}
		\centering
        \vspace{-0.2em}
        \caption{\textbf{Iteration $280$}}
	\end{subfigure}
	\hfill
	\begin{subfigure}[b]{0.19\linewidth}
		\centering
        \vspace{-0.2em}
        \caption{\textbf{Iteration $290$}}
        \label{fig:local_min_end}
	\end{subfigure}
    \caption{\textbf{Empirical Analysis of Reconstruction Loss}. We illustrate the evolution of two cuboid abstractions using our reconstruction loss with Chamfer distance and the truncated bi-directional loss of Tulsiani \etal \cite{Tulsiani2017CVPRa}.}
    \vspace{-0.2em}
\end{figure}

In this section, we provide empirical evidence regarding our claim that our Chamfer-based reconstruction loss leads to more stable training compared to the truncated bi-directional loss of Tulsiani \etal \cite{Tulsiani2017CVPRa}. Towards this goal, we directly optimize/train for the primitive parameters, \ie, not optimizing the weights of a neural network but directly fitting the primitives.
We perform this experiment on a 2D toy example 
and compare the results when using the proposed loss to the results using the truncated distance formulation in \cite{Tulsiani2017CVPRa}.
We visualize the evolution of parameters for both optimization objectives as training progresses.
We observe that the truncated loss proposed in \cite{Tulsiani2017CVPRa} is more likely to converge to local minima (\eg figures \ref{fig:local_min_start}-\ref{fig:local_min_end}), while our loss consistently avoids them.

\begin{figure}[h!]
	\centering
	\begin{subfigure}[b]{0.19\linewidth}
		\centering
		\includegraphics[width=\linewidth]{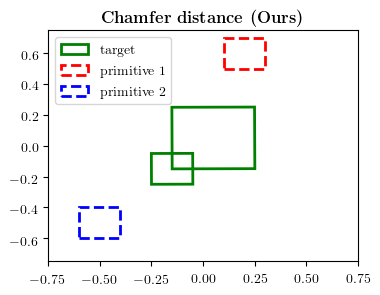}
	\end{subfigure}
	\hfill
	\begin{subfigure}[b]{0.19\linewidth}
		\centering
		\includegraphics[width=\linewidth]{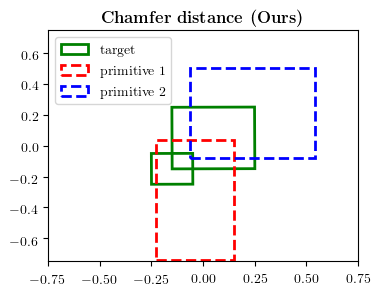}
	\end{subfigure}
	\hfill
	\begin{subfigure}[b]{0.19\linewidth}
		\centering
		\includegraphics[width=\linewidth]{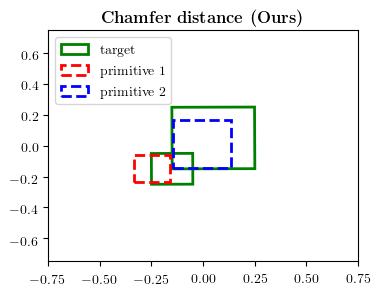}
	\end{subfigure}
	\hfill
	\begin{subfigure}[b]{0.19\linewidth}
		\centering
		\includegraphics[width=\linewidth]{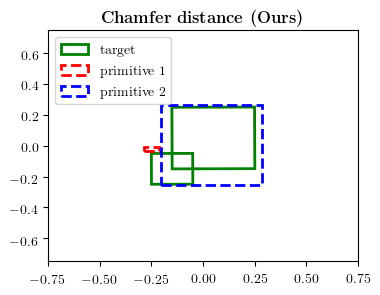}
	\end{subfigure}
	\hfill
	\begin{subfigure}[b]{0.19\linewidth}
		\centering
		\includegraphics[width=\linewidth]{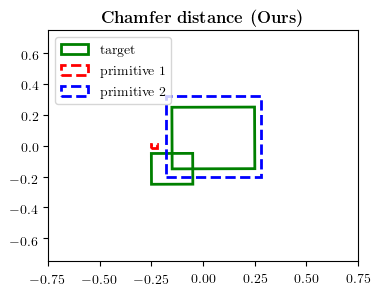}
	\end{subfigure}
	\vskip\baselineskip
	\begin{subfigure}[b]{0.19\linewidth}
		\centering
		\includegraphics[width=\linewidth]{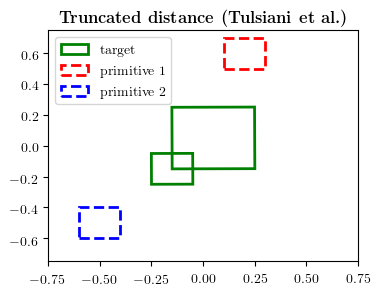}
	\end{subfigure}
	\hfill
	\begin{subfigure}[b]{0.19\linewidth}
		\centering
		\includegraphics[width=\linewidth]{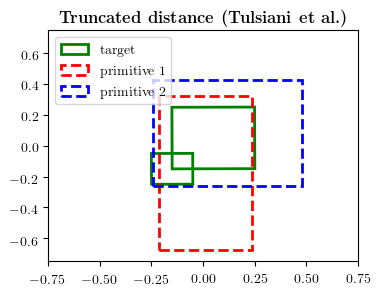}
	\end{subfigure}
	\hfill
	\begin{subfigure}[b]{0.19\linewidth}
		\centering
		\includegraphics[width=\linewidth]{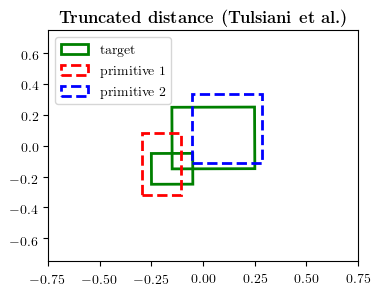}
	\end{subfigure}
	\hfill
	\begin{subfigure}[b]{0.19\linewidth}
		\centering
		\includegraphics[width=\linewidth]{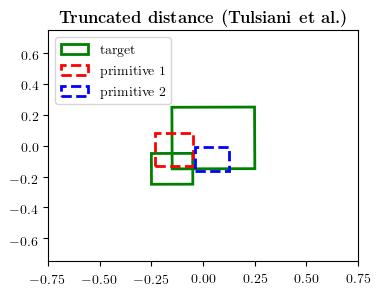}
	\end{subfigure}
	\hfill
	\begin{subfigure}[b]{0.19\linewidth}
		\centering
		\includegraphics[width=\linewidth]{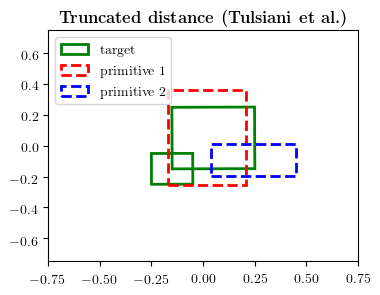}
	\end{subfigure}
	\vskip\baselineskip
	\begin{subfigure}[b]{0.19\linewidth}
		\centering
        \caption{\textbf{Iteration $0$}}
	\end{subfigure}
    \vspace{-0.2em}
	\hfill
	\begin{subfigure}[b]{0.19\linewidth}
		\centering
        \caption{\textbf{Iteration $10$}}
	\end{subfigure}
    \vspace{-0.2em}
	\hfill
	\begin{subfigure}[b]{0.19\linewidth}
		\centering
        \caption{\textbf{Iteration $20$}}
	\end{subfigure}
    \vspace{-0.2em}
	\hfill
	\begin{subfigure}[b]{0.19\linewidth}
		\centering
        \caption{\textbf{Iteration $30$}}
	\end{subfigure}
    \vspace{-0.2em}
	\hfill
	\begin{subfigure}[b]{0.19\linewidth}
		\centering
        \caption{\textbf{Iteration $40$}}
	\end{subfigure}
    \vspace{-0.2em}
	\vskip\baselineskip

	\begin{subfigure}[b]{0.19\linewidth}
		\centering
		\includegraphics[width=\linewidth]{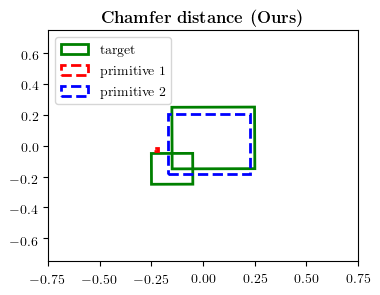}
	\end{subfigure}
	\hfill
	\begin{subfigure}[b]{0.19\linewidth}
		\centering
		\includegraphics[width=\linewidth]{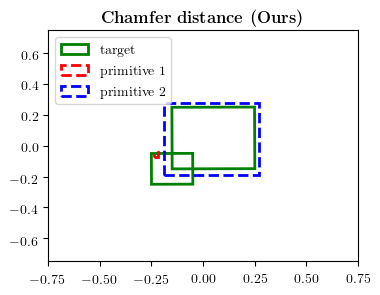}
	\end{subfigure}
	\hfill
	\begin{subfigure}[b]{0.19\linewidth}
		\centering
		\includegraphics[width=\linewidth]{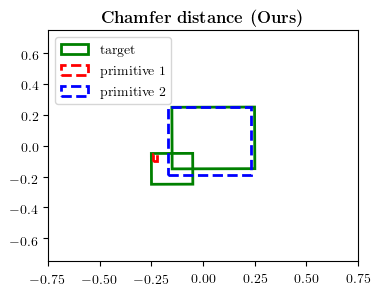}
	\end{subfigure}
	\hfill
	\begin{subfigure}[b]{0.19\linewidth}
		\centering
		\includegraphics[width=\linewidth]{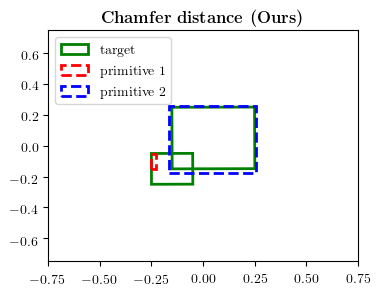}
	\end{subfigure}
	\hfill
	\begin{subfigure}[b]{0.19\linewidth}
		\centering
		\includegraphics[width=\linewidth]{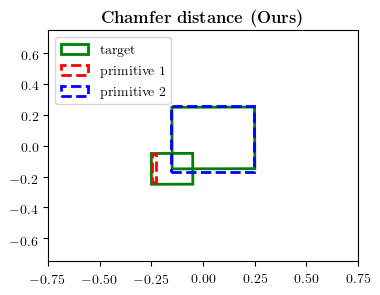}
	\end{subfigure}
	\vskip\baselineskip
	\begin{subfigure}[b]{0.19\linewidth}
		\centering
		\includegraphics[width=\linewidth]{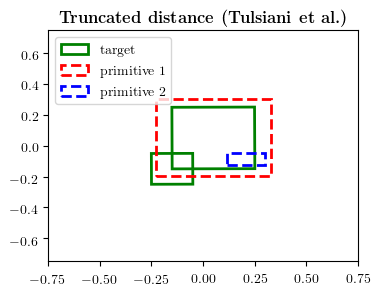}
	\end{subfigure}
	\hfill
	\begin{subfigure}[b]{0.19\linewidth}
		\centering
		\includegraphics[width=\linewidth]{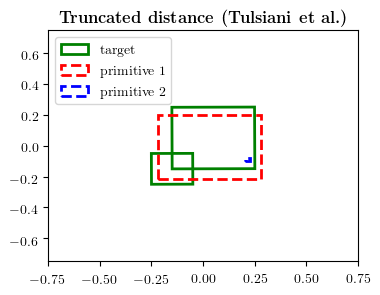}
	\end{subfigure}
	\hfill
	\begin{subfigure}[b]{0.19\linewidth}
		\centering
		\includegraphics[width=\linewidth]{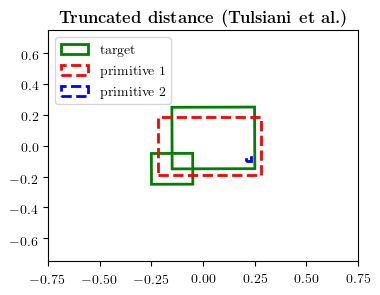}
	\end{subfigure}
	\hfill
	\begin{subfigure}[b]{0.19\linewidth}
		\centering
		\includegraphics[width=\linewidth]{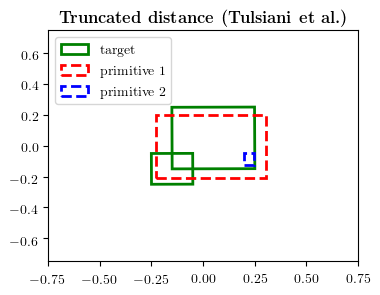}
	\end{subfigure}
	\hfill
	\begin{subfigure}[b]{0.19\linewidth}
		\centering
		\includegraphics[width=\linewidth]{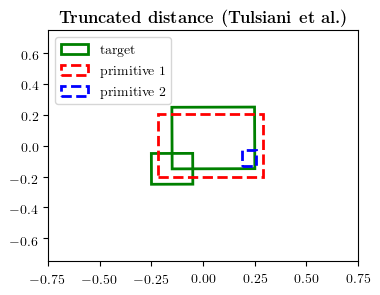}
	\end{subfigure}
	\vskip\baselineskip
	\begin{subfigure}[b]{0.19\linewidth}
		\centering
        \caption{\textbf{Iteration $50$}}
	\end{subfigure}
    \vspace{-0.2em}
	\hfill
	\begin{subfigure}[b]{0.19\linewidth}
		\centering
        \caption{\textbf{Iteration $60$}}
	\end{subfigure}
    \vspace{-0.2em}
	\hfill
	\begin{subfigure}[b]{0.19\linewidth}
		\centering
        \caption{\textbf{Iteration $70$}}
	\end{subfigure}
    \vspace{-0.2em}
	\hfill
	\begin{subfigure}[b]{0.19\linewidth}
		\centering
        \caption{\textbf{Iteration $80$}}
	\end{subfigure}
    \vspace{-0.2em}
	\hfill
	\begin{subfigure}[b]{0.19\linewidth}
		\centering
        \caption{\textbf{Iteration $90$}}
	\end{subfigure}
    \vspace{-0.2em}
	\vskip\baselineskip

	\begin{subfigure}[b]{0.19\linewidth}
		\centering
		\includegraphics[width=\linewidth]{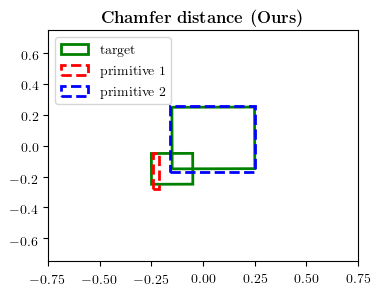}
	\end{subfigure}
	\hfill
	\begin{subfigure}[b]{0.19\linewidth}
		\centering
		\includegraphics[width=\linewidth]{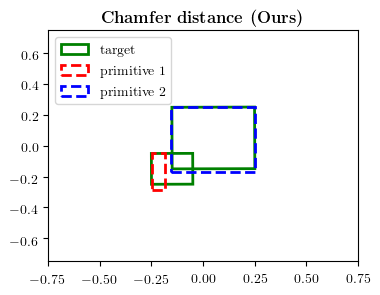}
	\end{subfigure}
	\hfill
	\begin{subfigure}[b]{0.19\linewidth}
		\centering
		\includegraphics[width=\linewidth]{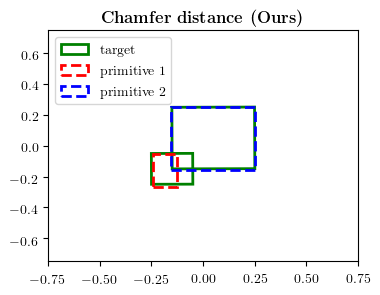}
	\end{subfigure}
	\hfill
	\begin{subfigure}[b]{0.19\linewidth}
		\centering
		\includegraphics[width=\linewidth]{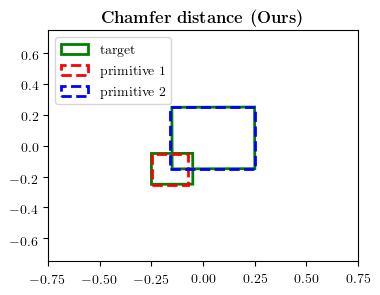}
	\end{subfigure}
	\hfill
	\begin{subfigure}[b]{0.19\linewidth}
		\centering
		\includegraphics[width=\linewidth]{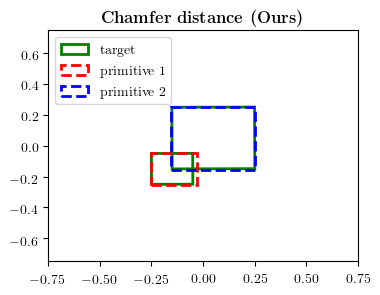}
	\end{subfigure}
	\vskip\baselineskip
	\begin{subfigure}[b]{0.19\linewidth}
		\centering
		\includegraphics[width=\linewidth]{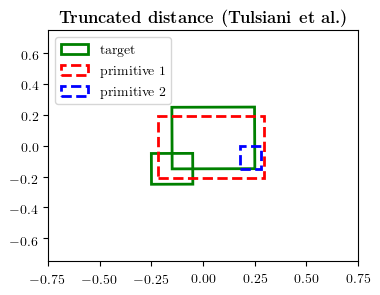}
	\end{subfigure}
	\hfill
	\begin{subfigure}[b]{0.19\linewidth}
		\centering
		\includegraphics[width=\linewidth]{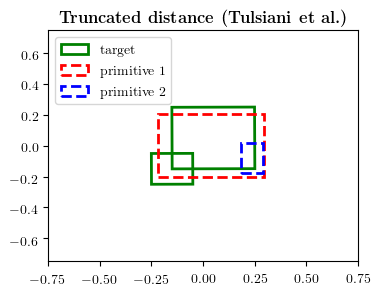}
	\end{subfigure}
	\hfill
	\begin{subfigure}[b]{0.19\linewidth}
		\centering
		\includegraphics[width=\linewidth]{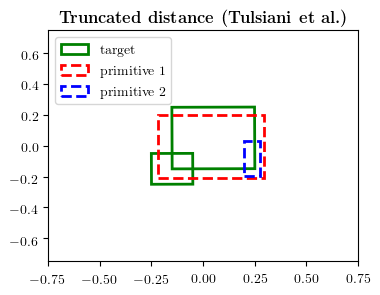}
	\end{subfigure}
	\hfill
	\begin{subfigure}[b]{0.19\linewidth}
		\centering
		\includegraphics[width=\linewidth]{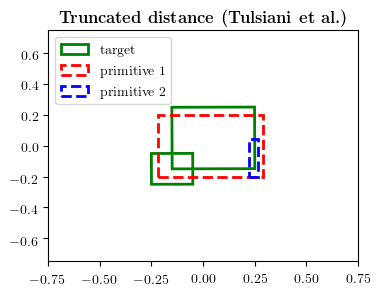}
	\end{subfigure}
	\hfill
	\begin{subfigure}[b]{0.19\linewidth}
		\centering
		\includegraphics[width=\linewidth]{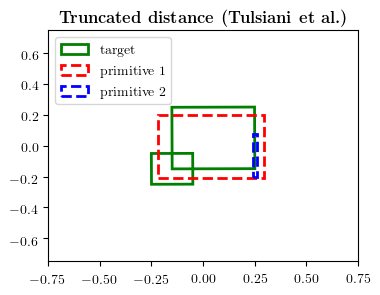}
	\end{subfigure}
	\vskip\baselineskip
	\begin{subfigure}[b]{0.19\linewidth}
		\centering
        \caption{\textbf{Iteration $100$}}
	\end{subfigure}
    \vspace{-0.2em}
	\hfill
	\begin{subfigure}[b]{0.19\linewidth}
		\centering
        \caption{\textbf{Iteration $110$}}
	\end{subfigure}
    \vspace{-0.2em}
	\hfill
	\begin{subfigure}[b]{0.19\linewidth}
		\centering
        \caption{\textbf{Iteration $120$}}
	\end{subfigure}
    \vspace{-0.2em}
	\hfill
	\begin{subfigure}[b]{0.19\linewidth}
		\centering
        \caption{\textbf{Iteration $130$}}
	\end{subfigure}
    \vspace{-0.2em}
	\hfill
	\begin{subfigure}[b]{0.19\linewidth}
		\centering
        \caption{\textbf{Iteration $140$}}
	\end{subfigure}
    \vspace{-0.2em}
\end{figure}
\vspace{-0.2em}

\end{document}